%% file: main.tex
\documentclass[10pt]{article} 

\usepackage[accepted]{tmlr}

\input{math_commands.tex}

\usepackage{url}
\usepackage{graphicx}
\usepackage{lscape}
\usepackage{amsmath}
\usepackage{amsfonts}
\usepackage{amssymb,amsfonts}
\usepackage{dsfont}
\usepackage{subcaption}
\usepackage{foreign}
\usepackage{hyperref}
\usepackage[capitalize]{cleveref}
\usepackage{url}
\usepackage{threeparttable}
\usepackage{epigraph}
\usepackage{tikz}
\usepackage{array} 
\usepackage{wrapfig}
\usepackage{booktabs}
\usepackage{multirow}
\usepackage{pifont}
\usepackage{xspace}
\definecolor{ForestGreen}{RGB}{24,139,34}%
\definecolor{BrickRed}{RGB}{203,65,84}%
\makeatletter
\DeclareRobustCommand\onedot{\futurelet\@let@token\@onedot}
\def\@onedot{\ifx\@let@token.\else.\null\fi\xspace}
\def\eg{\emph{e.g}\onedot} 
\def\ie{\emph{i.e}\onedot}

\def\wrt{w.r.t\onedot} 

\newcommand{\Update}[1]{#1}
\newcommand{\CameraReady}[1]{#1}

\title{\textit{What Time Tells Us?} An Explorative Study of\\ Time Awareness Learned from Static Images}

\author{
\name Dongheng Lin\thanks{These authors contributed equally.} \quad
Han Hu\footnotemark[1] \quad
Jianbo Jiao\thanks{Corresponding Author}
\email \{d.lin.2@, hxh347@student., j.jiao@\}bham.ac.uk \\
\addr The MIx Group, University of Birmingham, Birmingham, UK \\
Project page: \url{https://rathgrith.github.io/timetells_release/}
}

\def\month{09}  
\def\year{2025} 
\def\openreview{\url{https://openreview.net/forum?id=f1MYOG4iDG}} 

\begin{document}

\maketitle

\input{sec/0_abstract}  
\input{sec/1_intro}
\input{sec/2_related_works}

\input{sec/3_methodology}

\input{sec/4_dataset}

\input{sec/5_experiment}

\input{sec/6_conclusion}

\appendix
\input{sec/X_suppl}
\newpage
\bibliography{main}
\bibliographystyle{tmlr}

\end{document}

%% file: math_commands.tex
\usepackage{amsmath,amsfonts,bm}

\def\eqref#1{equation~\ref{#1}}

\def\1{\bm{1}}

\DeclareMathAlphabet{\mathsfit}{\encodingdefault}{\sfdefault}{m}{sl}
\SetMathAlphabet{\mathsfit}{bold}{\encodingdefault}{\sfdefault}{bx}{n}



%% file: sec/0_abstract.tex
\begin{abstract}
Time becomes visible through illumination changes in what we see. Inspired by this, in this paper we explore the potential to learn time awareness from static images, trying to answer: \textit{what time tells us?} To this end, we first introduce a Time-Oriented Collection (TOC) dataset, which contains 130,906 images with reliable timestamps. Leveraging this dataset, we propose a Time-Image Contrastive Learning (TICL) approach to jointly model timestamps and related visual representations through cross-modal contrastive learning. 
We found that the proposed TICL, 1) not only achieves state-of-the-art performance on the timestamp estimation task, over various benchmark metrics, 2) but also, interestingly, though only seeing static images, the time-aware embeddings learned from TICL show strong capability in several time-aware downstream tasks such as time-based image retrieval, video scene classification, and time-aware image editing. Our findings suggest that time-related visual cues can be learned from static images and are beneficial for various vision tasks, laying a foundation for future research on understanding time-related visual context. 

\end{abstract}

%% file: sec/1_intro.tex
\epigraph{``Time is the moving image of eternity.''}{\textit{Plato}}

\section{Introduction}
\label{sec:intro}

On our planet, the day-night cycle occurs every 24 hours, a phenomenon recorded systematically by various clock systems developed by human society. Surprisingly, such clock systems emerged much earlier than our recognition of Earth as a ``blue marble'' engaged in constant orbital movement \citep{dohrn1996history}. Although most people possess a vague, intuitive sense of current time \citep{MOORE1992101}, the origin of this metaphysical consciousness of time, which is a key concept for both our bodies and society, remains elusive. Research in neuroscience has revealed that visual stimuli from photoreceptors are crucial for the adaptation of mammals to day-night rhythms \citep{DUFFY2009165}. This implies that the concept of time for humankind could emerge from various visual experiences. Given the implicit relations between clock time and visual experiences, we are interested in asking: \CameraReady{\textit{1) Can neural networks gain similar awareness to clock time from 
solely visual stimuli \ie static images? 2) If so, what implications does such time awareness tell us towards understanding the world?}}

To answer these questions, in this study, we propose an approach to learn and disentangle the visual cues related to time from static images, via a pre-text task estimating the clock timestamps from images, and exploration on various downstream tasks to find their visual implications. 

\begin{figure*}
    \centering
    \begin{subfigure}[b]{0.29\linewidth}
        \centering
        \includegraphics[width=\linewidth]{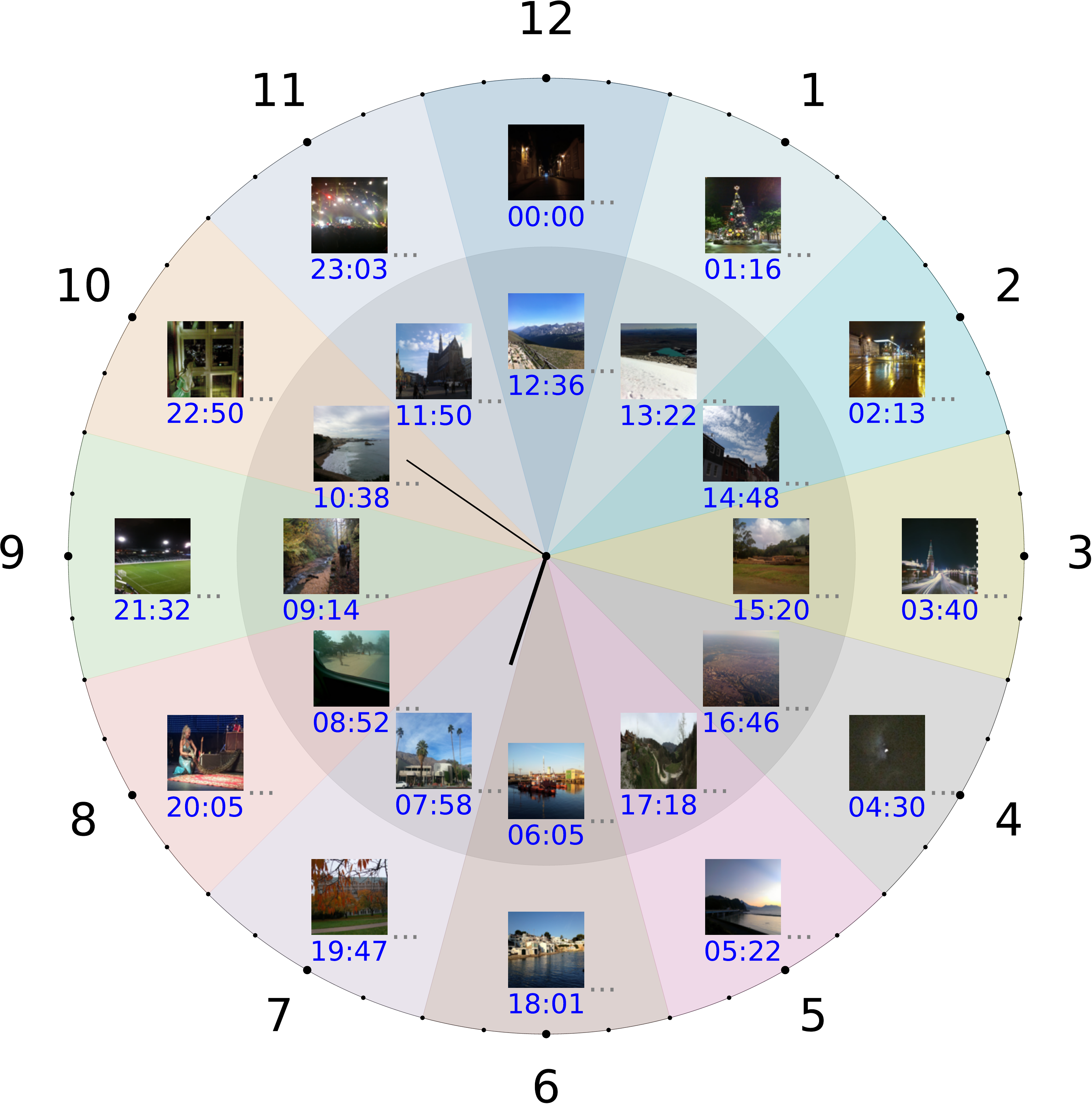}
        \caption{TOC dataset}
        \label{fig:dataset_overview}
    \end{subfigure}
    \hfill
    \begin{subfigure}[b]{0.32\linewidth}
        \centering
        \includegraphics[width=\linewidth]{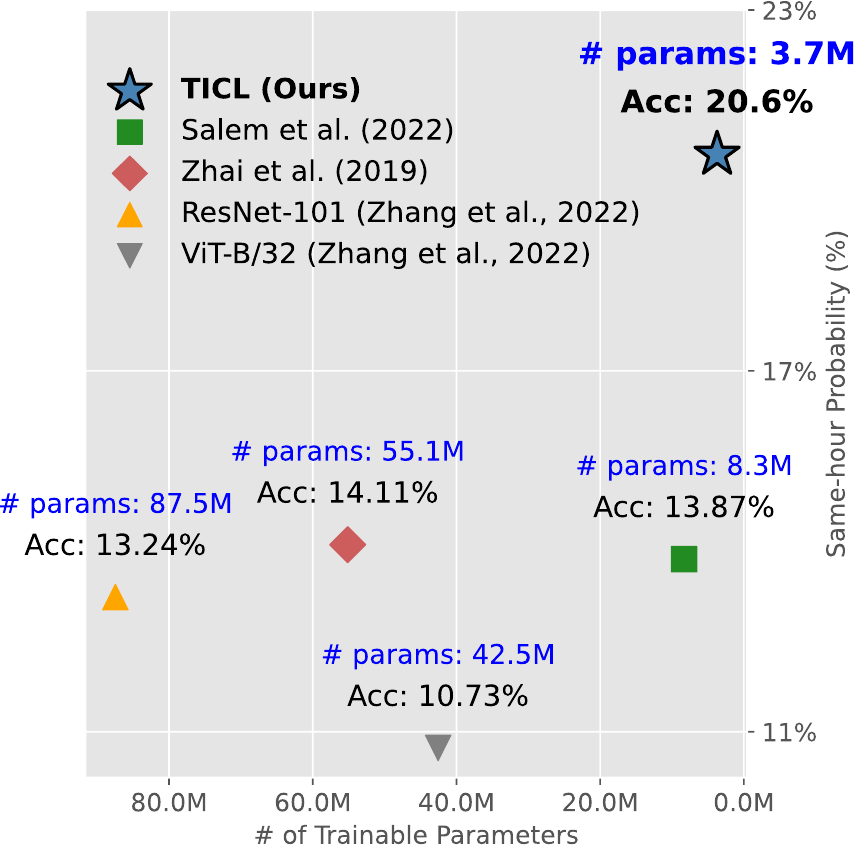}
        \caption{Time-estimation performance}
        \label{fig:TE_performance}
    \end{subfigure}
    \hfill
    \begin{subfigure}[b]{0.37\linewidth}
        \centering
        \includegraphics[width=\linewidth]{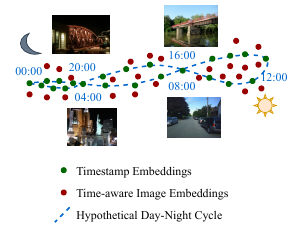}
        \caption{{Expected time aware embedding space}}
        \label{fig:Problem_FORM}
    \end{subfigure}
    \hfill
    \caption{\textbf{An overview of our study,} in which we presented a new high-quality dataset for time-of-day estimation (a), based on which we propose a new approach, achieving state-of-the-art performance (b). {We further explore the implications of learned time-aware embeddings (c)}, showing effectiveness over several time-related downstream tasks.}
    \label{fig:Contribution}
\end{figure*}

Learning to model captured timestamps of images requires a reliable natural image dataset with timestamps. There are previous surveillance camera datasets with fixed views, such as the Time of Year Dataset (TYD) \citep{CVL_CAMS} and other subsets of the Archive of Many Outdoor Scenes (AMOS) \citep{jacobs09webcamgis}, featuring images captured by a few stationary webcams at different times of the day. However,  these datasets do not reflect the complexity and diversity of views in real-world applications. To address this issue, \citet{salem2020dynamic} proposed a mixed subset of AMOS and YFCC100M \citep{thomee2016yfcc100m}, containing diverse samples. However, many images in this dataset suffer from incorrect timestamps due to unsynchronised time zones \citep{padilha2022content}, which undermines its reliability for learning robust time-awareness.

In addition to the challenge of lacking reliable datasets, designing effective solutions for the pre-text task also faces significant difficulty. Providing accurate clock time estimates requires the model to go beyond understanding basic illumination, as the task is complicated by inherent ambiguities between the clock timestamp and images. These ambiguities arise because daylight time is influenced by additional metadata, such as regional climate and seasonal variations that affect the duration of daylight hours \citep{CVL_CAMS, Sharma2016, ZhangAIPR}.
To cope with this issue, \citet{salem2022timestamp,zhai2019learning} introduced additional metadata inputs, aiming to model the joint conditional probabilities between geolocation, hour and date to provide performance improvements to the estimation task. While these works made reasonable and valuable improvements, they have introduced extra dependencies on additional metadata, limiting the generalisation ability when such metadata is unavailable as reported. On the other hand, they primarily focus on the specific task of clock time estimation, without exploring further implications of time to other applications. Whereas in this work, in addition to estimating more accurate timestamps, we further utilise the learned time and time-aware image features to investigate their impact on several other downstream tasks.

Specifically, due to the lack of high-quality data, we first curate a new benchmark dataset comprising social media images featuring diverse views and objects, along with manually verified reliable timestamps. Such a dataset has the potential to become the new de facto choice for future research. {Secondly, we propose a Time-Image Contrastive Learning (TICL) approach that extracts time-of-day awareness from rich semantics from foundation vision-language model via contrastive learning} outperforms all existing methods on the pre-text timestamp estimation task. Moreover, we conduct explorations of utilising such time-awareness on several downstream tasks, including time-based image retrieval, video scene recognition, and time-aware visual editing, showing the indirect relations between time and scene understanding. 

Note that this work is not aiming at purely time estimation, but more about an exploration of what the learned embedding tells us, through such a pre-text task. Our key contributions can be summarised as follows: 

\begin{itemize}
    \item We introduce Time-Oriented Collection (TOC), a new benchmark dataset containing 130,906 images with reliable timestamps (examples shown in \cref{fig:dataset_overview}).
    \item We propose TICL, an approach jointly modelling time and related visual representations, achieving state-of-the-art (SOTA) performance on timestamps estimation from static images. \cref{fig:TE_performance} shows the achieved performance, boosting SOTA from \textbf{14.11\%} to \textbf{20.6\%}, while keeping small number of trainable parameters.

    \item {We study the potential of the learned time-aware visual embeddings (\cref{fig:Problem_FORM}) by validating them on several downstream tasks (\eg time-based image retrieval, video scene classification, and time-aware image editing), showing clear evidence of their effectiveness.}  
\end{itemize}

%% file: sec/2_related_works.tex
\section{Related Works}
\label{sec:related_works}
\subsection{Image datasets with timestamps}
\label{sec:time_data}

Estimating the time of day from static images is a notable and underexplored challenge. Earlier studies were hampered by the scarcity of datasets with images paired with accurate local timestamps. Many images from social networks often have metadata that is inaccurate, missing, or uncalibrated to local timezones. To cope with this, some researchers have turned to webcam image datasets, which naturally include accurate timestamps. However, these datasets are limited to fixed views and are often degraded by noise, low light, or obstructions, hindering their generalisation to diverse applications.
 
For example, established social media image datasets such as MIRFLICKR-1M \citep{MIRFLICKR1M} and YFCC100M \citep{thomee2016yfcc100m} were found to contain many unnatural non-photographic images (\eg memes, scribbles) and inaccurate timestamps due to unsynchronised clocks and other sources of inconsistency. On the other hand, webcam datasets contain only fixed stationary views, such as AMOS \citep{jacobs07amos} and TYD dataset \citep{CVL_CAMS}, which fail to represent the complexities of temporal variations within diverse environments. The CVT-Time dataset \citep{salem2020dynamic}, despite combining stationary webcam images with YFCC100M subsets with images captured by smartphone, still struggles with unreliable timestamps and low-quality webcam images.

\subsection{Clock timestamp estimation}
\label{sec:time_esti}
Previous \Update{work} have studied joint attribute estimation of images, including captured clock time, date, and geolocation. {\textit{In this work, we focus solely on clock time estimation regardless of other fields in timestamps (\Update{e.g., date, year}),} since we are primarily interested in relations between clock time itself originated from human activities \citep{MOORE1992101} to visual cues.} \citet{CVL_CAMS} used VGG-16 to classify temperature, month, and hour intervals from images taken by 6 webcams during daylight, which is insufficient for comprehensive day-long analysis. In addition to such earlier simple approaches, \citet{zhai2019learning} worked with a mixed dataset of Flickr and webcam images, classifying images taken at the same hour but in different months into 288 classes, optionally incorporating geolocation inputs. Similarly, \citet{salem2022timestamp} used webcam images, predicting month, week, and hour as dependent tasks trained jointly while considering geolocations as optional inputs. Such joint predictions improve hour-based time-of-day classification by leveraging metadata cues on regions and climate, which correlate with daylight length. However, such dependency also puts risks on generalization ability when there are no reliable geolocation or date metadata available \citep{salem2020dynamic, zhai2019learning}. {Therefore, we deliberately chose to use only input images for clock time prediction, without utilising any additional metadata with regard to the generalization problem acknowledged in previous \Update{work}. }

%% file: sec/3_methodology.tex
\section{Time-Image Contrastive Learning}
\label{sec:TICL_method}

Before introducing the proposed method, we revisit the problem formulation \Update{and goal of our study. Unlike previous timestamp-prediction architectures that simply attach a classifier to an image backbone \citep{salem2020dynamic, zhai2019learning, CVL_CAMS}, our goal is \emph{not} to engineer another bespoke head. Instead, we introduce a minimal, generic recipe that transforms a certain environmental attribute (in our case, the clock time) into \emph{embeddings} that elucidates the implications of world understanding from the attribute. The novelty of TICL therefore lies in providing explicit semantic-aware time embeddings for further investigations with its robustness on time encoding validated by superior pre-text time estimation task performance.

For the clock-time estimation of images specifically,} we seek to train a model $f_{\theta}(\cdot)$, which predicts a timestamp $\hat{t}$ given input image $x$. The estimate can be written as $\hat{t}=f_{\theta}(x)$. While regression seems ideal owing to the continuous nature of time, it faces significant challenges. The cyclic nature of the clock introduces discontinuity to regression methods that treat target values as scalars within a range that is a disconnected set \citep{Zhou_2019_CVPR}. In regression, cyclic data often causes $\hat{t}$ to cluster near the midpoint of the range \citep{vamplew}. For instance, timestamps like 23:59 and 00:00, despite their visual similarity, are treated as opposite extremes on the time scale. In such cases, the regression model tends to reach a sub-optimal solution around 12:00, which is far from accurate. Apart from this extremal case, the sensitivity of the scalar time values also encourages the model to predict an average ground-truth value across visually similar images. Encoding time into cyclic space partially mitigates scalar discontinuities \citep{vamplew,kazemi2019time2veclearningvectorrepresentation}, but sensitivity issues still limit performance (see detailed analysis in \cref{sec:precision_limits}). 

This justified why prior studies have employed classification over discrete time periods (\eg hours), in which $\hat{t}$ has finite value options corresponding to classes. Classification mitigates the above issue in regression by simplifying the model to give a coarser estimate from mutually orthogonal discrete classes. Even for boundary cases like 23:59 and 00:00, the classification model tends to predict one of the adjacent classes (\eg 23:00 or 00:00), which is more reasonable. However, the orthogonality of one-hot vectors \citep{rodriguez2018beyond} overlooks the relationships (partial order, cyclic) between time periods.

On top of these observations, we propose \emph{Time-Image Contrastive Learning} (TICL), a multi-modal approach that jointly learns time and image representations via cross-modal contrastive learning, inspired by GeoCLIP \citep{geoclip}. Each input image $x_i$ is associated with a label $t_i\!\in\!\mathbb{R}^{C}$ indicating its time period. Empirically, we fix $C\!=\!24$ for all the results in the main paper for a fair comparison with previous \Update{work} (see further discussions on the choice of $C$ in \cref{sec:class_num}). Each one-hot vector $t_i$ is projected into a high-dimensional representation space $\mathbb{R}^K$ using a \emph{Time Encoder} $T_i=f_{\theta_T}(\cdot)$, where $K\!=\!768$ to match the dimensionality of the image representation. 

\begin{figure*}[h!]
    \centering
    \includegraphics[width=0.8\linewidth]{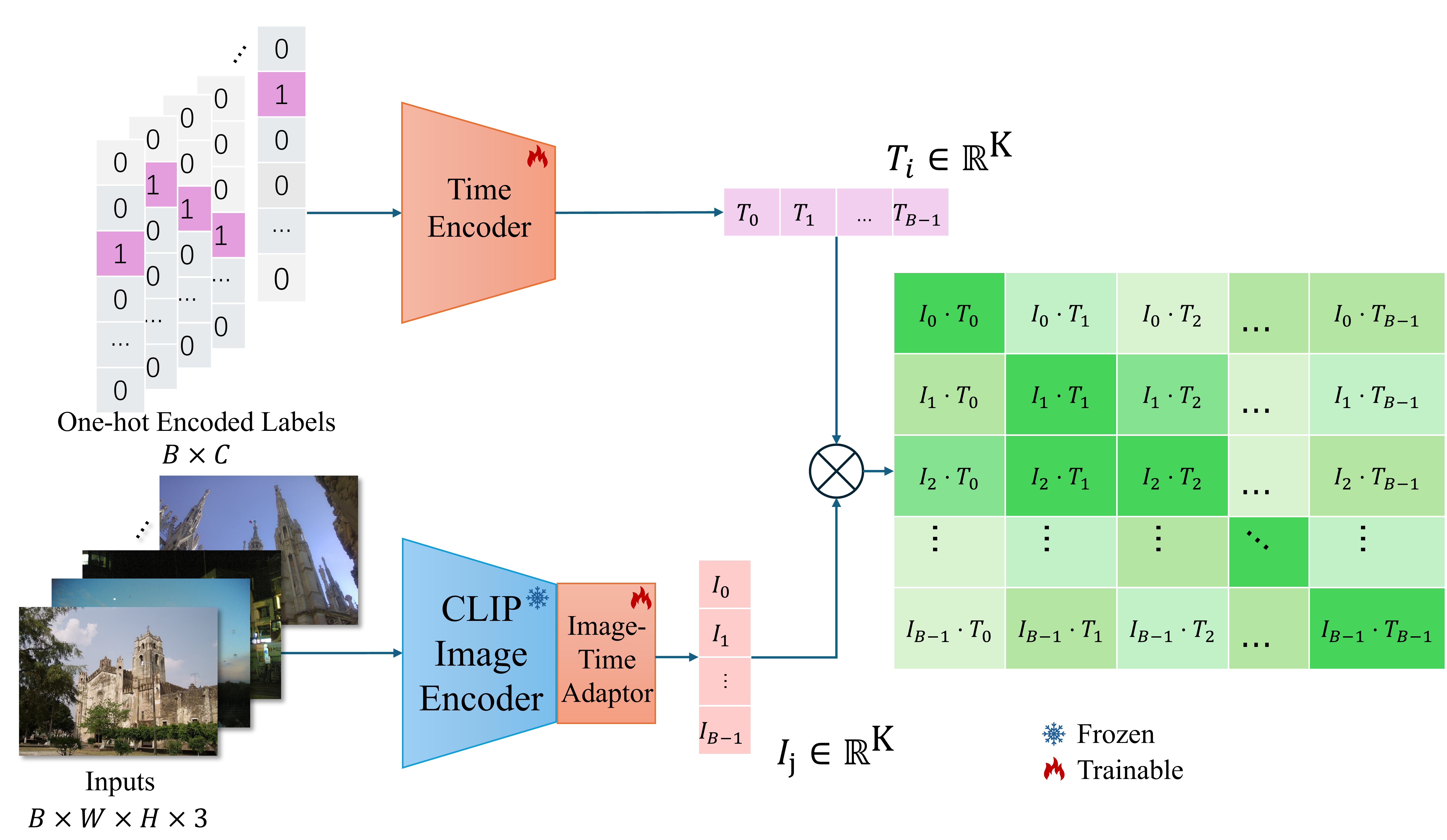}
    \caption{\textbf{Overview of TICL.} Given static images and one-hot time labels, two encoders (Time Encoder and image encoder + ITA) project inputs into a shared feature space; a contrastive loss aligns the corresponding pairs.}
    \label{fig:model_arch}
\end{figure*}

As visualised in \cref{fig:model_arch}, during training we maximise the cosine similarity between the CLIP image feature $I_i=f_{\theta_I}(x_i)$ and its corresponding time-class embedding $T_i=f_{\theta_T}(t_i)$. Here, $f_{\theta_I}(\cdot)$ denotes the combined operation of the frozen CLIP image encoder and an \emph{Image-Time Adaptor} (ITA). The alignment is optimised by minimising the contrastive loss \citep{he2019moco}, as defined in \cref{eq:loss}, where $\tau$ is a learnable temperature that controls the sharpness of the softmax distribution \citep{wu2018unsupervised}. At inference, TICL flexibly supports both classification at any class granularity and nearest-neighbour inference (see \Update{\cref{sec:model_details}}, \cref{sec:class_num} and \Update{\cref{sec:inference}}).

\begin{equation}
    \label{eq:loss}
    \mathcal{L}_{B}
    =-\sum_{i=0}^{B-1} 
    \log\!
    \frac{\exp\!\left(I_i\!\cdot T_i/\tau\right)}
         {\sum_{j=0}^{B-1}\exp\!\left(I_i\!\cdot T_j/\tau\right)}
\end{equation}

Several key intuitions support this design. Previous work has shown that combining additional geolocation and date information can improve the performance of time estimation, but reliance on such attributes propagates errors from prior to posterior predictions \citep{salem2020dynamic}. We observed that the CLIP image encoder is a powerful foundation model capturing rich semantic context, exhibiting strong zero-shot capabilities in geo-localisation and scene recognition \citep{radford2021learningtransferablevisualmodels,agarwal2021evaluatingclipcharacterizationbroader,geoclip}. These results suggest that CLIP implicitly encodes cues (\eg climate, region) relevant to timestamp estimation. Therefore, we \emph{freeze} the CLIP backbone and extract these cues directly, rather than ingesting raw geolocations or season inputs that may themselves be noisy \citep{salem2022timestamp}.

Another benefit comes from the learnable time embedding in the contrastive scheme. In a vanilla classifier, the final output $\hat{y}$ is confined to the simplex $\bigl\{\|\hat{y}\|_1=1,\hat{y}\in\mathbb{R}^C\bigr\}$, where each target is a fixed one-hot vector. Samples are optimised only toward their own target, and activations to related classes may be suppressed \citep{5128907}. In contrast, our method endows each time class with a \emph{trainable} vector that absorbs shared information among temporally adjacent samples, aligning timestamps and visual inputs more effectively—especially for tail classes. These learnable embeddings also prove useful in downstream tasks (\cref{sec:downstream_tasks}). 

\Update{In summary, we expect TICL combines the benefit from the orthogonality of one-hot encoded labels and flexibility of learnable high-dimensional embeddings. This simple yet principled design delivers consistent gains over prior estimators and alternative time-encoding schemes \citep{NIPS2007_013a006f,kazemi2019time2veclearningvectorrepresentation,salem2022timestamp,zhai2019learning} (see \cref{sec:te_perf,sec:ablation}), while remaining computationally lightweight and conceptually generalisable.}

%% file: sec/4_dataset.tex
\section{Benchmark Dataset TOC}

\label{sec:dataset}
\begin{table}[t]
\caption{Classification accuracy on the TOC test set of the baseline model \citet{salem2022timestamp} when using different training datasets.}
\label{tab:dataset_ablate}
\centering
\resizebox{0.55\linewidth}{!}{%
\begin{threeparttable}
\begin{tabular}{l|cccc}
\toprule
\textbf{Training Dataset} \tnote{$\dagger$} & \textbf{Top-1} {$\uparrow$} & \textbf{Top-3} {$\uparrow$} & \textbf{Top-5} {$\uparrow$} \\     
\midrule
Original \citet{salem2020dynamic}                & 12.02\%         & 34.05\%         &  56.45\%       \\

\textbf{\Update{Cleaned TOC (ours)\tnote{$\ddagger$}}}                        & \textbf{13.87\%}    & \textbf{39.36\%}    & \textbf{60.71\%}   \\
\bottomrule
\end{tabular}
\begin{tablenotes}
\footnotesize 

\item[$\dagger$] In training datasets, overlapped samples from the test set are excluded.
\item[$\ddagger$] \Update{ The new dataset details are introduced in \cref{sec:dataset} and \cref{sec:dataset_details}}
\end{tablenotes}
\end{threeparttable}
}
\end{table}

With regards to problems \cref{sec:time_data}, we introduce a new benchmark Time-Oriented Collection (TOC) dataset consisting of high-quality images sourced from social media, featuring reliable image metadata. We collected 117,815 training samples and 13,091 test samples from the Cross-View Time (CVT) \citep{salem2020dynamic}, mitigating various limitations in previous datasets. This dataset reflects real-world scenarios and human activities, making time-of-day estimation more applicable to potential practical applications. 

\begin{figure*}[h!]
    \centering
    \includegraphics[width=0.8\linewidth]{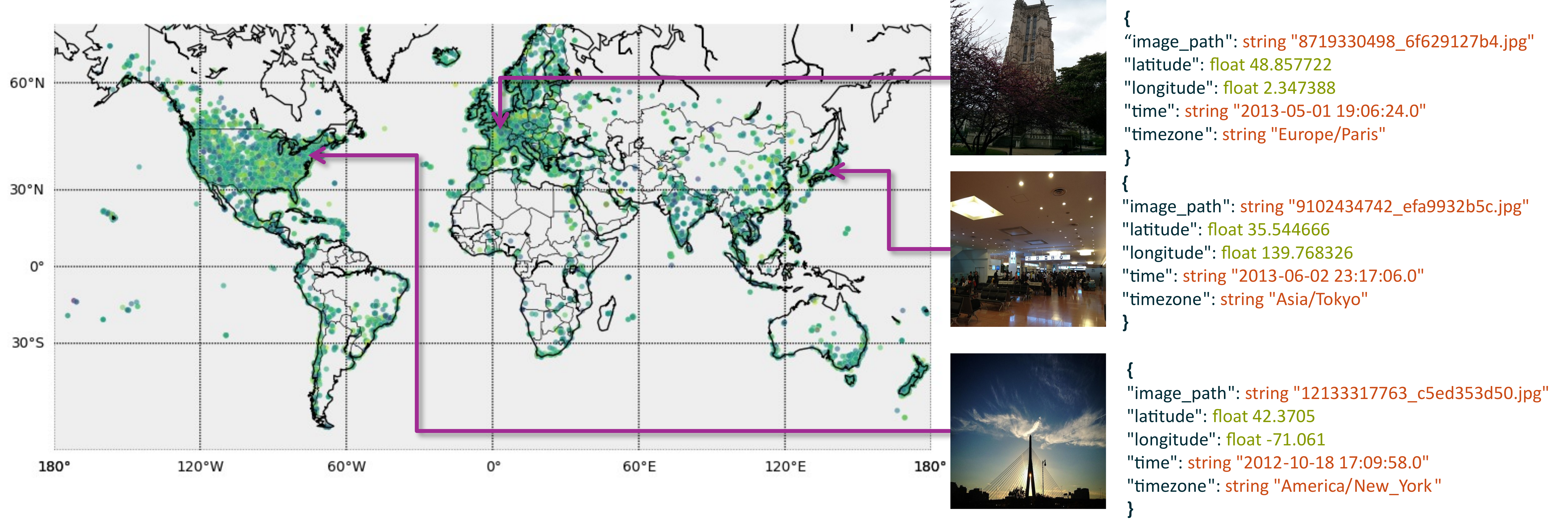}
    \caption{\textbf{Sample images and metadata from the TOC dataset \wrt GPS coordinates.} Metadata contains several fields indicating timestamps and geolocations. The samples spread across all the continents and show a natural distribution of internet images, where the southern hemisphere has relatively fewer samples due to a sparser population of photo capturing.}
    \label{fig:sample-TOC}
\end{figure*}

During dataset curation, we manually filtered out unnatural, non-photographic images from the CVT dataset and calibrated the timestamps to match the images. To accelerate the process, we utilized ResNet18 features of the images to quickly identify the outliers in deep image feature space for different periods of the day. After which, we conducted meticulous manual inspection for each outlier image to check if their timestamp or contents are natural and valid (see more details in \cref{sec:dataset_details}). This revised dataset reflects natural variations in human activity throughout the day, with improved reliability in terms of time metadata. As evidence, \cref{tab:dataset_ablate} shows a performance gap on the same test set using different levels of filtering on the CVT dataset, justifying the effectiveness of the filtering process indicated by improvements in baseline model performance, suggesting that repetitive surveillance-camera-sourced and unnatural non-photographic images we removed do not help the model in better time recognition for images in the wild. A few examples of the exact format and appearances of the remaining samples within the TOC dataset are provided in \cref{fig:sample-TOC}. Geolocation distribution of the sample images within our final TOC dataset in \cref{fig:sample-TOC} also suggests that, due to the inherent geographic distribution of the internet, the northern hemisphere has more data captured by nature. Our dataset well represents such natural distribution. Further information and statistics about the dataset are available in \cref{sec:dataset_details}.

%% file: sec/5_experiment.tex
\section{Experiments}
\label{sec:experiments}
\subsection{Experiment Setting}

\paragraph{Dataset and metrics:} We use different evaluation metrics to measure performance on image clock time estimation tasks: top-k classification accuracy with \(k = 1, 3, 5\), and Time Mean Absolute Error (MAE) on a minute basis. In addition to the TOC test set, to better evaluate the generalisation ability of the proposed method, we selected a subset of the AMOS dataset \citep{jacobs07amos} as an additional test set. This additional test set contains 3,556 images with high SNR (which ensures good sample quality) captured by 53 stationary surveillance cameras with a more balanced time label distribution (see curation process and statistics in \cref{sec:AMOS_detail}). That is, all the compared models are trained solely on the TOC training set and evaluated on different test sets to demonstrate generalisation ability across different domains. 

\paragraph{Implementation details:} For our proposed TICL, we use Adam optimiser with an initial learning rate of $5\times10^{-4}$ and a weight decay of $1\times10^{-6}$. The training process spans 20 epochs, with the learning rate halved every 2 epochs and a batch size of 512. The temperature parameter is initialized to 0.07. All input images are resized to \(224 \times 224 \). For a fair comparison, we retrained all the previous baseline methods on the cleaned TOC train set, using the best training configurations reported in \citet{ZhangAIPR, zhai2019learning, salem2022timestamp} respectively. Additional details about implementations are available in \cref{sec:model_details}.

\subsection{Time estimation performance}
\label{sec:te_perf}

As shown in \cref{tab:time_prediction}, TICL not only outperforms all previous pure vision methods but also outperforms previous methods that require additional geolocation inputs on most metrics. TICL also demonstrates better performance in the additional AMOS test set, thereby indicating better generalisation ability. 
In summary, our experimental results indicate an overall improvement of the proposed methods in clock time estimation, especially in terms of accuracy and generalisation ability.
\vspace{-3mm}

\paragraph{Additional error analysis on pre-text tasks}

\begin{figure*}[h]
    \centering
    \includegraphics[width=0.8\linewidth]{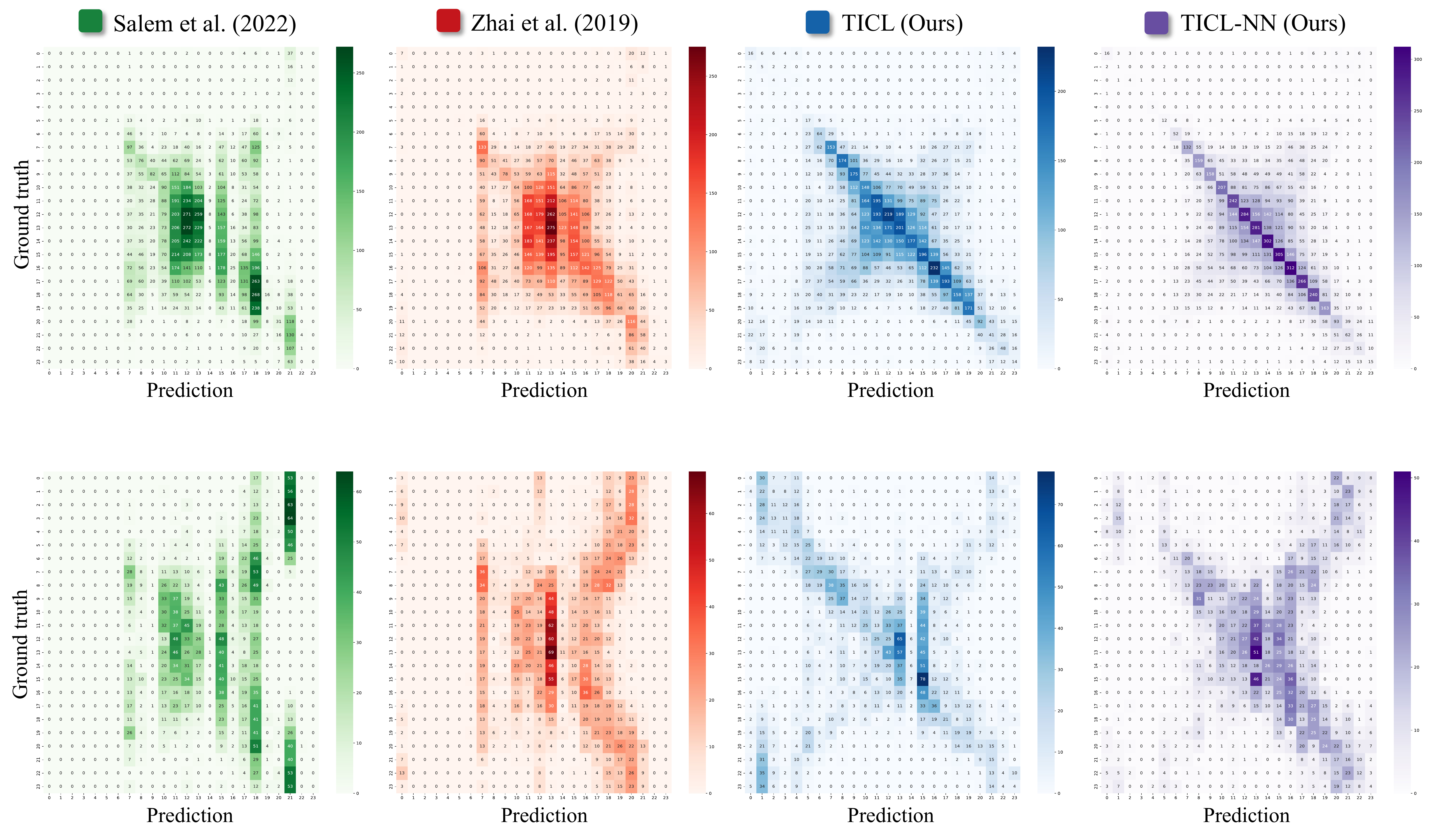}
    \caption{\textbf{Confusion matrices.} They provide more detailed comparisons throughout the 24 hours on our TOC test set (top), and the AMOS test set (bottom).}
    \label{fig:conf_mat}
\end{figure*}

In addition to the quantitative results, we also visualised the confusion matrices in \cref{fig:conf_mat} to provide a more in-depth evaluation of the task. An interesting finding is that both \citet{salem2022timestamp} and \citet{zhai2019learning} overlooked minority classes in the training set (classes from 1 a.m. to 5 a.m.), resulting in nearly no predictions for these classes on both test sets. This indicates a notable bias in these models towards classes during hours of intense human activity, when more images are present in dataset. In contrast, our proposed TICL method exhibits more balanced class-wise distributions of positive predictions on both test sets, suggesting better estimation fairness. The general trend in all the confusion matrices also suggests the remaining challenges faced by all methods. Notable anti-diagonal patterns indicate inherent visual ambiguities of the clock system \wrt appearances. 

\begin{table*}[t]
\centering
\caption{Time estimation performance on our TOC dataset and the AMOS test set.}
\label{tab:time_prediction}
\resizebox{\textwidth}{!}{%
\begin{threeparttable}
\begin{tabular}{l|cccc|cccc}
\toprule
                           & \multicolumn{4}{c|}{\textbf{TOC test set}}           & \multicolumn{4}{c}{\textbf{AMOS test set}\tnote{$\dagger$}}            \\
                           & Top-1 acc {$\uparrow$} & Top-3 acc {$\uparrow$} & Top-5 acc {$\uparrow$} & Time MAE (min.) {$\downarrow$} & Top-1 acc & Top-3 acc & Top-5 acc & Time MAE (min.) \\
\midrule

\Update{BLIP-VQA-base \citep{BLIP} \tnote{$\mathparagraph$}} & \Update{9.36\%} & \Update{-} & \Update{-} & \Update{241.58} & \Update{7.28\%} & \Update{-} & \Update{-} & \Update{302.82} \\
\Update{GPT-4o-mini \citep{openai2024gpt4ocard} \tnote{$\mathparagraph$}} & \Update{15.35\%} & \Update{-} & \Update{-} & \Update{\underline{161.39}} & \Update{11.15\%} & \Update{-} & \Update{-} & \Update{\underline{216.83}} \\
\midrule
\citet{ZhangAIPR} (ResNet-101)  & 13.24\%      & 37.30\%      & 58.23\%      & 177.84
      & 7.85\%       & 24.26\%      & 40.10\%      & 261.89      \\
\citet{ZhangAIPR} (ViT-B/32)        & 10.73\%      & 31.21\%      & 49.05\%      & 195.33      & 7.25\%       & 21.03\%      & 32.93\%      & 263.87      \\
\citet{zhai2019learning}             & 14.11\%      & 40.47\%      & 65.94\%      & 188.78      & 9.14\%       & 27.95\%      & 45.36\%      &  262.68  \\
\citet{salem2022timestamp}           & 13.87\%      & 39.36\%      & 60.71\%      &  186.44     & 8.63\%       & 26.49\%      & 42.58\%      & 255.20      \\
\textbf{TICL (Ours)}                   & \underline{20.60\%}      & \underline{49.01\%} & \textbf{67.82\%}      & {171.65} & \textbf{13.55\%} & \textbf{38.49\%} & \textbf{57.28\%} & \textbf{187.87} \\
\textbf{TICL-Nearest-Neighbour (Ours)}\tnote{$\ddagger$} & \textbf{25.67\%} & \textbf{49.32\%}     & \underline{66.74\%}      & \textbf{156.24}      & \underline{11.14\%}      & \underline{31.01\%}      & \underline{48.84\%}      & {220.94}      \\
\midrule
\citet{zhai2019learning}\tnote{$\mathsection$} & 15.01\%      & 42.54\%      & 68.24\% & 185.34      & 8.85\%       & 24.12\%      & 38.63\%      & 268.41      \\
\citet{salem2022timestamp}\tnote{$\mathsection$} & 13.53\%      & 38.47\%      & 59.10\%      & 176.70      & 8.16\%       & 23.88\%      & 39.67\%      & 257.00      \\
\bottomrule
\end{tabular}%
\begin{tablenotes}
\footnotesize
\item[$\dagger$] Experiments on this test set are conducted in a zero-shot manner, in which we directly evaluate models trained solely on the TOC dataset.
\Update{\item[$\mathparagraph$] The result is obtained via direct Visual Q\&A on the Visual Language Model with details in \cref{sec:inference}.}
\item[$\ddagger$] Results in this row are achieved via Nearest-Neighbour style inference. We choose the time labels of nearest neighbours from the train dataset as estimations\Update{~(see details in \cref{sec:inference}).}
\item[$\mathsection$] These methods take additional known geolocation metadata inputs. Therefore, it's unfair to directly compare them with other methods. So we put them here just for reference.
\end{tablenotes}
\end{threeparttable}
}

\end{table*}
\subsection{\Update{Detailed component analysis}}
\label{sec:ablation}
\begin{table*}[t]
\caption{\Update{Detailed component analysis} of the proposed method design.}
\label{tab:ablation_image}
\centering
\resizebox{\textwidth}{!}{%
\begin{threeparttable}
\begin{tabular}{l|c|c|cccc|cccc}
\toprule
\textbf{Image Encoder}\tnote{$\dagger$}  & \textbf{$f_{\theta_T}$}\tnote{$\ddagger$} & \textbf{$f_{\theta_{\text{ITA}}}$}\tnote{$\mathsection$} & \multicolumn{4}{c|}{\textbf{TOC test set}}           & \multicolumn{4}{c}{\textbf{AMOS test set}}            \\
                           &  &  & Top-1 acc $\uparrow$ & Top-3 acc $\uparrow$ & Top-5 acc $\uparrow$ & Time MAE (min.) $\downarrow$ & Top-1 acc & Top-3 acc & Top-5 acc & Time MAE (min.) \\
\midrule

\multirow{4}{*}{DINOv2-base}    & \textcolor{BrickRed}{\textbf{\ding{55}}} & \textcolor{BrickRed}{\textbf{\ding{55}}}\tnote{$\mathparagraph$} & 7.69\% & 23.36\% & 38.61\% & 302.84 & 5.65\% & 17.12\% & 27.28\% & 319.09 \\  
                                & \textcolor{ForestGreen}{\textbf{Ours}} & \textcolor{BrickRed}{\textbf{\ding{55}}} & 8.01\% & 23.84\% & 39.06\% & 295.34 & 5.23\% & 17.35\% & 29.22\% & 320.76 \\     
                                & \textcolor{BrickRed}{\textbf{\ding{55}}} & \textcolor{ForestGreen}{\textbf{\ding{51}}} & 1.02\% & 3.29\% & 12.04\% & 486.77 & 4.11\% & 11.41\% & 19.62\% & 381.92 \\
                                & \textcolor{ForestGreen}{\textbf{Ours}} & \textcolor{ForestGreen}{\textbf{\ding{51}}} & 9.53\% & 27.34\% & 44.17\% & 254.49 & 5.09\% & 14.74\% & 25.16\% & 327.72 \\ \midrule
\multirow{4}{*}{SwinV2(B)}        & \textcolor{BrickRed}{\textbf{\ding{55}}} & \textcolor{BrickRed}{\textbf{\ding{55}}}\tnote{$\mathparagraph$} & 11.45\% & 32.27\% & 51.08\% & 240.77 & 7.87\% & 22.49\% & 36.81\% & 281.80 \\
                                  & \textcolor{ForestGreen}{\textbf{Ours}} & \textcolor{BrickRed}{\textbf{\ding{55}}} & 11.64\% & 32.13\% & 50.33\% & 243.86 & 7.51\% & 22.36\% & 37.54\% & 288.21 \\
                                  & \textcolor{BrickRed}{\textbf{\ding{55}}} & \textcolor{ForestGreen}{\textbf{\ding{51}}} & 12.74\% & 33.65\% & 52.06\% & 222.76 & 6.75\% & 23.76\% & 38.41\% & 284.30 \\
                                  & \textcolor{ForestGreen}{\textbf{Ours}} & \textcolor{ForestGreen}{\textbf{\ding{51}}} & 13.37\% & 34.94\% & 52.93\% & 216.17 &  7.37\% & 22.98\% & 38.08\% & 276.66\\ \midrule
\multirow{4}{*}{ConvNeXt(L)}   & \textcolor{BrickRed}{\textbf{\ding{55}}} & \textcolor{BrickRed}{\textbf{\ding{55}}}\tnote{$\mathparagraph$}  & 11.59\% & 32.93\% & 50.88\% & 240.64 & 6.41\% & 21.68\% &  37.63\%& 300.74 \\  
                                & \textcolor{ForestGreen}{\textbf{Ours}} & \textcolor{BrickRed}{\textbf{\ding{55}}}  & 11.86\% & 32.81\% & 50.18\% & 240.80 & 6.10\% & 20.66\% &  35.85\%& 302.45 \\
                                  & \textcolor{BrickRed}{\textbf{\ding{55}}} & \textcolor{ForestGreen}{\textbf{\ding{51}}} & 13.51\% & 35.29\% & 52.76\% & 216.28 & 7.71\% & 24.33\% &  39.96\%& 275.23 \\
                                  & \textcolor{ForestGreen}{\textbf{Ours}} & \textcolor{ForestGreen}{\textbf{\ding{51}}} & 14.67\% & 36.75\% & 54.60\% &  204.19 & 8.27\% & 24.78\% & 40.86\% & 263.03\\ \midrule

\multirow{6}{*}{\textbf{CLIP (ViT-L/14)}} & \textcolor{BrickRed}{\textbf{\ding{55}}} & \textcolor{BrickRed}{\textbf{\ding{55}}}\tnote{$\mathparagraph$}& 16.66\% & 44.43\% & 65.07\% & 193.66 & 12.37\% & 36.95\% & 55.96\% & 200.93 \\   
                                  & \textcolor{ForestGreen}{\textbf{Ours}} & \textcolor{BrickRed}{\textbf{\ding{55}}} & 16.73\% & 44.05\% & 63.99\% & 195.41 & 13.50\% & 38.49\% & \textbf{58.30\%} & 189.99 \\  
                                  & \textcolor{BrickRed}{\textbf{\ding{55}}} & \textcolor{ForestGreen}{\textbf{\ding{51}}} & 18.60\% & 46.41\% & 65.98\% & 181.22 & 12.57\% & 37.51\% & 57.23\% & 189.69\\ 
                                  & \textcolor{BrickRed}{\textbf{\ding{55}}} & \textcolor{ForestGreen}{\(f_{\theta_{\text{ITA}}} \oplus f_{\theta_T}\)} & 19.26\% & 45.40\% & 62.92\% & 189.97 & 11.42\% & 35.65\% & 54.06\% & 197.09\\ 

                                  & \textcolor{ForestGreen}{\textbf{RFF}} & \textcolor{ForestGreen}{\textbf{\ding{51}}} & 16.75\% & 35.14\% & 46.61\% &  206.50 & 6.07\% & 15.78\% & 22.27\% & 290.70\\ 
                                 & \textcolor{ForestGreen}{\textbf{T2V}} & \textcolor{ForestGreen}{\textbf{\ding{51}}} & 17.70\% & 45.69\%  & 66.11\% & 185.89 & 7.37\% & 21.74\% & 35.10\% & 264.25\\ 
                                 & \textcolor{ForestGreen}{\textbf{Ours}} & \textcolor{ForestGreen}{\textbf{\ding{51}}} & \textbf{20.61\%} & \textbf{49.01\%} & \textbf{67.83\%} & \textbf{171.65} & \textbf{13.55\%} & \textbf{38.50\%} & 57.28\% & \textbf{187.87}\\
                                  
\bottomrule
\end{tabular}%
\begin{tablenotes}
\footnotesize
\item[$\dagger$] All image encoders are frozen feature extractors with pretrained features provided by corresponding PyTorch libraries \citep{wolf2020huggingfacestransformersstateoftheartnatural, torchvision2016}.
\item[$\ddagger$] $f_{\theta_T}$ denotes the Time Encoder module. When $f_{\theta_T}$ is absent, only one-hot encoding is used to represent the clock timestamp, and the outputs of $f_{\theta_I}$ need to be projected to 24 dimensions. RFF, T2V means that we uses off-the-shelf encoding methods for low-dimension/cyclic vectors from \citet{NIPS2007_013a006f, kazemi2019time2veclearningvectorrepresentation}.
\item[$\mathsection$] $f_{\theta_{\text{ITA}}}$ denotes the Image-Time Adaptor. When it is absent, only the backbone feature extractor and time encoder are used. 
\item[$\mathparagraph$] The baseline with neither of the $f_{\theta_T}$, $f_{\theta_{\text{ITA}}}$ components simply has a linear layer after Image Encoder projecting the features to 24 dimensions.
\end{tablenotes}
\end{threeparttable}
}
\end{table*}

In this section, we present the ablation study investigating the effectiveness of each module in the proposed TICL model across different configurations. To ensure a fair comparison, we use a classification-based inference pipeline for all experiments (see implementation details in \cref{sec:inference}). \Cref{tab:ablation_image} provides performance comparisons under various settings, including different backbones \citep{ tan2021efficientnetv2smallermodelsfaster, oquab2023dinov2, liu2022swintransformerv2scaling, liu2022convnet} within the image encoders.

\paragraph{\Update{Impact of different backbone image encoders:}} The differences in performance across the image encoder backbones highlight the effectiveness of the CLIP Image Encoder. Thanks to its rich semantic representations, the CLIP Image Encoder consistently achieves better results across all configurations than other backbones. 

\paragraph{Ablation on proposed modules:} We observed that the Time Encoder $f_{\theta_T}$ and the Image-Time Adaptor $f_{\theta_{\text{ITA}}}$ have varying effects when used individually, either slightly improving or degrading the baseline. However, when both modules are employed simultaneously, they lead to universal improvements across all image encoder backbones, underscoring the joint contribution of the Time Encoder and Image-Time Adaptor.

\paragraph{Ablation on different time encoding methods: }  {We also tested the performance using other variants of Time Encoder. RFF \citep{NIPS2007_013a006f} encodes input (hour, minute) into 512-dim vectors to align with ITA outputs directly using the same dynamic queue as in \citet{geoclip}, which is outperformed by our methods on TOC test set and does not generalise well on AMOS test set. In addition, T2V \citep{kazemi2019time2veclearningvectorrepresentation} based Time Encoder also shows similar problems on its performances.} These comparisons suggest that the one-hot class embeddings exhibit better generalisation ability and performance on most metrics. {A possible explanation could be that, the sensitivity of accurate time encoding results in some clock timestamp embeddings being assigned with very limited training samples to represent them. This makes them not robust against the visual ambiguity of time, as images with the same clock time could have very different appearances due to variations in geolocation, season, and climate. In contrast, clock time class embeddings for each hour are vaguely associated with many samples that lie in the same hour interval. Representing target clock timestamp embeddings using spectrums of temporally close samples may reflect the ambiguity of clock time \wrt image appearances, making the estimates more robust and generalizable.} (See more analysis in the \cref{sec:regression_limits}).

\section{What Time Tells Us on Downstream Tasks?}
\label{sec:downstream_tasks}

\Update{Apart from the possible media forensics application that time estimation can be applied to \citep{padilha2022content}. We also interested in} the relation to other computer vision tasks and the capability of the learned time awareness. \Update{Therefore, in this section, we explore 1) time-based image retrieval which is a direct use of the model in retrieval \& recommendation applications (\cref{sec:retrieval}), 2) video scene classification, revealing an interesting connection between static images and dynamic visual scenes learned from timestamp supervision (\cref{sec:vsc}), and 3) time-based image editing, showing the model learned can provide proper perceptual guidance to the generative models (\cref{sec:image_editing})}. 

\subsection{Time-based image retrieval}
\label{sec:retrieval}

\begin{wrapfigure}{r}{0.5\linewidth}
    \centering
    \includegraphics[width=\linewidth]{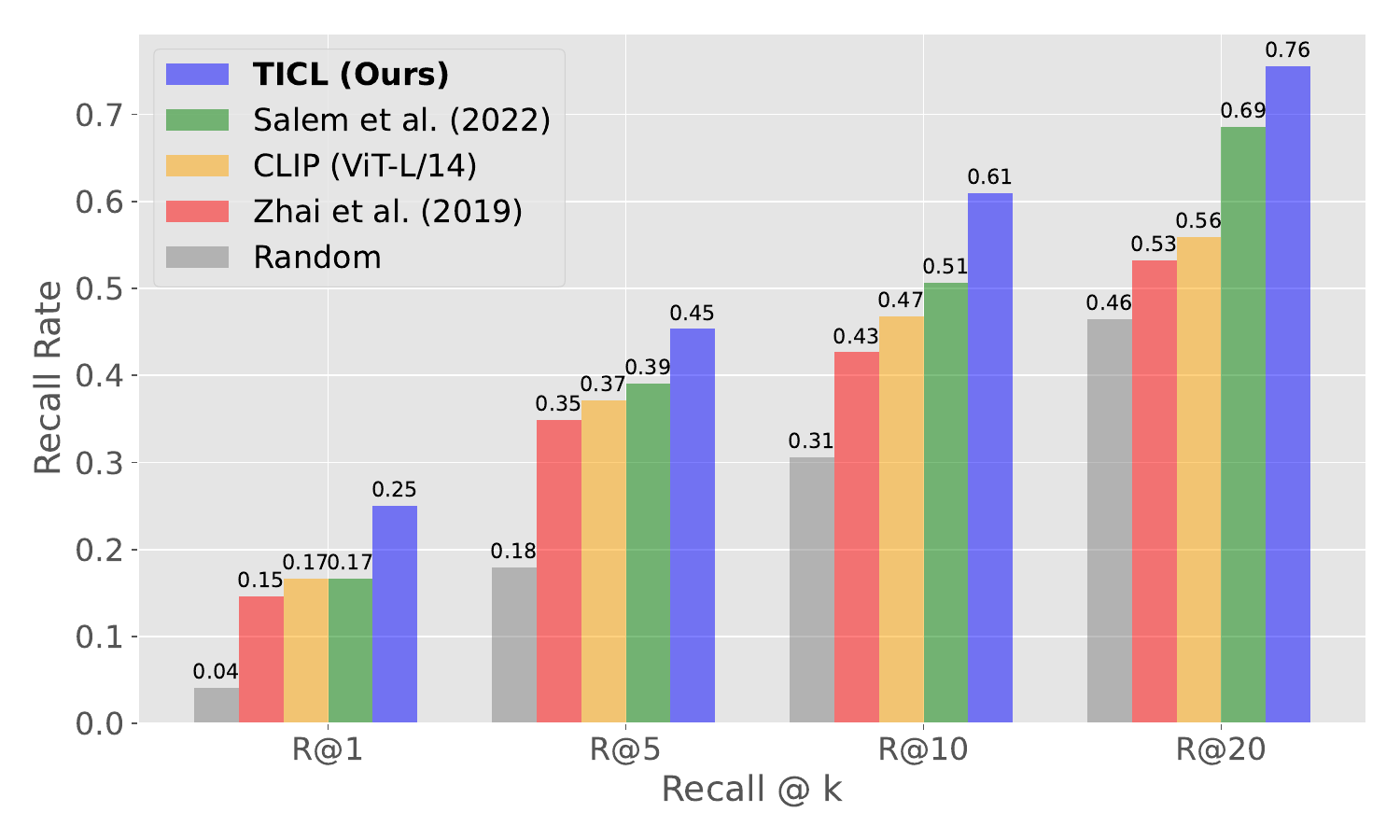}
    \caption{Recall@k for time-based image retrieval.}
    \label{fig:combined_recall_comparison}
\end{wrapfigure}
An intuitive application of the time-aware model is time-based image retrieval. It aims to effectively retrieve images from a database with a similar captured time of day to the query images. We consider a zero-shot vector search engine that retrieves the nearest neighbours of query images based on their time-aware feature similarities. To evaluate this task, we separated the TOC test set into 13,043 database images and 48 query images spanning all 24 hours. The performance is measured using Recall@k reported in \cref{fig:combined_recall_comparison}. Images retrieved with a time difference of no more than $30$ minutes from the queries are considered as positives. The results clearly show that the proposed TICL model achieves the best performance across all Recall@k metrics.

\begin{figure*}[t]
    \centering
    \begin{subfigure}[b]{0.49\linewidth}
        \centering
        \includegraphics[width=\linewidth]{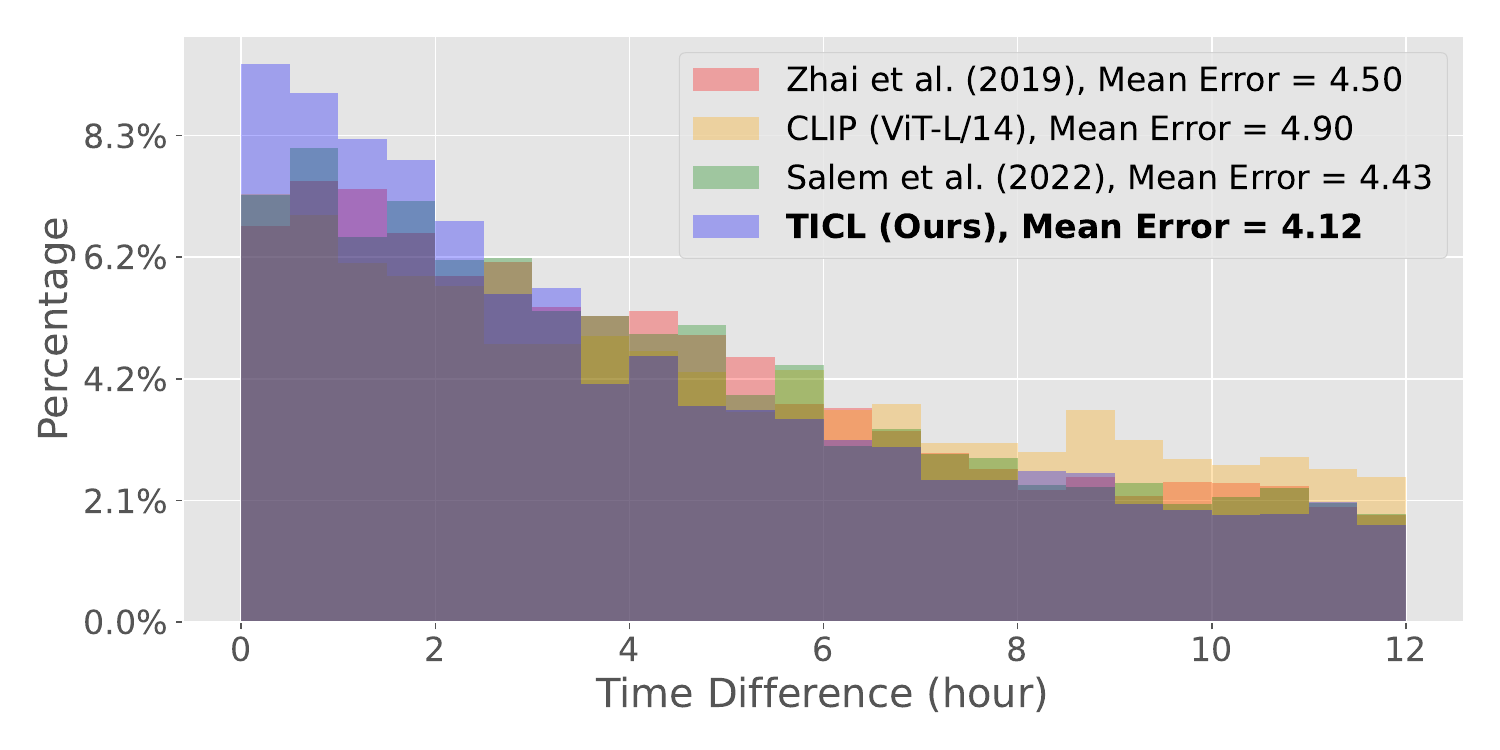}
        \caption{Retrieval time error distribution}
        \label{fig:error_dist_tepid}
    \end{subfigure}%
    \hfill
    \begin{subfigure}[b]{0.49\linewidth}
        \centering
        \includegraphics[width=\linewidth]{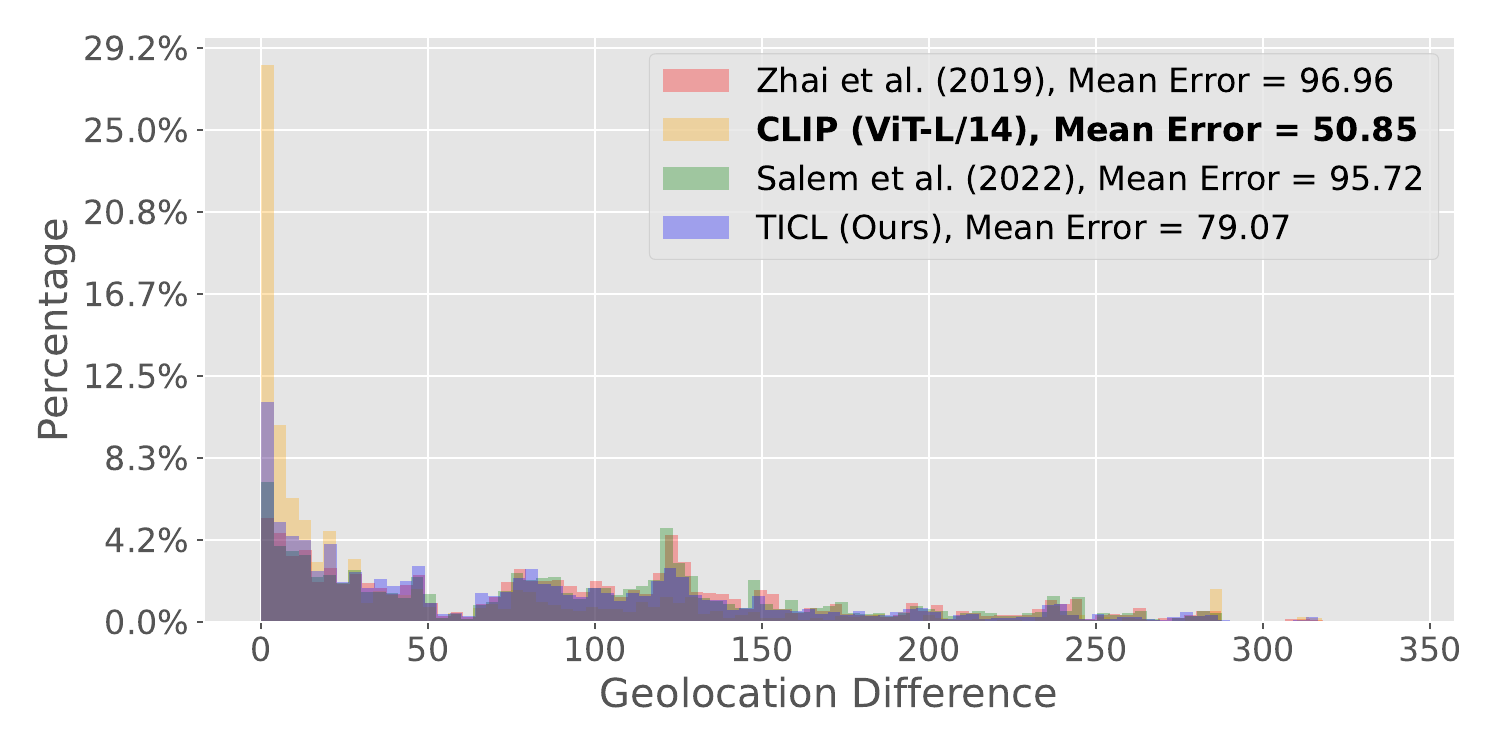}
        \caption{Retrieval geolocation error distribution}
        \label{fig:geo_error_dist_tepid}
    \end{subfigure}%
    \caption{\Update{\textbf{Comparison of geolocation and time error distribution.} It is collected among top-100 retrieved images using different feature extractors.}}
    \label{fig:error_dist}
\end{figure*}

{
We also analysed the distribution of metadata differences between the retrieved images and their corresponding query images. Specifically, \cref{fig:error_dist_tepid} illustrates the distribution of clock time errors among the top-100 retrieved samples for different features. The results show that TICL retrieves a higher percentage of images with smaller time errors compared to other features.} \cref{fig:geo_error_dist_tepid} further shows the geolocation error distribution. Images retrieved by vanilla CLIP embeddings are geographically closest to the queries, suggesting that CLIP represents a rich understanding of scene priors strongly related to geolocations, which was delineated in some previous Visual Place Recognition (VPR) \Update{work} using CLIP backbone \citep{radford2021learningtransferablevisualmodels, 10361537, geoclip}. We suspect that this contextual awareness is partly inherited by TICL, which achieved moderately better performance than other time-aware features of previous \Update{work}. {From this observation, we suspect that TICL disentangled time-aware features from other metadata attributes. To validate this hypothesis, for each query image, we consider an additional task of localising geolocation and time jointly using retrieval. As shown in \cref{tab:geotemp}, the advantage of TICL suggests it has a more balanced capability of understanding geolocation and time jointly than other models.}

\begin{table}
    \centering
    \caption{\textbf{Joint localisation of geolocation and time.} probabilities that the top-1 retrieved image has GPS coordinates' L1-difference $\le$ \texttt{0.01} and a clock time L1-difference $\le$ \texttt{30} minutes to query images.}
    \label{tab:geotemp}
    \resizebox{0.65\linewidth}{!}{%
    \begin{tabular}{c|cccc}
    \toprule
     Chance & \citet{salem2022timestamp} & \citet{zhai2019learning} & CLIP (ViT-L/14) & \textbf{TICL (Ours)} \\ \midrule
      0.03\% & 2.08\% & 4.17\% & 6.25\% & \textbf{10.42\%} \\ \bottomrule
    \end{tabular}%
    }
\end{table}

\subsection{Video scene classification}
\label{sec:vsc}

Understanding dynamic scenes is an important challenging problem that visual models currently face \citep{miao2021vspw}. A fundamental task in this domain is video scene classification. Pretraining models on static images with object categories have been proven to be helpful in video classification \citep{8099985}. Intuitively, dynamic scenes represented in videos have temporally consistent frames within. {Therefore, despite dynamic scene categories seeming to be irreverential to the time of day, we are particularly curious about whether the proposed TICL model, which is pretrained to estimate clock time for input static images, can provide additional understanding of a continuous sequence of frames other than discrete moments represented by static images.}

\paragraph{Experiment setup:} To assess whether our time-aware models provide valuable priors for understanding different categories of dynamic scenes, we provide classification results under two different constructions. {1) We concatenate the time-aware features from different frozen feature extractors to pretrained VideoMAE backbone \citep{tong2022videomae},  2) directly run linear probing on video frames with the frozen feature extractors.}  We compared the performances under different feature extractors on various scene datasets including Hollywood2-Scene \citep{marszalek09}, YUP++ \citep{6247815} and 360+x \citep{chen2024x360}. Please refer to \cref{sec:video_supp} for implementation details and other experimental settings.

\begin{figure}[t]
    \centering
    \includegraphics[width=0.66\linewidth]{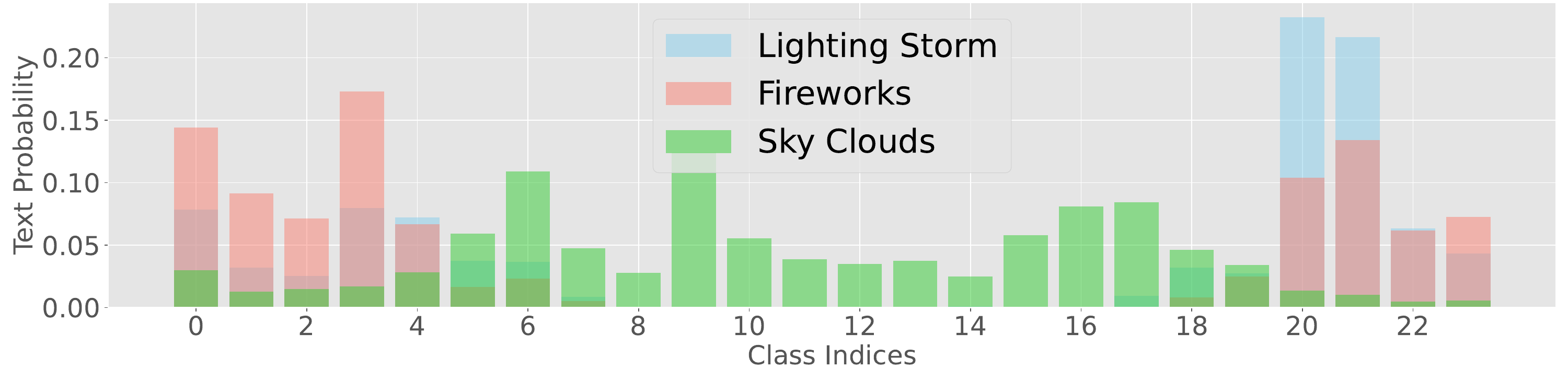}
    \caption{\textbf{The probability distribution for different input text queries of scenes that are seemingly irrelevant to time.} This is calculated by $\mathbf{Softmax} = \frac{\exp \left(T_{CLIP} \cdot T_{i}\right)}{\sum_{j=0}^{|C|-1} \exp \left(T_{CLIP} \cdot T_{j} \right)}$ where \(T_i, T_j, T_{CLIP}\) are the TICL clock time class embeddings and CLIP text embeddings. }
    \label{fig:tq_breakfast}
\end{figure}

\paragraph{Possible correlations between the time of day and scene:} According to \cref{tab:Video-Classifier}, TICL features provide consistent improvements to the scene classification task under different settings. The most straightforward explanation for this boost is that scene classes are correlated with the learned time of day by definition. To prove this, we visualized the cosine similarity between certain text embeddings of certain scenes that clock time class embeddings, as shown in \cref{fig:tq_breakfast}. The imbalanced distributions proved the conceptual correlation of scenes to time due to human activity patterns.

\begin{figure}[h!]
    \centering
    \begin{subfigure}[b]{0.3\linewidth}
        \centering
        \includegraphics[width=\linewidth]{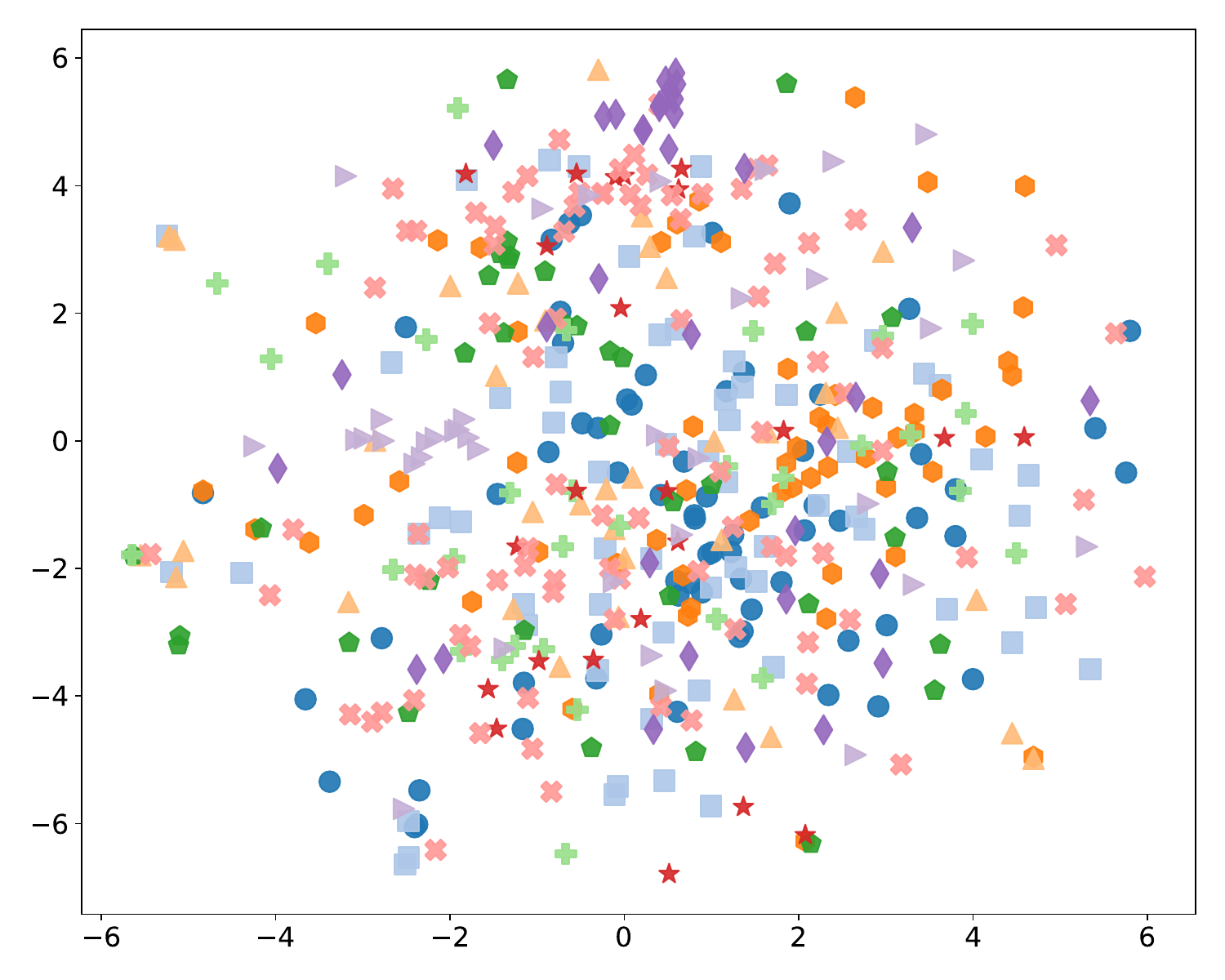}
        \caption{CLIP}
        \label{fig:tsne_clip}
    \end{subfigure}
    \begin{subfigure}[b]{0.3\linewidth}
        \centering
        \includegraphics[width=\linewidth]{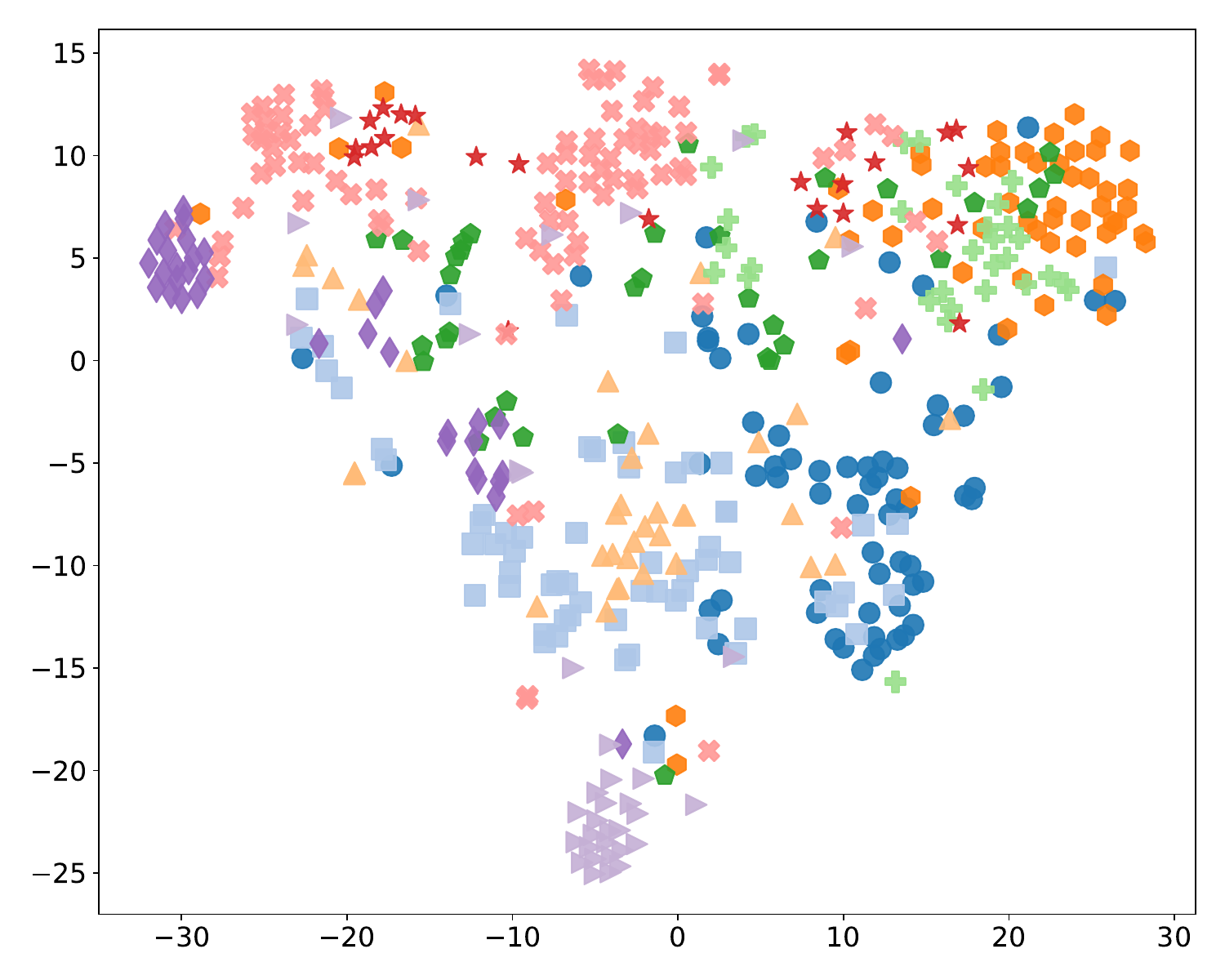}
        \caption{TICL (ours)}
        \label{fig:tsne_ticl}
    \end{subfigure}
    \caption{\textbf{t-SNE visualisation comparison.} It compares video features before the final classifier layer using either (a) CLIP or (b) TICL, on the Hollywood2-Scene dataset \citep{marszalek09}, each different scatter point shape/colour corresponds with classes.}
    \label{fig:tsne_VIDEO}
\end{figure}

\paragraph{Consistency in time-aware frame embeddings:} As shown in \cref{sec:retrieval}, the TICL representations can capture similarities between images with close clock times. Natural videos, although they sometimes involve drastic subjects or view movement, frames within each should still represent continuous time periods. TICL features for frames across the whole video should be more consistent than those of vanilla CLIP, which have stronger locality per frame \citep{10.1145/3474085.3479207}. This intra-video consistency allows for more general time-aware priors extracted using TICL. The t-SNE visualisation of the video features in \cref{fig:tsne_VIDEO} supports this claim, showing that TICL features are more separable than vanilla CLIP features (see \cref{sec:video_exp_setup_supp} for a more in-depth analysis of the phenomena and claims above).

\begin{table*}[t]
\caption{Performances on the video scene classification task.}
\label{tab:Video-Classifier}
\centering
\resizebox{0.85\linewidth}{!}{%
\begin{threeparttable}
\begin{tabular}{l|cccc}
\toprule
     Classifier               & Hollywood2-Scene $\uparrow$ & YUP++ \tnote{$\dagger$} $\uparrow$ & 360+x (Panoramic) $\uparrow$ & 360+x (Third-view) $\uparrow$ \\
\midrule
VideoMAE \citep{tong2022videomae}    &      48.83\%      &   97.29\%    &        53.70\%           &         54.55\%          \\
VideoMAE + CLIP (ViT-L/14) &   52.92\%    &   \textbf{98.33\%}    &        57.40\%           &          50.91\%         \\
VideoMAE + \citet{salem2022timestamp} &   45.53\%    &   97.50\%    &        44.45\%           &        52.72\%            \\
VideoMAE + \citet{zhai2019learning} &   51.03\%    &   97.71\%    &        48.15\%           &         56.36\%           \\
\textbf{VideoMAE + TICL} &       56.53\%     &  \textbf{98.33\%}     &       \textbf{59.26\%}            &     \textbf{58.18\%}   \\ \midrule
CLIP (ViT-L/14) (Linear Probing) & 39.69\% & 97.08\% & 35.19\% & 11.10\% \\
\textbf{TICL (Linear Probing)}  & \textbf{57.04\%} & \textbf{98.33\%} & 51.85\% & 42.59\%  \\

\bottomrule
\end{tabular}%
\begin{tablenotes}
\footnotesize
\item[$\dagger$] We use an unofficial train/val/test split of 5:1:4, since the original 1:9 train/test split overfit prematurely.
\end{tablenotes}
\end{threeparttable}
}
\end{table*}

\subsection{Time-aware image editing}
\label{sec:image_editing}
As aforementioned in \cref{sec:TICL_method}, the TICL model can provide the corresponding embeddings for certain periods of the day. Therefore, it is natural to consider using these clock timestamp embeddings as guidance to edit images toward different classes. To assess the extent to which clock time embeddings aid this task, we adopted the following experiment framework from \citet{Patashnik_2021_ICCV} that conducts image editing via latent vector searching through optimisation steps instead of tuning the models directly.

\begin{figure*}[ht]
    \centering
    \includegraphics[width=0.8\linewidth]{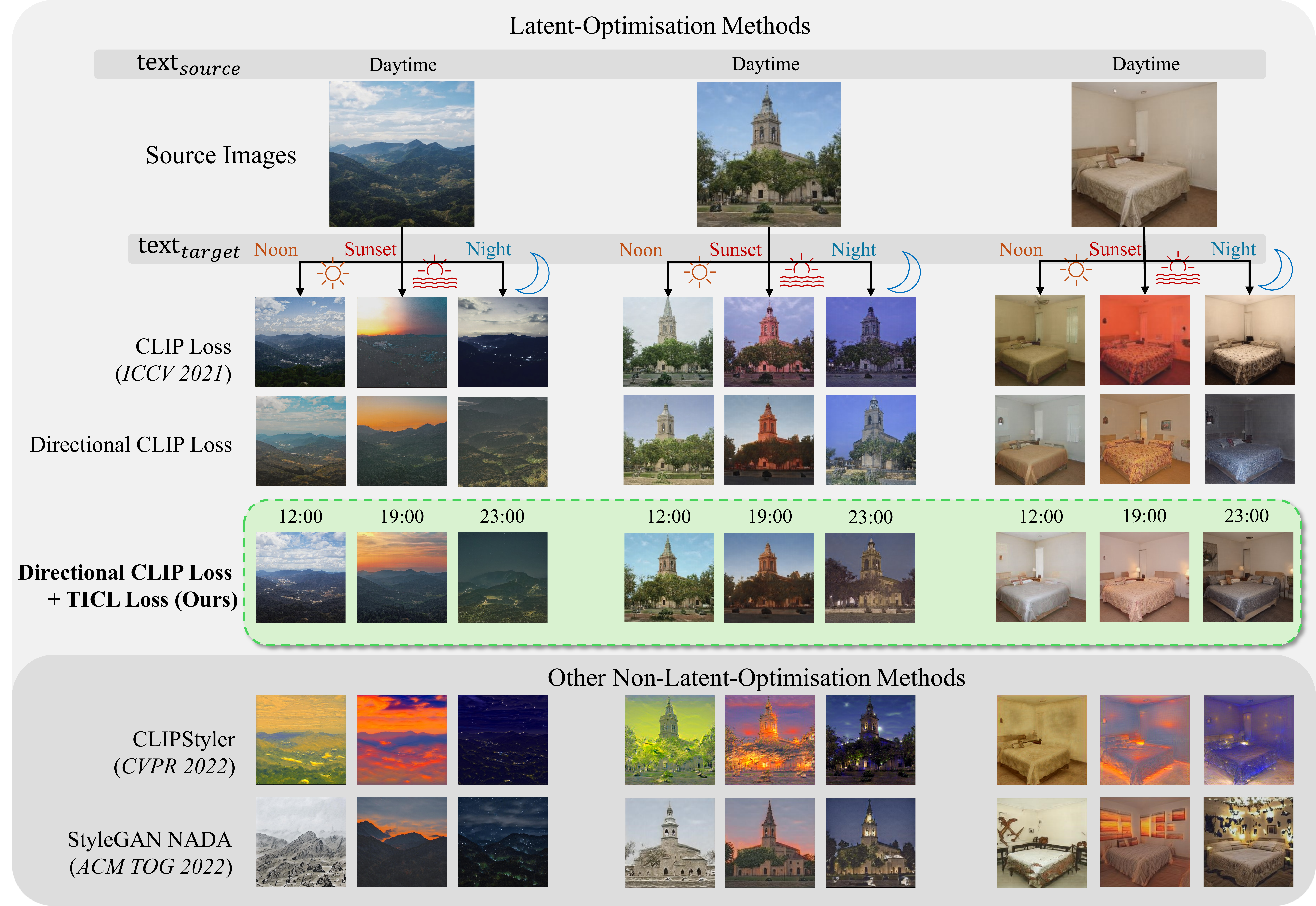}
    \caption{\textbf{Time-aware image editing.} It shows the results of applying our time-aware editing method (green overlay) on three different StyleGAN2 models trained on LHQ-Landscape \citep{ALIS}, LSUN-Church, and LSUN-Bedroom \citep{yu15lsun} datasets. The results of other non-latent optimisation methods are also demonstrated (under \textcolor{gray}{grey} overlay).}
    \label{fig:StyleGAN}
\end{figure*}

\paragraph{Experiment setup:} To provide comprehensive evaluations, we conducted experiments on three different baseline StyleGAN2 models \citep{Karras2019stylegan2} focusing on different subjects trained on \citep{ALIS,yu15lsun}. The pretrained generator weights are adopted from existing codebases \citet{pinkney_lhq-sg2-1024,epstein2022blobgan,Karras2020ada}. The editing pipelines were restricted to follow the same latent optimisation baseline method introduced in StyleCLIP \citep{Patashnik_2021_ICCV}. Additionally, we designed a new time-aware synergy loss combining directional CLIP loss and TICL feature similarity loss. Specifically, the editing process can be formulated as: 
\[
\underset{w \in \mathcal{W^{+}}}{\arg \min }\left(\lambda_{1} \mathcal{L}_{T I C L}+\lambda_{2} \mathcal{L}_{\text {CLIPdir }}+\lambda_{l2}\left\|w-w_{\text {source}}\right\|_{2}\right)
\] in which \(w, w_{source}\) represents latent vectors for ongoing edit outcomes and original images, 
(design, hyperparameter and implementation details in \cref{sec:lo_appendix}).

\paragraph{Qualitative evaluation:} The proposed time-aware synergy loss yields the most plausible synthesis outcome as illustrated in \cref{fig:StyleGAN}. The limitations of solely text-guided image editing methods could be due to their susceptibility to certain adversarial solutions fooling CLIP image encoders with certain patterns only \citep{liu2021fusedream}. Specifically, \cref{fig:StyleGAN} shows the vanilla StyleCLIP edits using the CLIP loss tend to focus on the general tint of the image but fail to reflect realistic illuminations. We find that replacing the CLIP loss with a directional variant introduced in previous \Update{work} \citep{gal2021stylegannada, kwon2022clipstylerimagestyletransfer} can assist in overcoming larger domain gaps. Despite showing improvements over the baseline editing method, the results still show unrealistic artefacts and shape distortions. These limitations show the necessity of incorporating additional time-aware features other than just guidance text embeddings when computing loss functions for image edits. Our qualitative evaluations demonstrated the effectiveness of the TICL embeddings on the specific task. We also included other baseline method results that work under different frameworks other than latent optimisation for a more comprehensive comparison. {See more quantitative evaluations \Update{(\cref{tab:fid_score})}, user studies \Update{(\cref{tab:human_evaluation})} and results on TICL-aided editing with diffusion models in \cref{sec:lo_appendix} and \cref{sec:diffusion} respectively}.

%% file: sec/6_conclusion.tex
\section{Conclusion}

In this paper, we tried to answer the question of \textit{what time tells us}, through exploring the pretext task of time-of-day estimation and downstream tasks. A new reliable benchmark dataset, \textit{TOC} was introduced to support the pretext task, consisting of images captured in natural settings with verified timestamps. This dataset addresses the limitations of existing datasets by providing a more diverse and realistic collection of images that better reflect daily visual experiences. Building upon that, a new learning paradigm (\textit{TICL}) was proposed, which aligns clock timestamp and image in representation space via a pretext time prediction task, surpassing previous \Update{work} in time-of-day estimation. The learned time-aware representations were further studied via validations on several downstream tasks. The strong performance in these downstream tasks highlighted its capability to recognise the similarity of the captured time (in time-based image retrieval), frame-coherent priors in TICL for video scene understanding (significantly improved video scene classification), and produce realistic and time-consistent performance in time-aware image editing (accurately reflecting typical lighting conditions for different times of day). 

\section*{Acknowledgement} 
This project is partially supported by the Royal Society grants (SIF\textbackslash R1\textbackslash231009,  IES\textbackslash R3\textbackslash223050) and an Amazon Research Award.
The computations in this research were performed using the Baskerville Tier 2 HPC service. Baskerville was funded by the EPSRC and UKRI through the World Class Labs scheme (EP\textbackslash T022221\textbackslash1) and the Digital Research Infrastructure programme (EP\textbackslash W032244\textbackslash1) and is operated by Advanced Research Computing at the University of Birmingham.

%% file: sec/X_suppl.tex
\vspace{6mm}
\appendix

\section*{Appendix Roadmap}
This is the appendix for the main paper. Here is a general roadmap describing the contents of each part of this document supporting the main paper:
\begin{itemize}
    \item We first provide additional details to the datasets in the \cref{sec:dataset_details}, which includes how we cleaned the originally noisy data into datasets that reflect the diversity of natural images paired with accurate metadata.
    \item  In \cref{sec:ticl_appendix}, we cover the detailed illustration of the implementation of the model and the setup of the experiment. Along with additional performance and error analysis. {We also provide additional results testing the ability to jointly predict date related metadata other than just the clock time.}
    \item In \cref{sec:precision_limits}, we explore various scalar encoding methods to time variables on the pre-text task through an regression example in \cref{sec:regression_limits}. We also discussed the inherent trade-off of fine-grained classification via an additional ablation to the number of classes in \cref{sec:class_num}.
    \item \cref{sec:retrieval_supp} provides additional qualitative evaluation to the time-based image retrieval task.
    \item \cref{sec:video_supp} gives experimental setup details, as well as more evidences of the intra-video consistency identified in the main paper in \cref{sec:frame_variance}.
    \item In \cref{sec:lo_appendix}, we provide a detailed setup of the experiment along with additional qualitative and quantitative evaluation of the capability of time-aware features in image editing tasks. 
    \item \cref{sec:diffusion} also shows results of using time-aware features to further improve the fidelity \wrt clock time via time-aware features under more advanced diffusion model baselines.
    \item In the main paper, we discussed about the implications of time-awareness in visual scene understanding, in \cref{sec:text_query}, we provide more examples of text query about the conceptual relations between clock time and scene/action/objects text embeddings.
    
\end{itemize}

\begin{figure*}[h]
    \centering
    \begin{subfigure}[b]{0.49\linewidth}
        \centering
        \includegraphics[width=\linewidth]{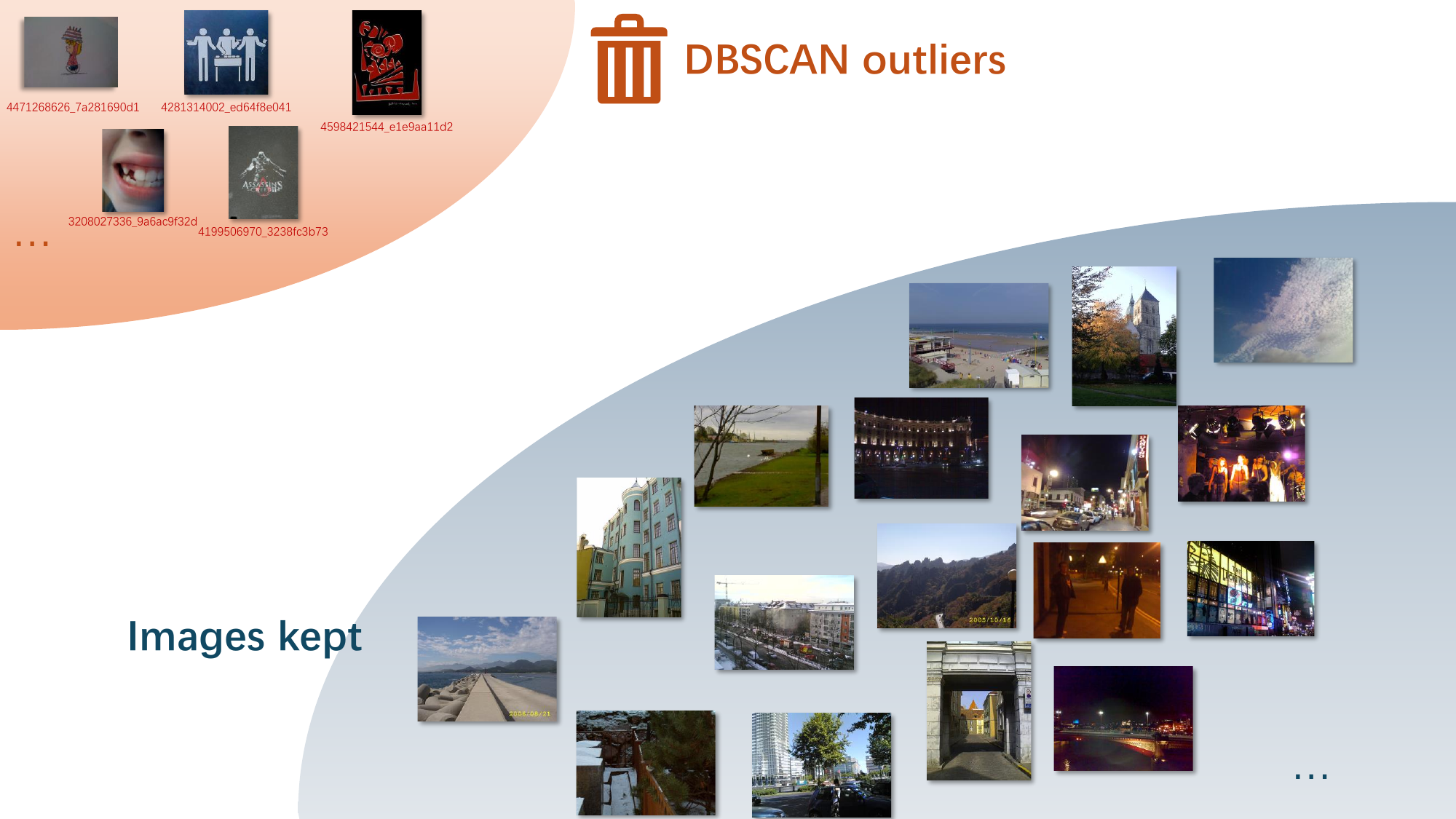}
        \caption{}
        \label{fig:Cleansing1}
    \end{subfigure}
    \hfill
    \begin{subfigure}[b]{0.49\linewidth}
        \centering
        \includegraphics[width=\linewidth]{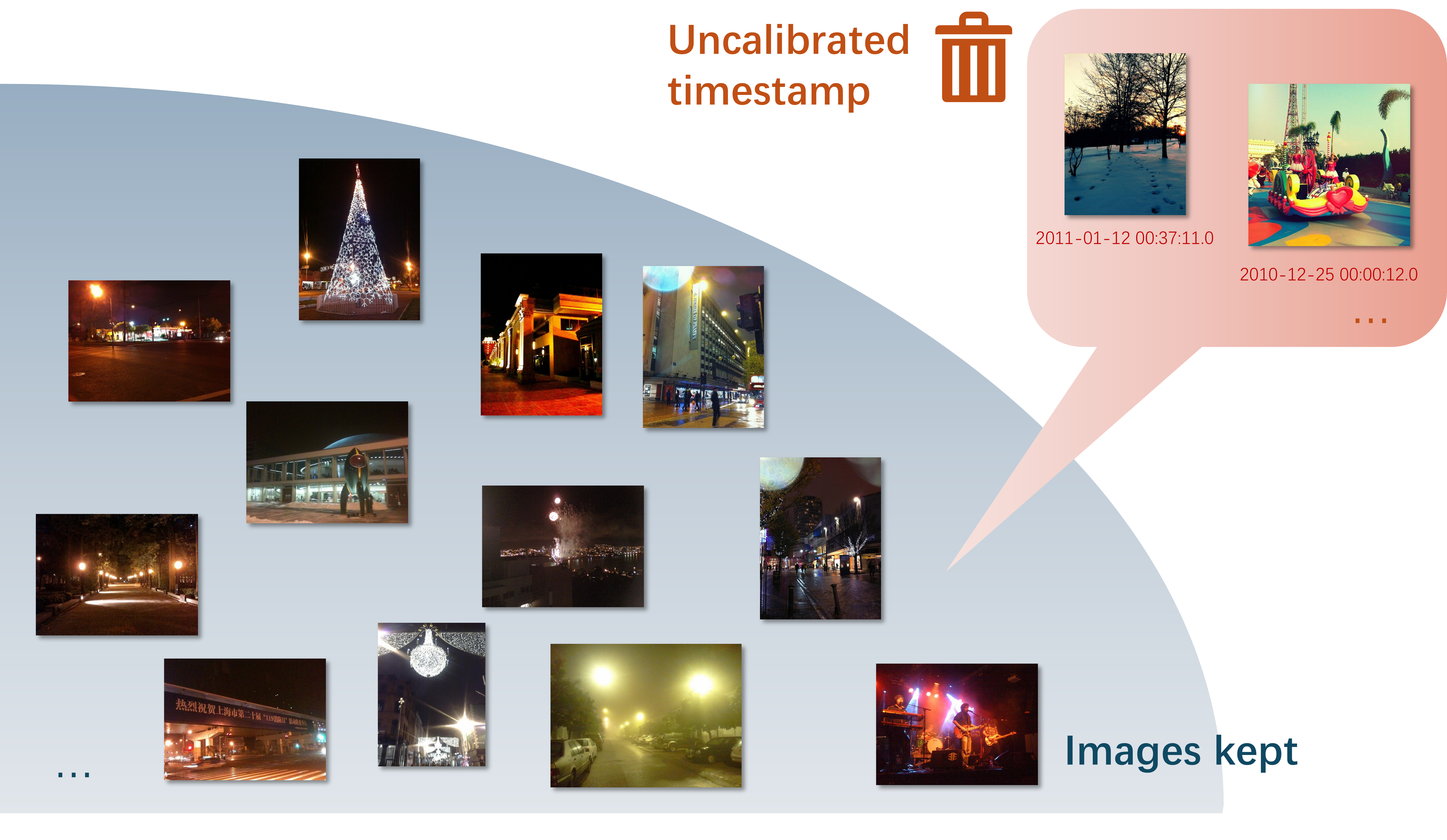}
        \caption{}
        \label{fig:Cleansing2}
    \end{subfigure}
    \caption{\textbf{Dataset filtering process,} where (a) shows examples of finding unnatural images in DBSCAN \citep{DBSCAN} outliers that may degrade dataset quality, and (b) shows examples of removed images with uncalibrated clock timestamps.}
    \label{fig:Cleansing}
\end{figure*}
\section{More Details on Datasets}
\label{sec:dataset_details}
\subsection{The proposed TOC dataset}
\label{sec:tocdataset_details}
\begin{figure}
    \centering
    \includegraphics[width=1\linewidth]{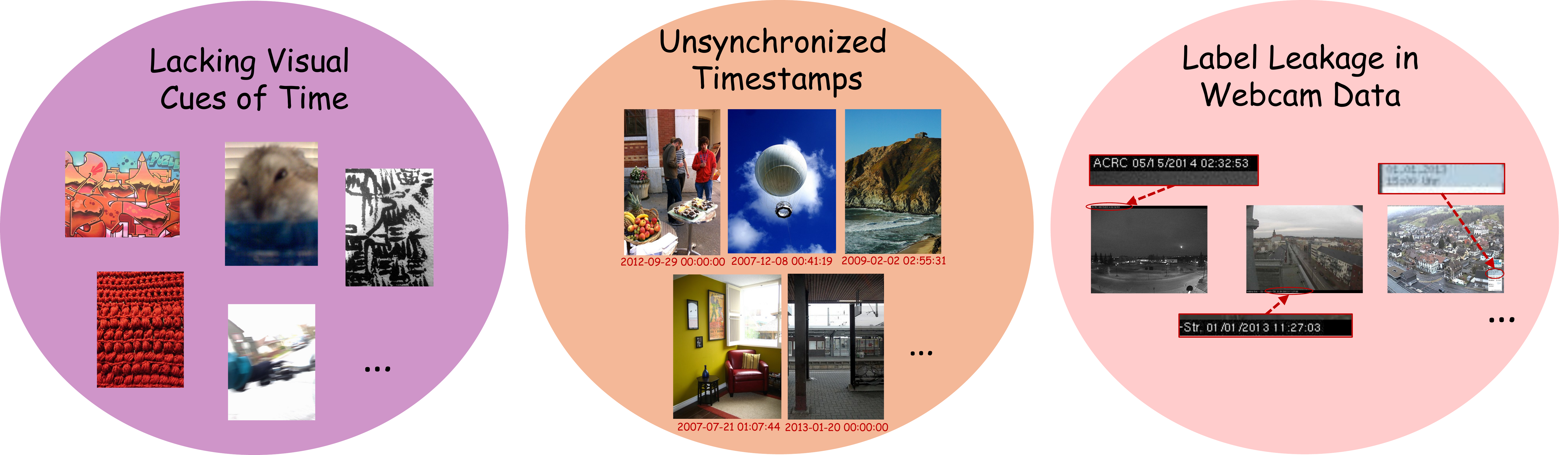}
    \caption{\Update{Existing issues in previous datasets \citep{salem2020dynamic, thomee2016yfcc100m, jacobs07amos}}}
    \label{fig:existing_issues}
\end{figure}

\Update{\paragraph{Comparison to previous works:} In this work, we introduce a new benchmark dataset that combines images from the YFCC100M \citep{thomee2016yfcc100m} and Cross-View Time datasets \citep{salem2020dynamic}. As we briefly summarised in \cref{sec:dataset}, the major differences of our datasets to previous static image datasets featuring time-metadata lies in the view/appearance diversity and metadata correctness. \Cref{tab:dataset_comparison} gives an overview of the differences. For previous generic social media image datasets, there have been persisting issues of unreliable timestamp metadata due to unsynchronized user/device activity. Apart from unreliable groundtruth, the visual appearances of time in some of the images are often undefined in non-photographic images which do not reflect any natural time-of-day. Despite such these problems are mitigated for datasets with proportional samples from static surveilance camera, the repetitive views and occasional ground-truth leakage overlayed on the camera footage suggests limitity usablilty. These issues are depicted separately in \cref{fig:existing_issues}. Therefore, our dataset resolved these issues by manually verifying metadata fidelity given the image on purely social media samples.}

\begin{table}[t]
\centering
\caption{\Update{Comparison of existing image datasets with timestamps.}}

\resizebox{\textwidth}{!}{%
\begin{tabular}{lllll}
\toprule
\textbf{Dataset} & \textbf{Image source} & \textbf{Timestamp reliability} & \textbf{Scene diversity} \\
\midrule
MIRFLICKR-1M \citep{MIRFLICKR1M} & mobile / miscellaneous & \textcolor{BrickRed}{\textbf{\ding{55}}} & high \\
YFCC100M \citep{thomee2016yfcc100m} & mobile / miscellaneous & \textcolor{BrickRed}{\textbf{\ding{55}}} & high \\
AMOS \citep{jacobs07amos} & fixed webcams & \textcolor{BrickRed}{\textbf{\ding{55}}} & limited \\
CrossView Time (CVT) \citep{salem2020dynamic} & webcams + mobile devices & \textcolor{BrickRed}{\textbf{\ding{55}}} & mixed \\
\textbf{TOC (ours)} & \textbf{wild-view natural photography images} & \textbf{verified \& timezone-aligned} & \textbf{high} \\
\bottomrule
\end{tabular}%
}

\label{tab:dataset_comparison}
\end{table}

\Update{\paragraph{Detailed curation steps:}} Now, we cover more details of the dataset curation process. \cref{fig:Cleansing} gives a clear illustration of the data filtering steps to the dataset, improving the sample quality and metadata reliability. We firstly inspected all the night-time images with average pixel brightness $\geq 100$ to determine whether they have clearly mislabeled timestamps. Specifically, extreme cases like polar day were considered, so images with $|\text{altitudes}| \geq 75$ were retained regardless of illumination. \Update{This step removes clearly unsynchronized images, which do not align with human consensus about nighttime illuminations.} To reduce the workload of filtering unnatural images, we firstly partitioned the images into 24 different hour intervals; within each of them, we apply DBSCAN ($\epsilon = 10, \text{minPts} = 100$) on ResNet-18 features, which gives a majority group and outliers. We recruited workers to manually review all the outlier images determining whether to add them back to the dataset. \Update{This pipeline allows for efficient removal of unnatural images without proper visual cues about time in the dataset by looking at samples groups with distinct features to the majority group containing natural photographs}.

\begin{figure*}[t]
 \centering
    \begin{subfigure}[b]{0.35\linewidth}
        \centering
        \includegraphics[width=\linewidth]{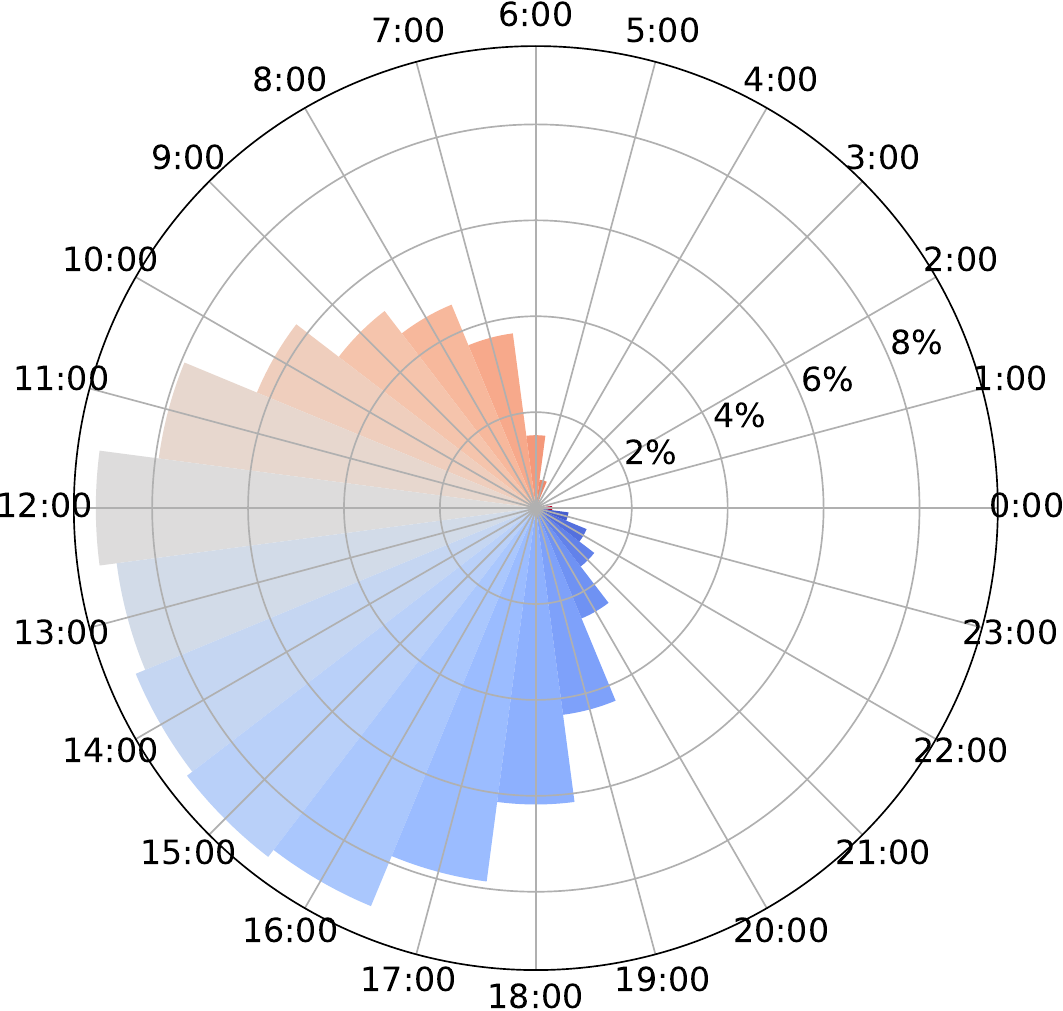}
        \label{fig:tepid_dist}
    \end{subfigure}
    \begin{subfigure}[b]{0.35\linewidth}
        \centering
        \includegraphics[width=\linewidth]{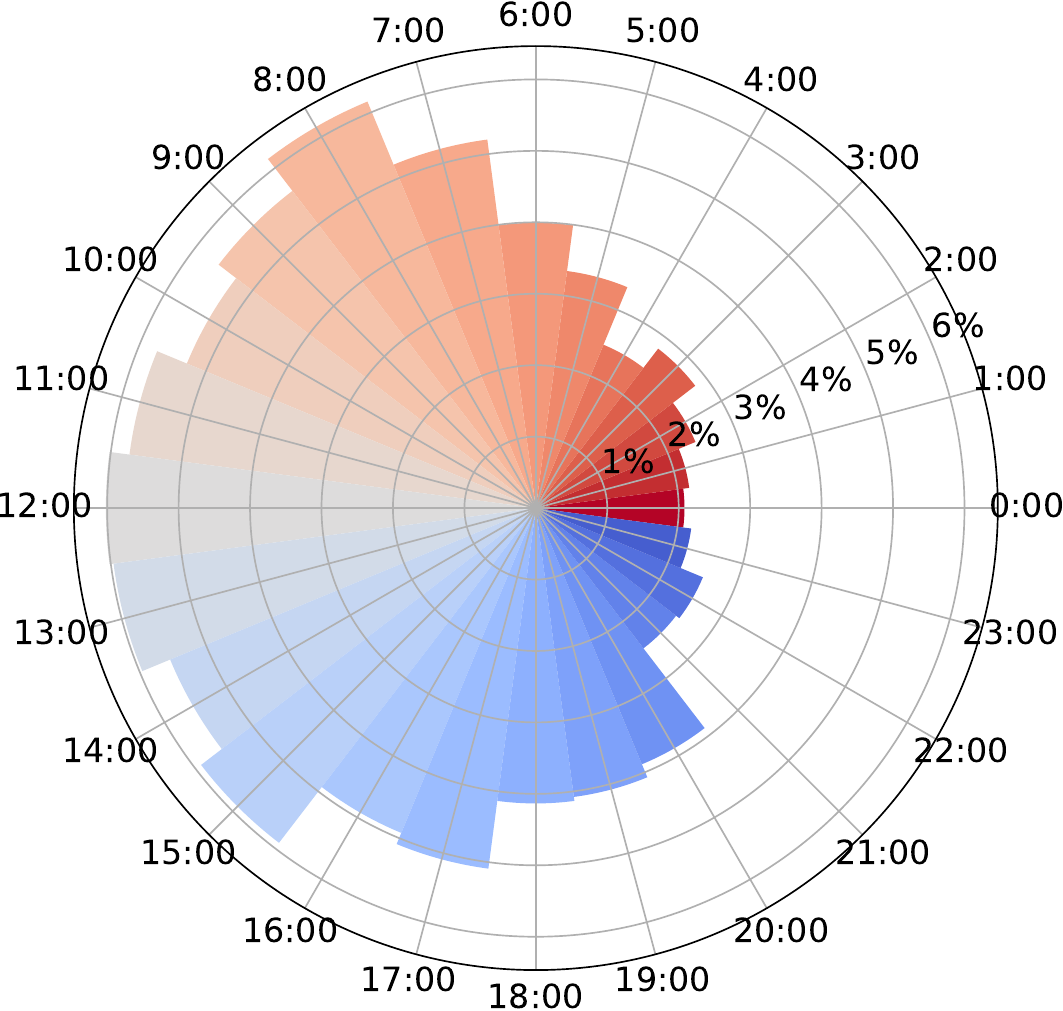}
        \label{fig:amos_dist}
    \end{subfigure}
    \caption{\textbf{Dataset hourly sample distribution,} where (a) shows hourly sample distribution for TOC dataset, in which daytime images are significantly more prevalent than nighttime images, and (b) shows hourly sample distribution for AMOS-test dataset displaying a similar skewed but more balanced distribution towards daylight hours.}
    \label{fig:domain_dist}
\end{figure*}

\begin{figure}
    \centering
    \includegraphics[width=0.95\linewidth]{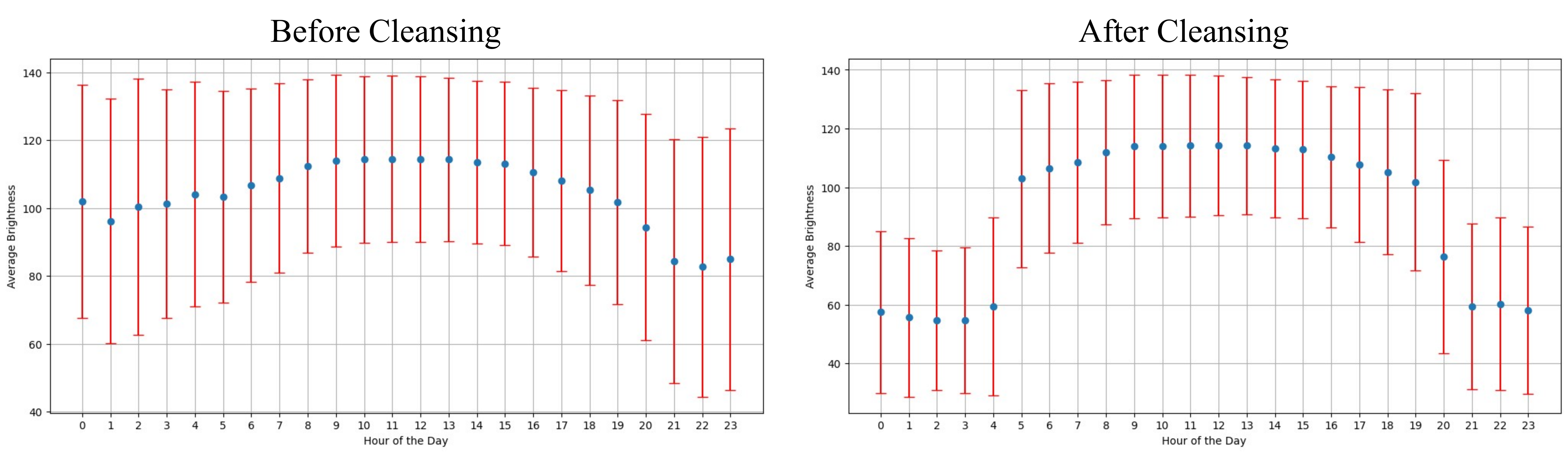}
    \caption{\Update{Mean brightness with standard deviation over images taken at different periods of the day.}}
    \label{fig:bright_dist}
\end{figure}

\Update{We conduct a statistical sanity check on the dataset–processing pipeline. The most intuitive visual cue correlated with the time of day is scene illumination, which we approximate by the statistical average of pixel brightness. In \cref{fig:bright_dist} we observe that, \emph{before} cleaning, images labelled 00:00–06:00 in the previous dataset have brightness comparable to daytime photos, and an unnatural gap appears between the 21:00–23:59 and 00:00–03:59 bins. Both patterns violate common sense, where 1) late-night images should be markedly darker than daytime images, and 2) illumination should change most rapidly around sunrise and sunset, while remaining relatively stable during midnight and noon periods. These anomalies point to unsynchronised capture timestamps in the unprocessed dataset. \emph{After} cleansing, the TOC dataset follows the expected smooth day–night trend, confirming that our pipeline yields data consistent with human intuition about illumination over a 24-hour cycle.}

Following the data filtering, we partitioned the TOC dataset into a training set and a test set at a $9:1$ ratio, with stratified sampling to ensure that the clock time distributions of both subsets were approximately equivalent. We observed a significant scarcity of images with reliable metadata captured at night compared to daytime images. This observation corroborates our hypothesis that the distribution of timestamps in images shared on social media is inherently unbalanced as depicted in \cref{fig:domain_dist}. 

Such imbalance presents challenges in learning equitable embeddings for class time periods that are under-represented due to limited sample availability. This imbalance necessitates strategic approaches to model training that can adequately compensate for these discrepancies.

\subsection{AMOS test dataset}
\label{sec:AMOS_detail}
\begin{figure*}[t]
    \begin{subfigure}[b]{0.27\linewidth}
        \centering
        \includegraphics[width=\linewidth]{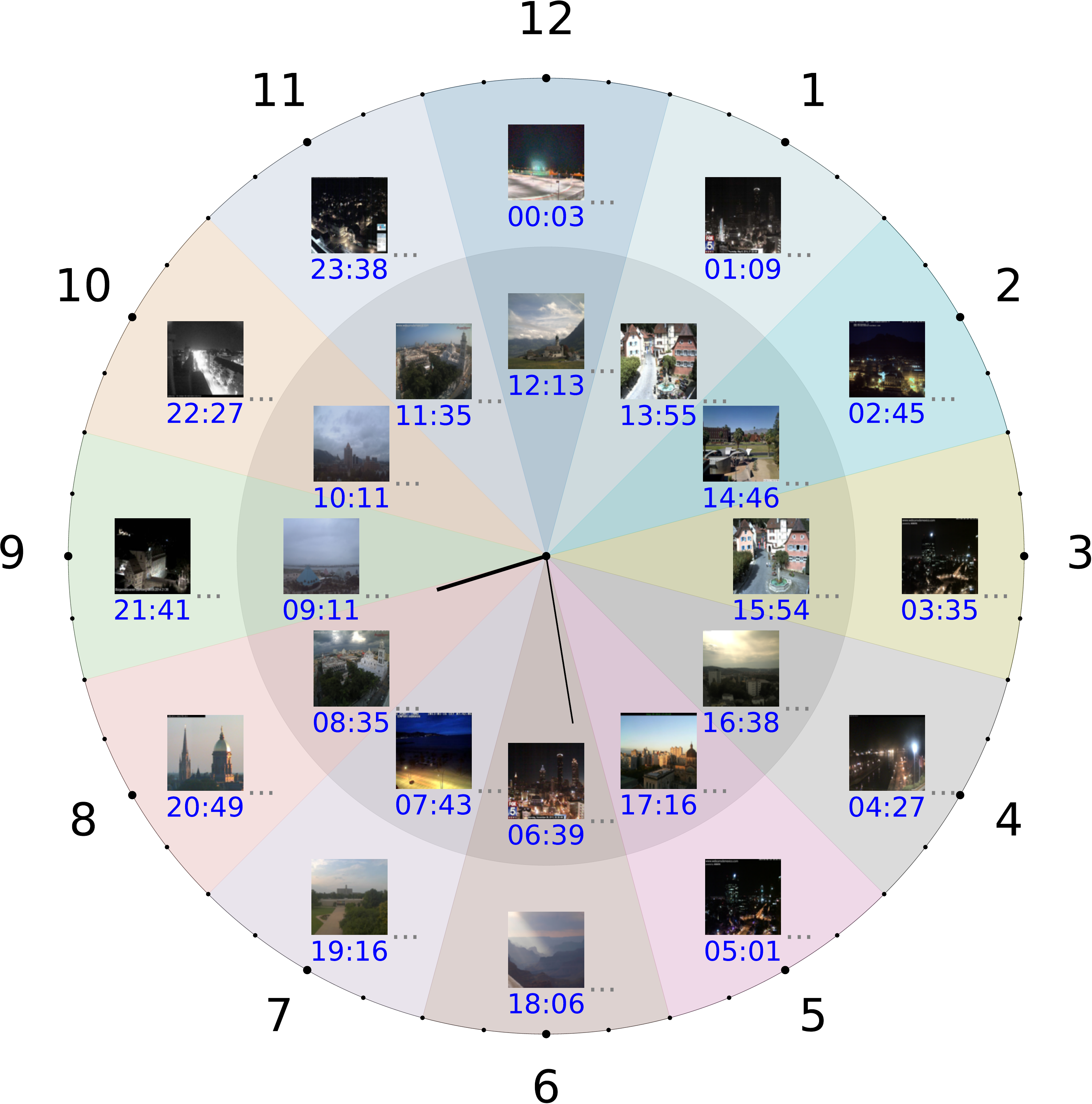}
    \end{subfigure}
    \centering
    \begin{subfigure}[b]{0.65\linewidth}
        \centering
        \includegraphics[width=\linewidth]{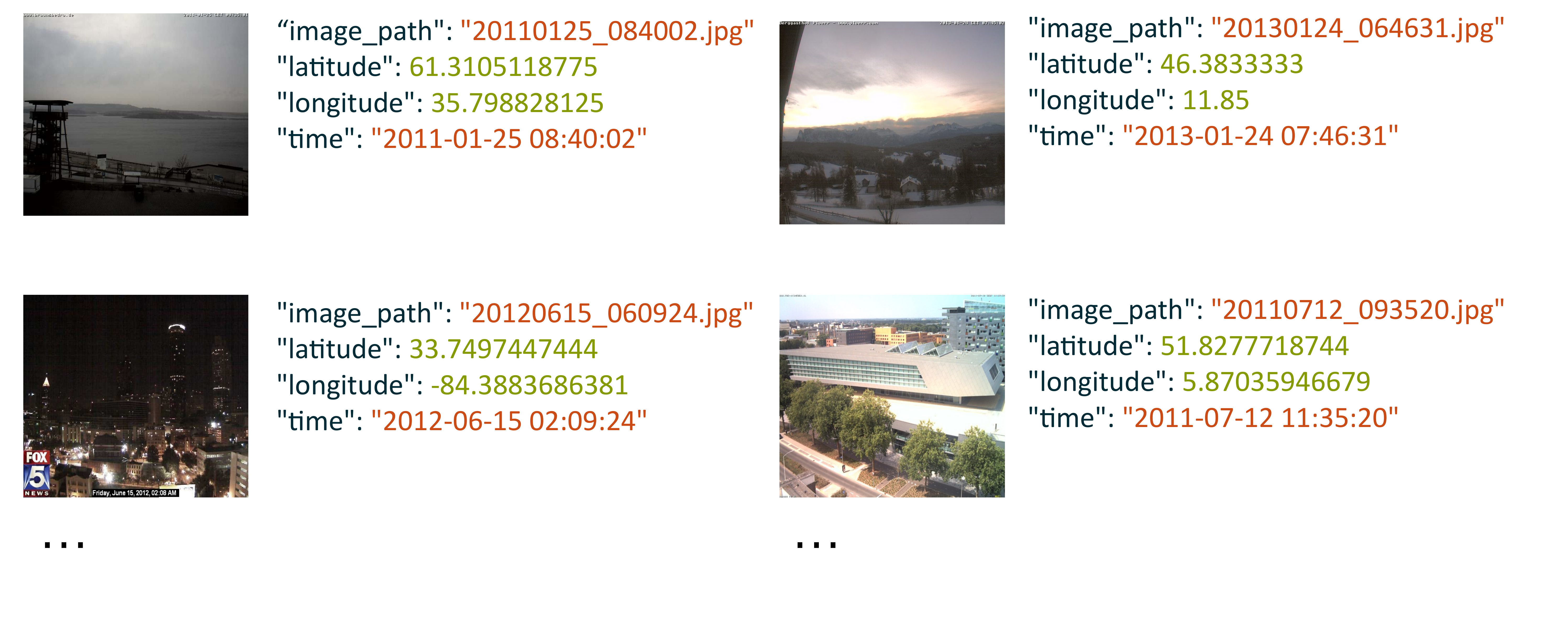}
    \end{subfigure}%

    \caption{\textbf{Sample images from the AMOS test dataset.} The images showcase different scenes captured by stationary surveillance cameras at various times of the day with decent visual quality.}
    \label{fig:AMOS}
\end{figure*}

\noindent \textbf{Dataset Filtering and SNR Estimation:} The AMOS-test dataset was selected from the CVT test set, containing 5,000 AMOS images, which was further reduced to 3,556 images. The dataset filtering involves several steps to ensure metadata reliability and sample quality. First, we calibrated the original UTC timestamps to their respective local timezones using the geolocation metadata. Then, we filtered out (1) noisy images with low Signal-to-Noise Ratio, where the SNR is estimated using a block-based variance method. Specifically, for an image \( I \) with \( N \) pixels, the SNR is computed as 
\[
\mathrm{SNR}(I) = 10 \cdot \log_{10} \left( \frac{\sigma^2_{\text{signal}}}{\sigma^2_{\text{noise}}} \right),
\]
where the noise variance \( \sigma^2_{\text{noise}} \) is estimated as the average variance over the lowest 10\% of non-overlapping blocks of size \(16 \times 16\) pixels, and the signal variance is given by 
\[
\sigma^2_{\text{signal}} = \sigma^2_{\text{total}} - \sigma^2_{\text{noise}},
\]
with \( \sigma^2_{\text{total}} \) being the variance of the entire image. Images with \( \mathrm{SNR}(I) \leq 15 \) were discarded. After cleansing, the average SNR improved from 1.93 (std = 10.35) to 3.38 (std = 3.50). This filtering ensured that only images with recognizable time-of-day related appearance were included in the evaluation. \Cref{fig:AMOS} shows a few sample images from the dataset.

As the images were captured automatically by surveillance cameras with fixed views, the AMOS test set represents a different domain to the proposed TOC dataset. Although the dataset contains repetitive visual appearances due to the stationary setup of the cameras, it benefits from a more balanced distribution of timestamps throughout the day, as shown in \cref{fig:amos_dist}.

\section{Implementation Details of TICL}
\label{sec:ticl_appendix}

In the main paper, we covered the high-level design of the TICL model we devised to learn time-awareness via a clock time estimation pre-text task. This section provides additional details.

\begin{figure*}[ht]
    \centering
    \includegraphics[width=0.75\linewidth]{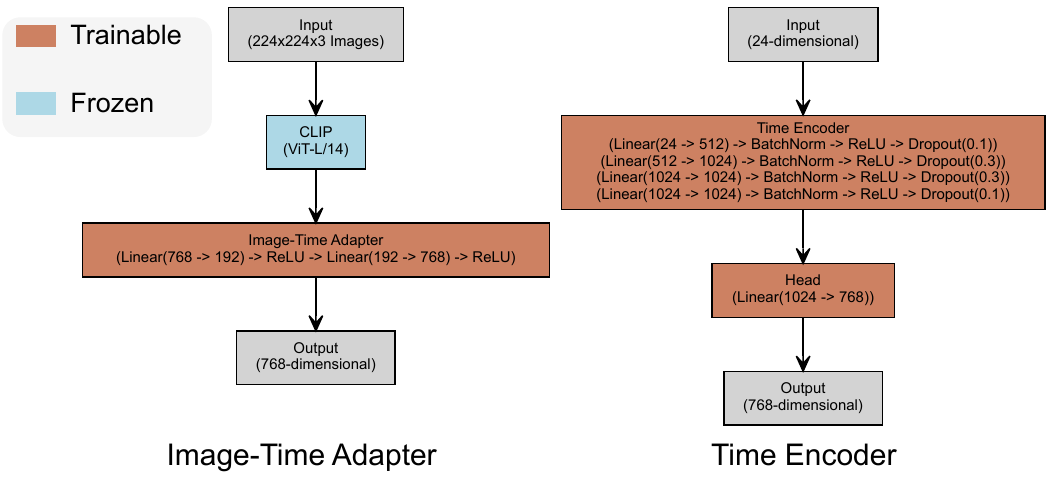}
    \caption{Visualisation of TICL sub-module architectures.}
    \label{fig:architectures}
\end{figure*}

\subsection{Model details}
\label{sec:model_details}
\paragraph{Time Encoder:} The Time Encoder consists of several fully-connected layers, with the detailed architecture shown in \cref{fig:architectures}. The raw timestamps are first preprocessed into 24 one-hot class embeddings. The Time Encoder then takes these input class embeddings and projects them to the desired representation space. 

\paragraph{Image-Time Adaptor module:} The Image-Time Adaptor module is employed to adapt the raw backbone features with Time Encoder outputs, as depicted in \cref{fig:architectures}. Training the Image-Time Adaptor module and Time Encoder jointly using a contrastive learning scheme allows for effective alignment between the two modalities.
\begin{figure*}[h!]
    \centering
    \begin{subfigure}[b]{0.48\linewidth}
        \centering
        \includegraphics[width=\linewidth]{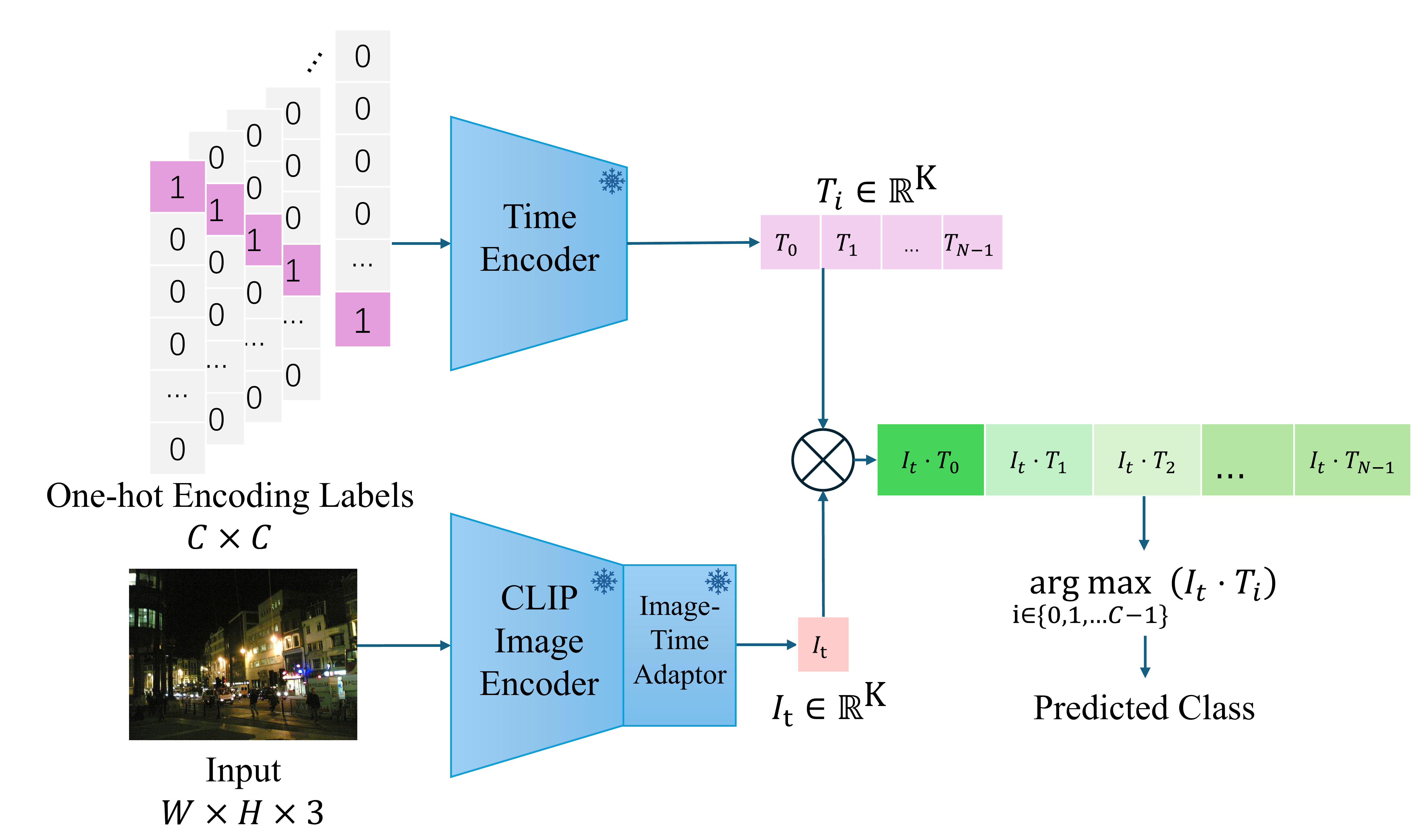}
        \caption{Classification-based inference}
        \label{fig:test_classification}
    \end{subfigure}%
    \hspace{0.02\linewidth}%
    \begin{subfigure}[b]{0.48\linewidth}
        \centering
        \includegraphics[width=\linewidth]{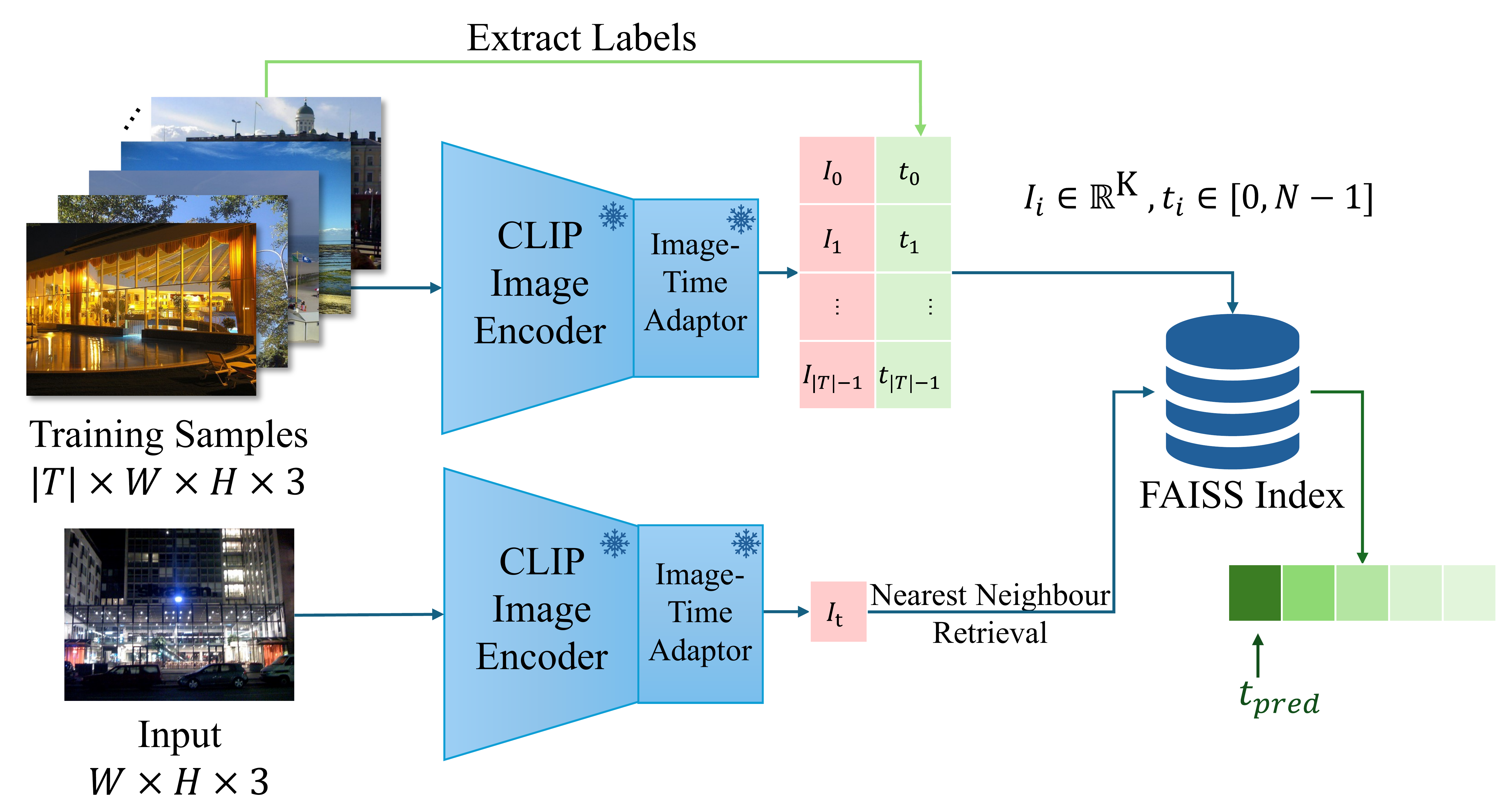}
        \caption{Nearest-neighbour inference}
        \label{fig:test_classification_retreival}
    \end{subfigure}
    \caption{\textbf{Detailed illustration of different inference pipelines.} In (a), the model selects the clock time with the highest similarity to the input image from a finite set of clock time class embeddings. (b) shows that the model estimates clock timestamp by finding the corresponding timestamp of the nearest-neighbour to the input images from the training set based on the sample-specific TICL embeddings.}
    \label{fig:TEST_arch}
\end{figure*}
\subsection{Details in clock timestamp estimation inference pipelines} 
\label{sec:inference}
Two different clock timestamp estimation inference pipelines were devised. The first pipeline, shown in \cref{fig:test_classification}, adheres to the classification scheme, selecting the timestamp with the highest similarity within a finite clock timestamp embedding pool encoded from \(C\) one-hot embeddings. The second pipeline, shown in \cref{fig:test_classification_retreival}, converts the problem to a retrieval-style formulation, using known image-timestamp pairs from the training set. The model returns the class-level timestamp of the most similar samples to it in the training set using an efficient vector search engine \citep{johnson2019billion}.

\Update{\paragraph{Settings for VQA baselines:} For baseline VQA based methods tested in \cref{tab:time_prediction}, namely the BLIP \citep{BLIP} and GPT-4o-mini \citep{openai2024gpt4ocard}, we simply use a pipeline with one round Q\&A directly outputing times as predictions, which takes the input prompt $p_\mathrm{query} =$ \textit{"Estimate~the~LOCAL~capture~clock~time~of~the~image,~answer~with~ONLY~one~24-hour~time~in~HH:MM~format."}.}

\begin{figure*}[t]
    \centering
    \begin{subfigure}[b]{0.48\linewidth}
        \centering
        \includegraphics[width=\linewidth]{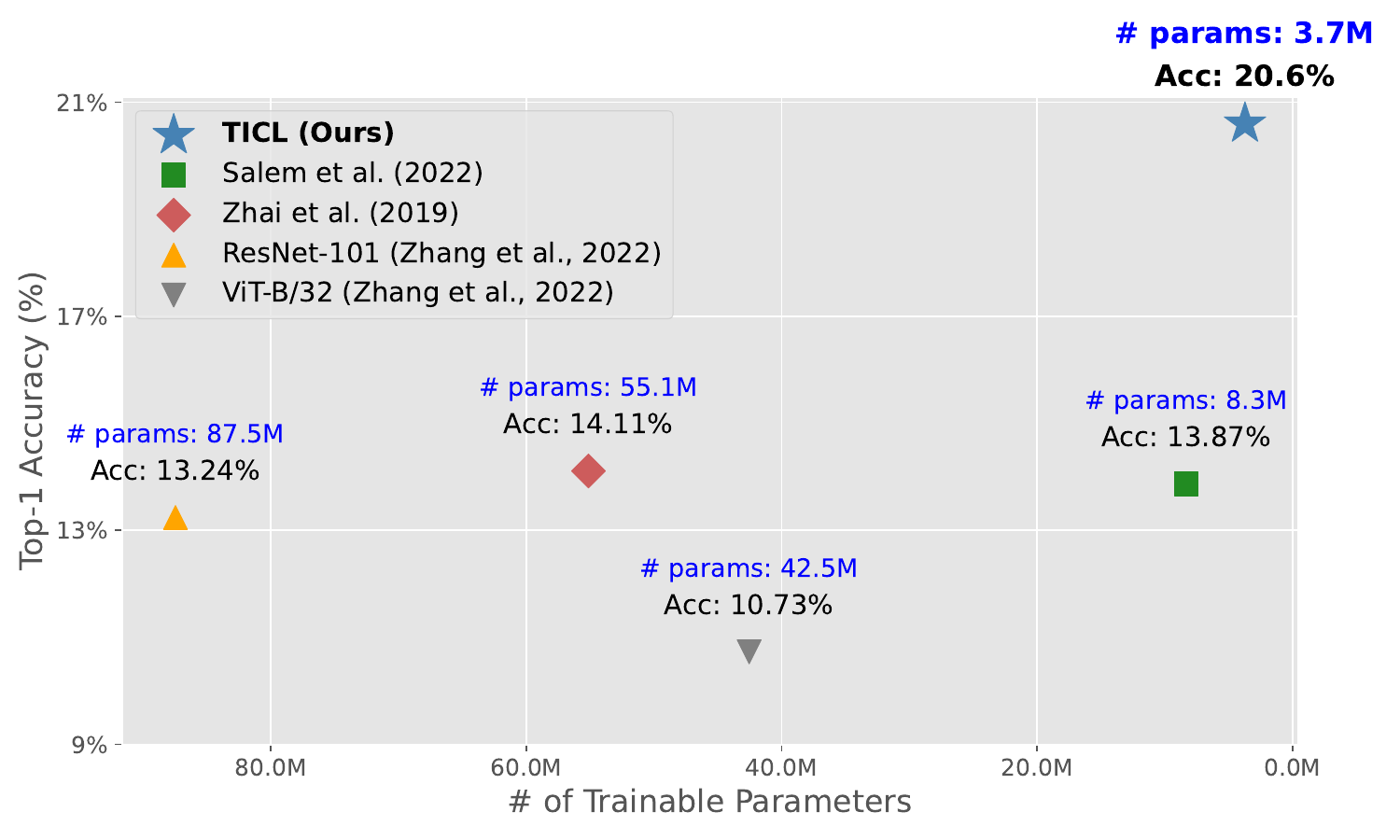}
        \caption{Trainable parameters and performance comparisons.}
        \label{fig:trainable_params_vs_accuracy}
    \end{subfigure}
    \hfill
    \begin{subfigure}[b]{0.48\linewidth}
        \centering
        \includegraphics[width=\linewidth]{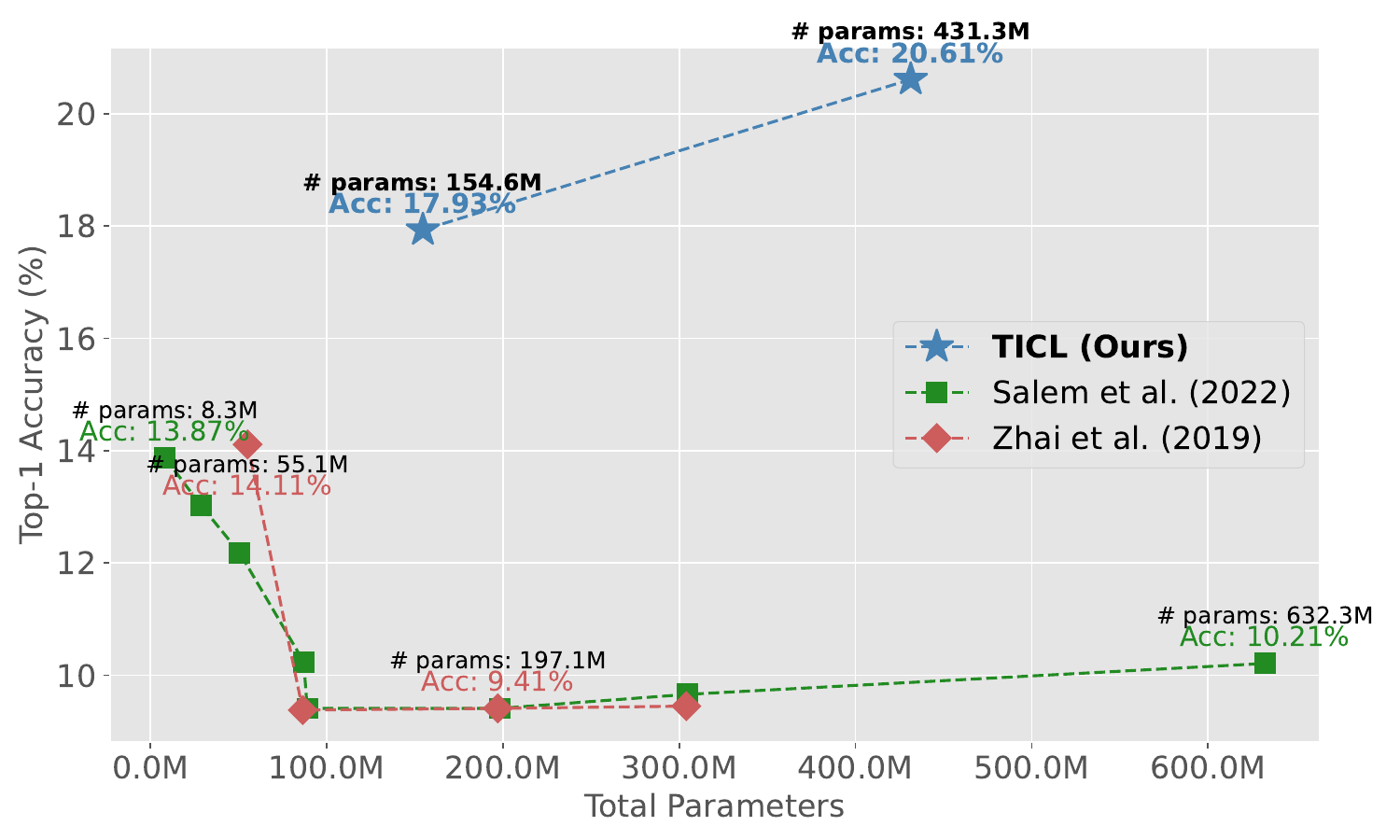}
        \caption{Total parameters and performance comparisons.}
        \label{fig:total_params}
    \end{subfigure}
    \caption{\textbf{Parameter efficiency and performance.} (a) Comparison of trainable parameters and performance. (b) Analysis of total parameters and performance.}
    \label{fig:parameter_efficiency}
\end{figure*}
\subsection{Computational efficiency}

Since the majority part of the TICL model, the CLIP image encoder, is frozen during training, the TICL training is thus efficient with a small number of trainable parameters. \cref{fig:trainable_params_vs_accuracy,fig:total_params} shows that TICL achieved the best performance with the minimum trainable parameters among existing methods. Benefiting from the fewer trainable parameters, training on precomputed image features is significantly faster. Also, \cref{fig:total_params} demonstrates that simply scaling up the model parameters for previous works may even degrade the performance. We suspect that it is due to the more severe overfitting of the larger models on training samples. In comparison, the TICL model reached better performance with a moderate total number of parameters.

\begin{table*}[t]
\centering
\caption{\textbf{Joint time estimation performance on our TOC dataset.} Namely, we jointly estimate the month and hour using the same setup in the previous baseline \citet{zhai2019learning} for fair comparison. }
\label{tab:joint_predition}
\resizebox{\textwidth}{!}{%
\begin{threeparttable}
\begin{tabular}{l|cccc|cccc}
\toprule
                           & \multicolumn{4}{c|}{\textbf{Hour Prediction}}           & \multicolumn{4}{c}{\textbf{Month Prediction}}            \\
                           & Top-1 acc {$\uparrow$} & Top-3 acc {$\uparrow$} & Top-5 acc {$\uparrow$} & Time MAE (min.) {$\downarrow$} & Top-1 acc & Top-3 acc & Top-5 acc & Month MAE \\
\midrule
\citet{salem2022timestamp} & 13.87\%      & 39.36\%      & 60.71\%      &  186.44      & 7.40\%       & 25.74\%      & 42.93\%      & 3.14      \\
\citet{zhai2019learning}  & 14.11\%      & 40.47\%      & 65.94\%      & 188.78      & 11.23\%       & 33.03\%      & 55.16\%      & 2.38      \\
\citet{salem2022timestamp}\tnote{$\dagger$} & 13.53\%      & 38.47\%      & 59.10\%      & 176.70      & 9.59\%       & 24.56\%      & 39.61\%      & 2.74      \\
\citet{zhai2019learning}\tnote{$\dagger$} & 15.01\%      & 42.54\%      & 68.24\% & 185.34      & 12.03\%       & 35.91\%      & 60.50\%      & 2.25      \\
TICL (Hour only) & \textbf{20.60\%}      & \textbf{49.01\%} & \textbf{67.82\%}      & 171.65 & - & - & - & - \\
TICL (Month only)\tnote{$\ddagger$} & - & - & - & - & \textbf{34.48\%} & \textbf{68.19\%} & \textbf{82.88\%} &  \textbf{1.45} \\
TICL (Month, Hour) & 19.45\% & 42.07\% & 55.57\% & 176.45 & 32.28\% & 52.00\% & 62.26\% & 1.77\\
\midrule
                           & \multicolumn{4}{c|}{\Update{\textbf{Hour Prediction}}}           & \multicolumn{4}{c}{\Update{\textbf{Season Prediction}}} 
                            \\
                           & \Update{Top-1 acc {$\uparrow$}} & \Update{Top-3 acc {$\uparrow$}} & \Update{Top-5 acc {$\uparrow$}} & \Update{{Time MAE (min.)} {$\downarrow$}} & \Update{Top-1 acc} & \Update{Top-3 acc} & \Update{Top-5 acc} & - \\
\midrule
\Update{TICL (Season, Hour)} & \Update{20.14\%} & \Update{45.98\%} & \Update{62.58\%} & \Update{\textbf{170.93}} & \Update{61.52\%} & \Update{71.74\%} & \Update{80.48\%} & - \\
\bottomrule
\end{tabular}%
\begin{tablenotes}
\footnotesize
\item[$\dagger$] These baselines take additional known geolocation metadata inputs, which boosted their performances on both prediction tasks.
\item[$\ddagger$] Predicting 12 classes for months.
\end{tablenotes}
\end{threeparttable}
}

\end{table*}

\subsection{Joint metadata estimation with time}
\label{sec:joint}

We noticed that some of the previous baselines support joint time estimation instead of just focusing on clock time only. They often consider the joint contribution of other metadata including geolocation, date, and time of day to the image appearances \citep{salem2022timestamp, zhai2019learning} to deal with the ambiguity of clock time when given only visual inputs. Therefore, in this section, we aim to explore such capability of estimating time and month jointly from only social media images.

\begin{wrapfigure}{r}{0.4\linewidth}
    \centering
    \includegraphics[width=\linewidth]{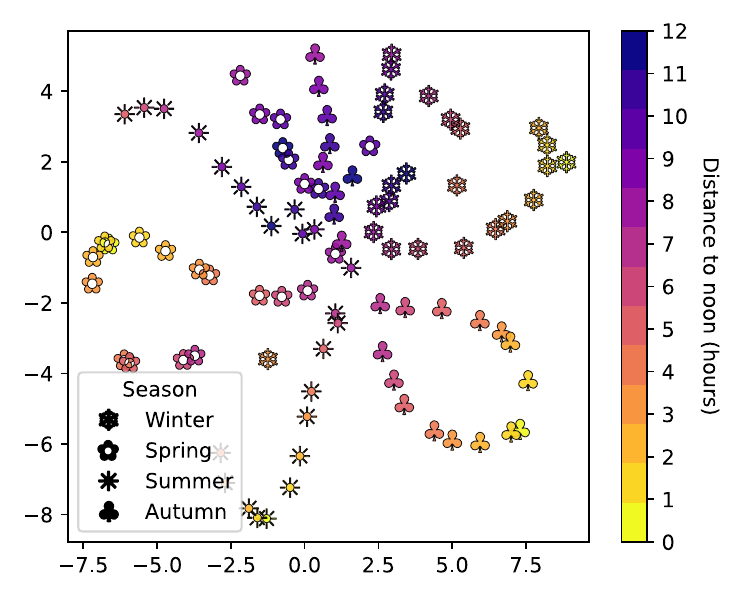}
    \caption{\Update{t-SNE visualisation of class embeddings \texttt{<clock time, season>}.}}
    \label{fig:season_time}
\end{wrapfigure}

We adjusted the network structure of TICL, enabling its ability to estimate month and hour jointly using a similar structure to \citet{zhai2019learning}, in which the model predicts $12 \times 24$ classes combining months and hours. We kept all the specified hyper-parameters and other setups the same. As for the compared baseline methods, we used the same hyper-parameters provided in previous works and picked the best performances from several trials, all the models are trained and tested on TOC dataset which contains only social media data to demonstrate the challenges on real-world samples. {As provided in \cref{tab:joint_predition}, TICL generally outperforms previous baselines when trained an tested on the more challenging TOC dataset without images with fixed views. In addition, under TICL paradigm, jointly predicting clock time and month \Update{or season} does not provide boosts to individual tasks. We suspect that it is because of the gaps between the visual cues between the two different target variables. Such gaps lead to difficulties to model a joint probability of $P(t, m|x)$ for clock time $t$ and month $m$ with only the input $x$. However, the prediction advantages of models focusing on each attributes only suggest the possibility of stacking such different metadata-aware models in joint metadata verification-related tasks focusing on $P(t|m, x), P(m|t, x)$.}

\Update{However, compared with using the date as additional supervision, the season supervision combines both date and geolocation gives a much smaller performance gap to the purely clock time supervised model. This suggests that learning meaningful combinations of other metadata could possibly provide a better clue to normalise the visual variance of the clock time. To support this point, we have included a visualization of learned clock time and season in terms of embedding in \cref{fig:season_time}. As visualised, the model clearly recognizes season variations of the same clock time, while keeping a general pattern for different time periods of the day. This shows the potential of producing a possible calibration mechanism by learning on unnormalised timestamps with the help of different metadata supervision. This indicate a hopeful direction that we may learn an calibration mechanism to the raw clock time to bio-clock by incorporating comprehensive metadata and scene priors as supervision.}

\section{Exploration of More Precise Time Encoding}
\label{sec:precision_limits}
\subsection{Scalar encoding}
\label{sec:regression_limits}
\begin{figure*}[ht]
    \centering
    \begin{subfigure}[b]{0.45\linewidth}
        \includegraphics[width=\linewidth]{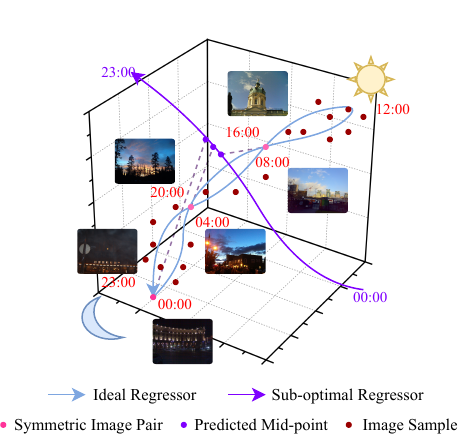}
        \caption{}
        \label{fig:regr}
    \end{subfigure}
    \begin{subfigure}[b]{0.42\linewidth}
        \includegraphics[width=\linewidth]{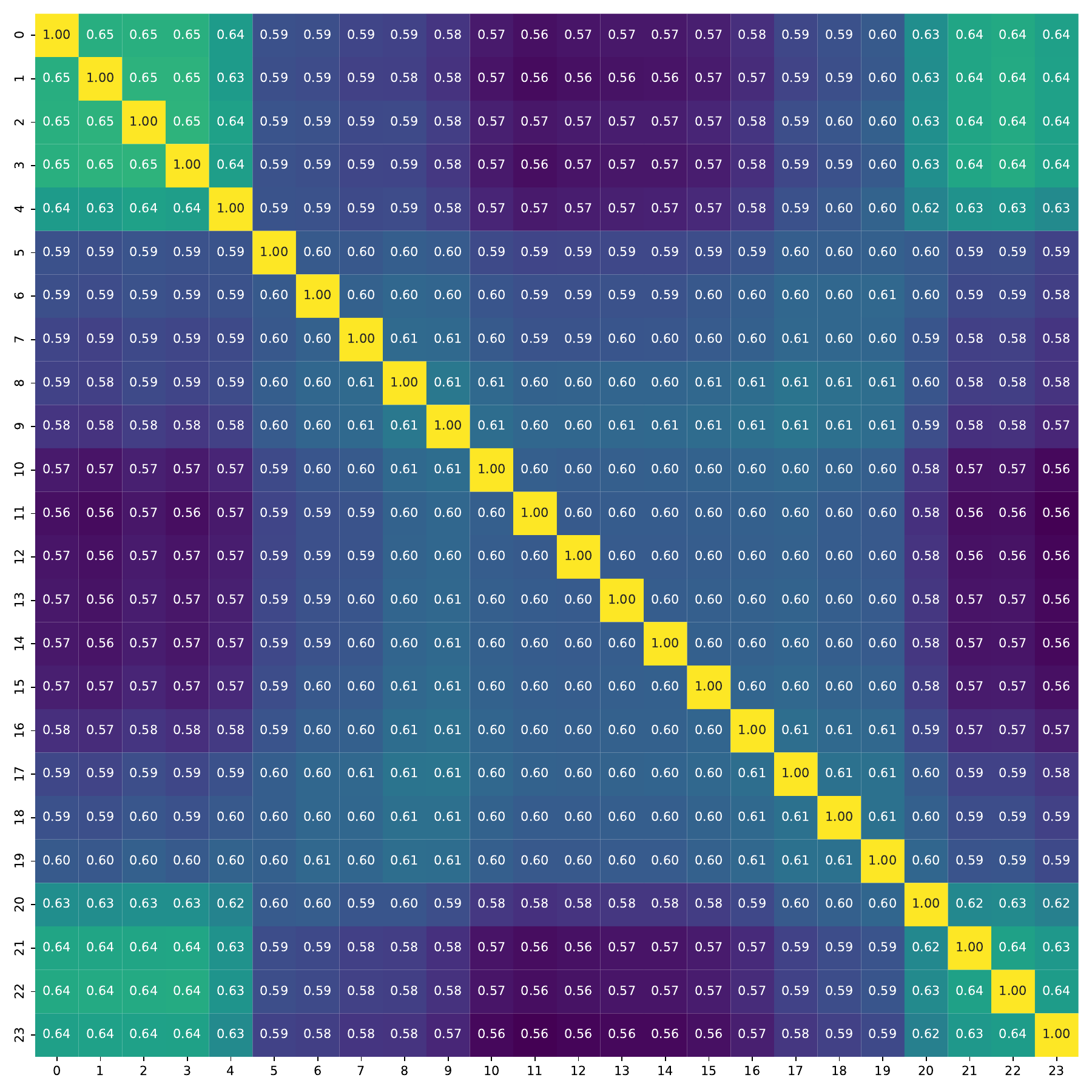}
        \caption{}
        \label{fig:cossim_resnet18}
    \end{subfigure}
    \\
    \centering
    \begin{subfigure}[b]{0.45\linewidth}
        \includegraphics[width=\linewidth]{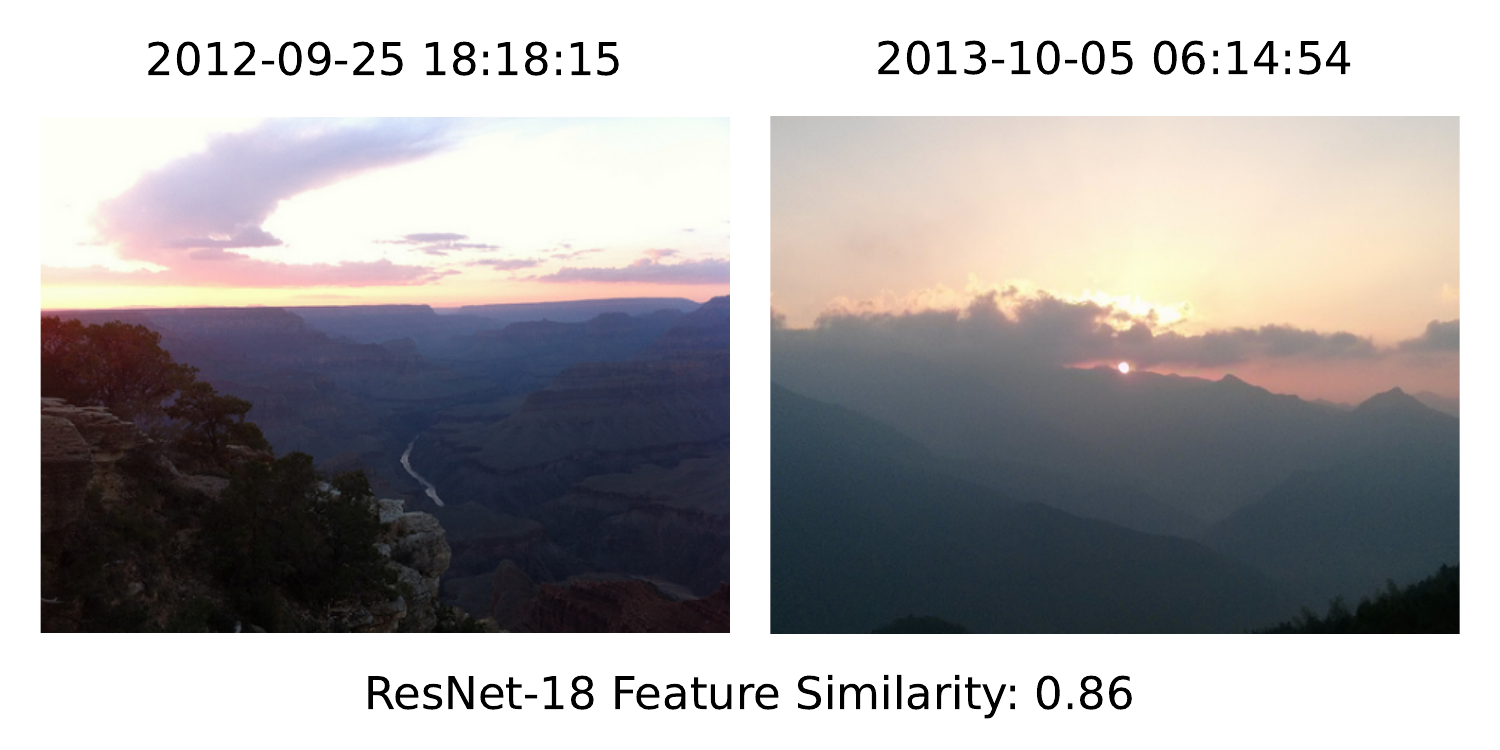}
        \caption{}
        \label{fig:sunset_sim}
    \end{subfigure}
    \begin{subfigure}[b]{0.45\linewidth}
        \includegraphics[width=\linewidth]{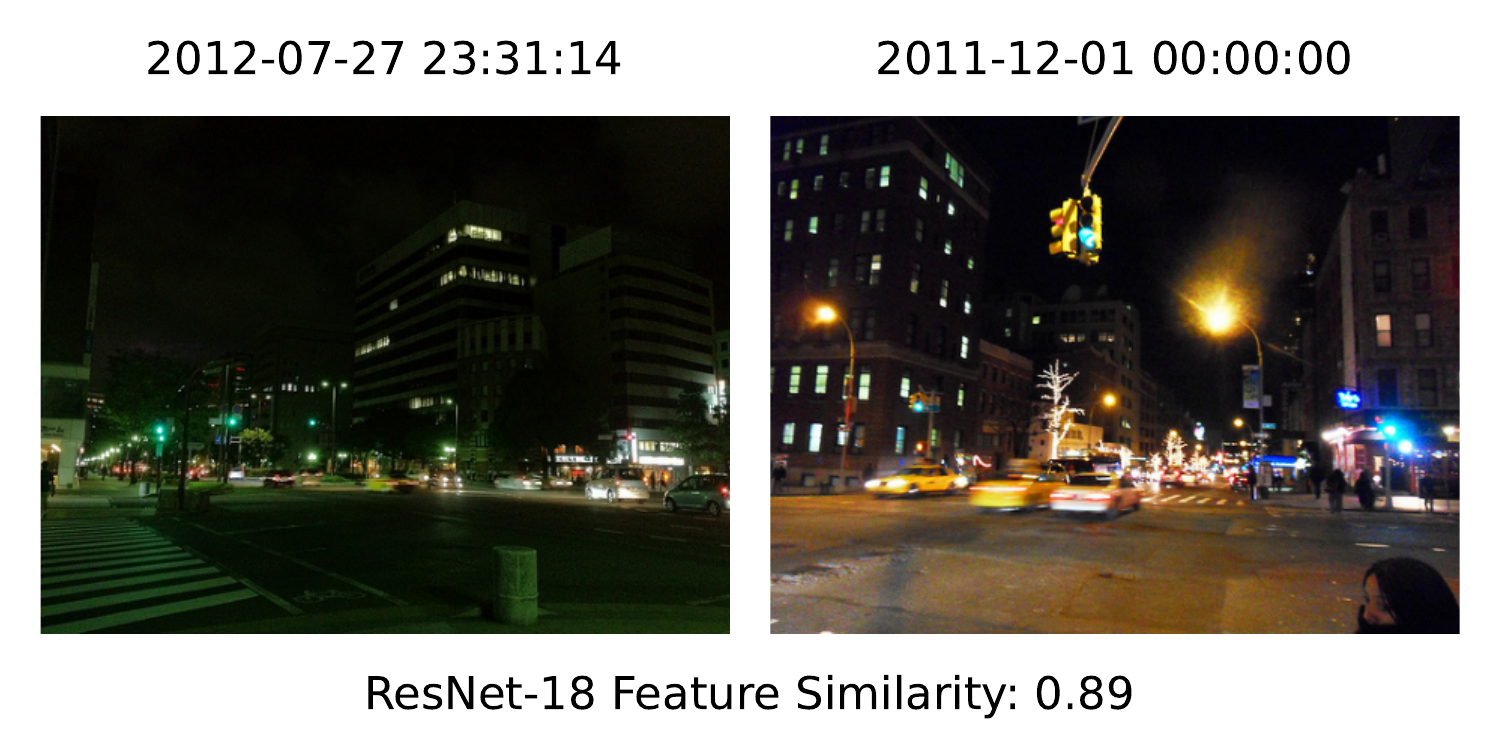}
        \caption{}
        \label{fig:night_sim}
    \end{subfigure}
    \caption{\textbf{Visual ambiguities for ground truth in regression.} (a) depicts a sub-optimal regression model where the predictions are biased towards the mid-point, and (b) shows a trend that images with more similar ResNet-18 features could have disparate timestamps. Few examples of such cases are provided in (c), (d).}
    \label{fig:comparison}
\end{figure*}

In this section, we explore limitations in a simple regression solution to the pre-text estimation task using scalar encoding of the clock time.  

\paragraph{Raw scalar encoding: } The regression style construction for clock time estimation from images presents significant challenges as covered in main text. There are different issues with regression models, including 1) loss function sensitivity and 2) discontinuity in the scalar range for regression. In the following paragraphs, we first provide a brief illustration of the issue on the regression loss function. Secondly, we present experiments of a regression model working in a circular space instead of the vanilla scalar range which is a disconnected set \citep{Zhou_2019_CVPR}. These experiments provide explanations for the limits of vanilla regression models.

\begin{figure*}[h!]
    \centering
    \begin{subfigure}[b]{0.52\linewidth}
        \centering
        \includegraphics[width=\linewidth]{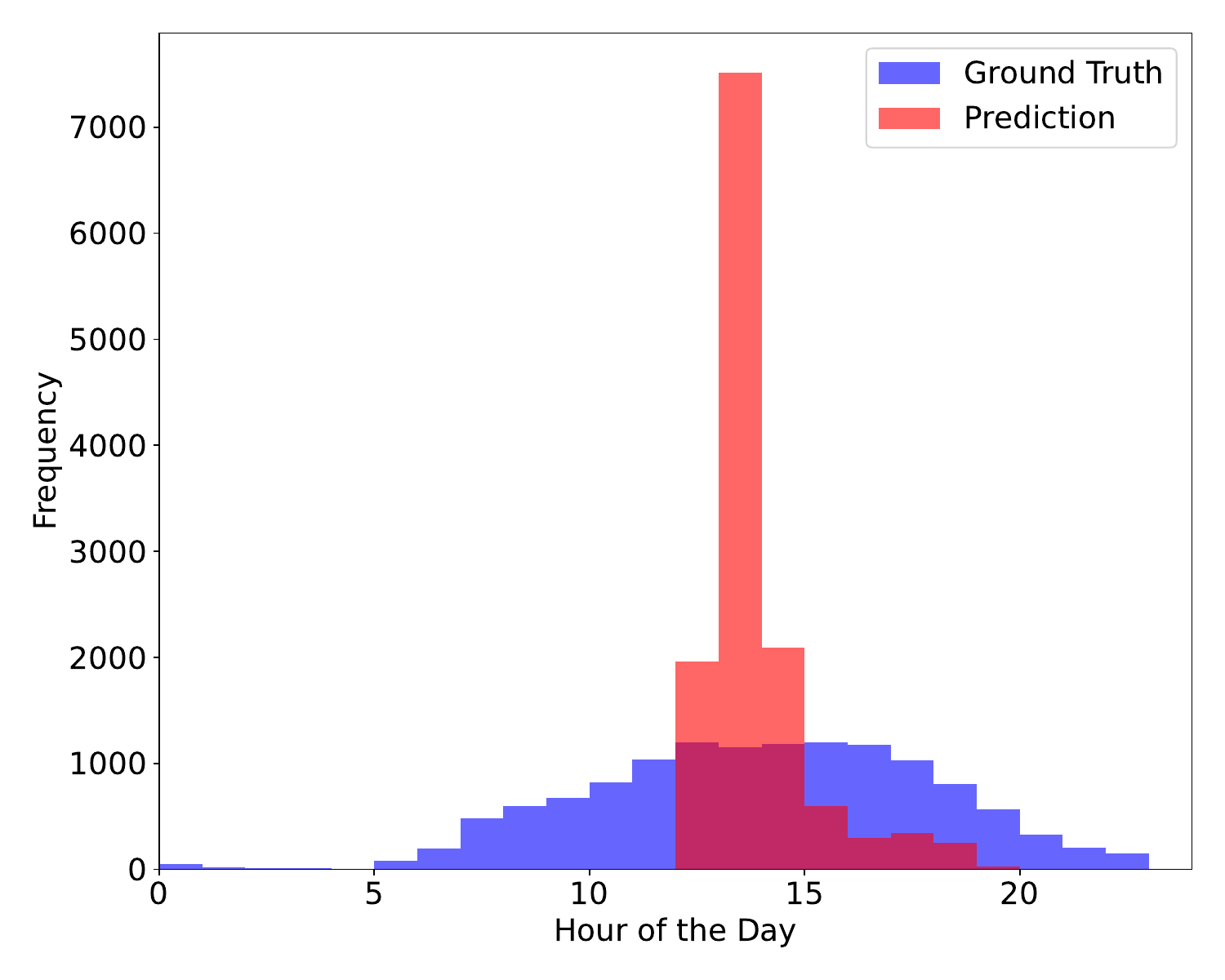}
        \caption{}
        \label{fig:regressor_prediction}
    \end{subfigure}
    \begin{subfigure}[b]{0.47\linewidth}
        \centering
        \includegraphics[width=\linewidth]{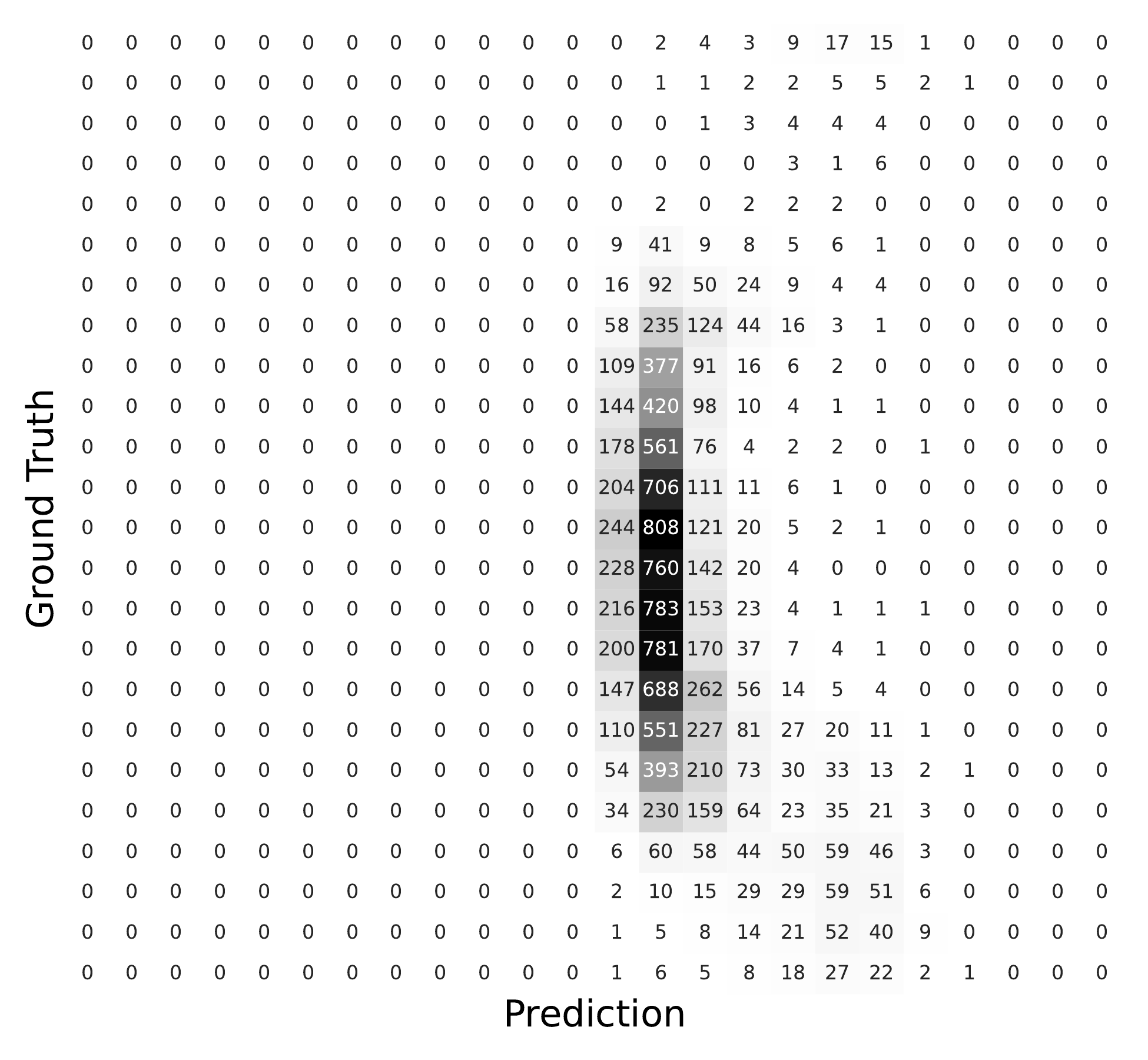}
        \caption{}
        \label{fig:regressor_conf}
    \end{subfigure}
    
    \centering
    \begin{subfigure}[b]{0.46\linewidth}
        \centering
        \includegraphics[width=\linewidth]{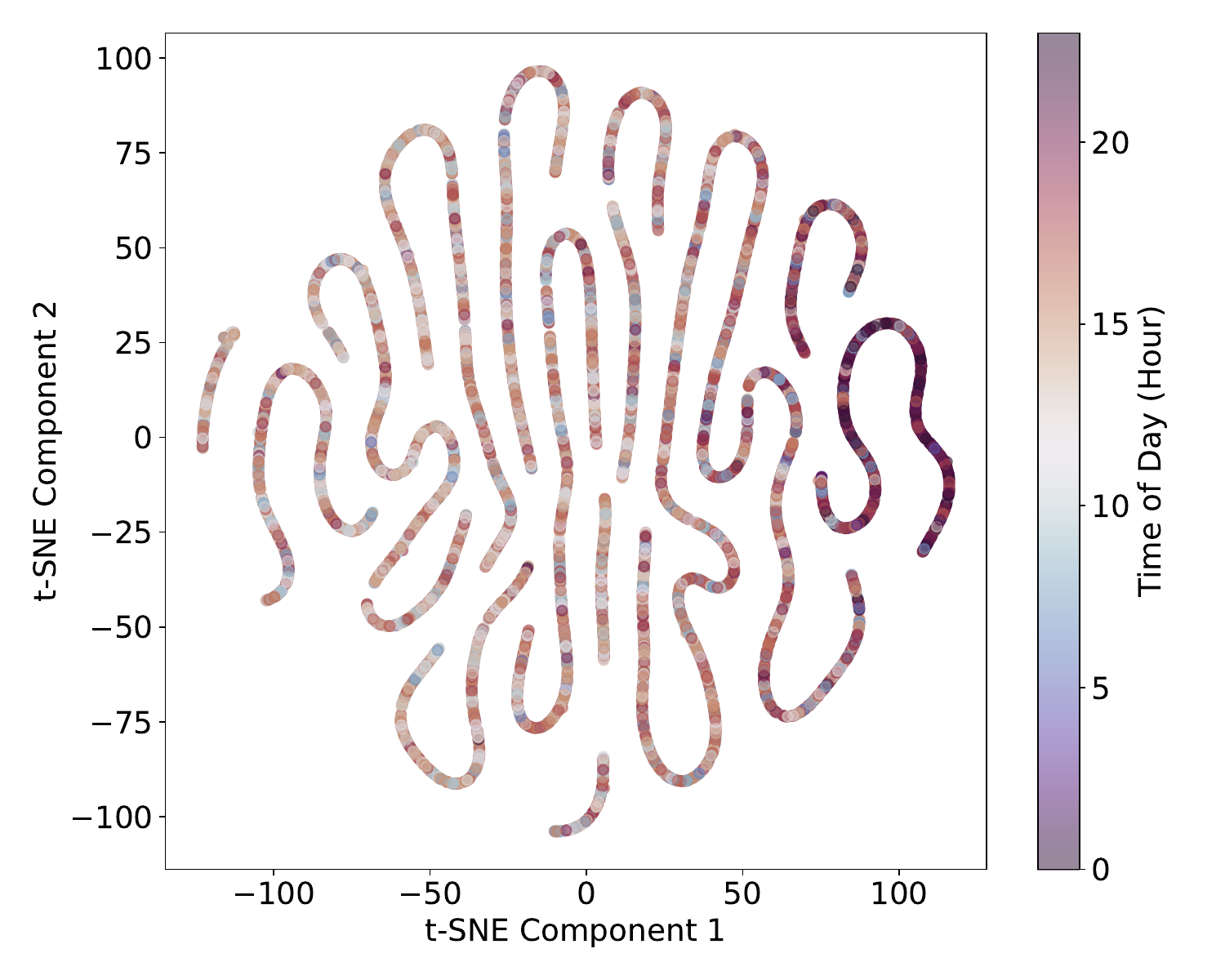}
        \caption{}
        \label{fig:regr_rep}
    \end{subfigure}
        \centering
    \begin{subfigure}[b]{0.46\linewidth}
        \includegraphics[width=\linewidth]{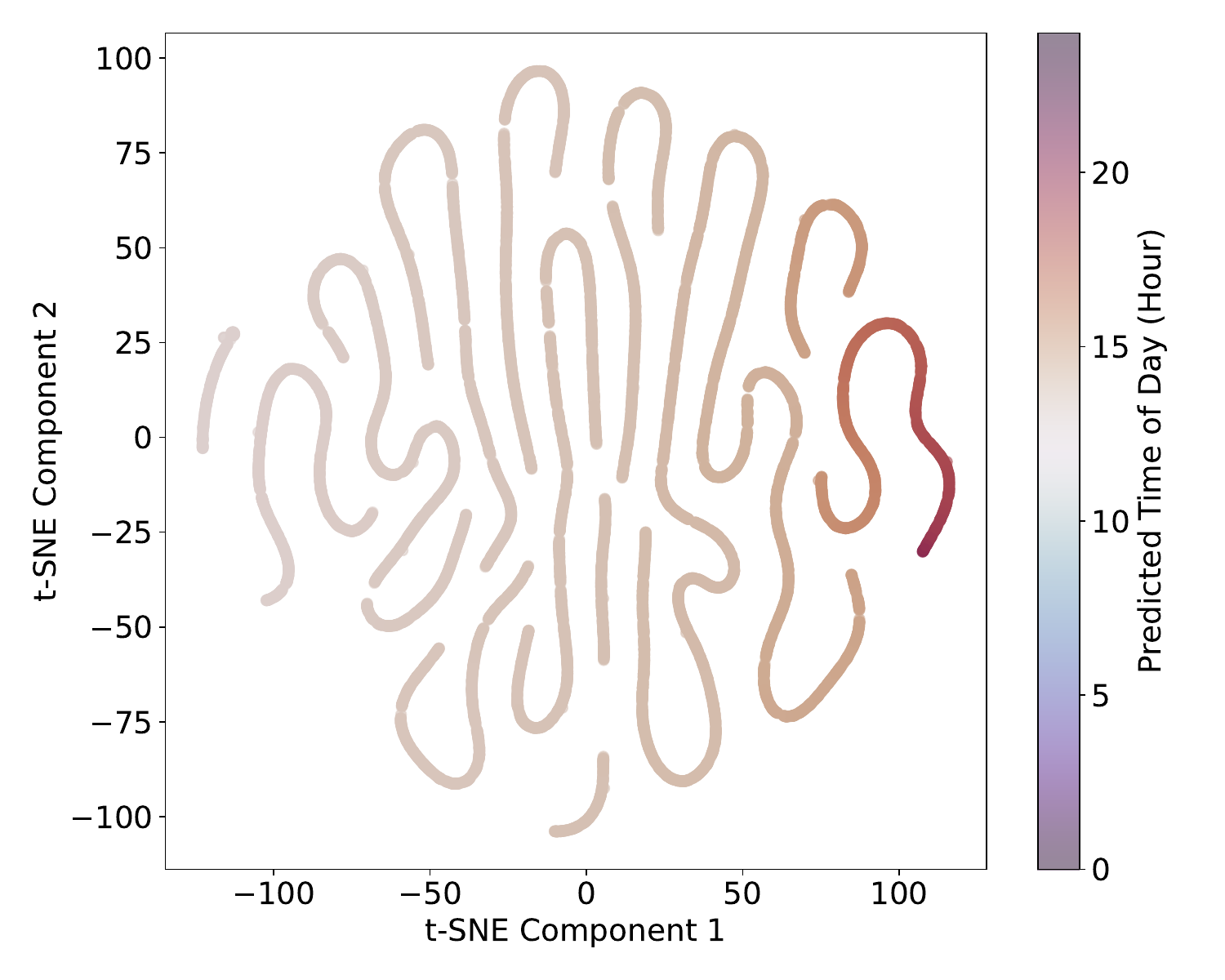}
        \caption{}
        \label{fig:pred_regr_rep}
    \end{subfigure}

    \caption{\textbf{Experiments on regression model.} (a) shows prediction distribution of regression model on TOC test set, (b) represents the confusion matrix by hour, (c) and (d) visualise t-SNE of regressor representations annotated with ground truth and predicted timestamps, respectively.}
\end{figure*}
\begin{figure*}[h!]
    \centering
    \begin{subfigure}[b]{0.50\linewidth}
        \centering
        \includegraphics[width=\linewidth]{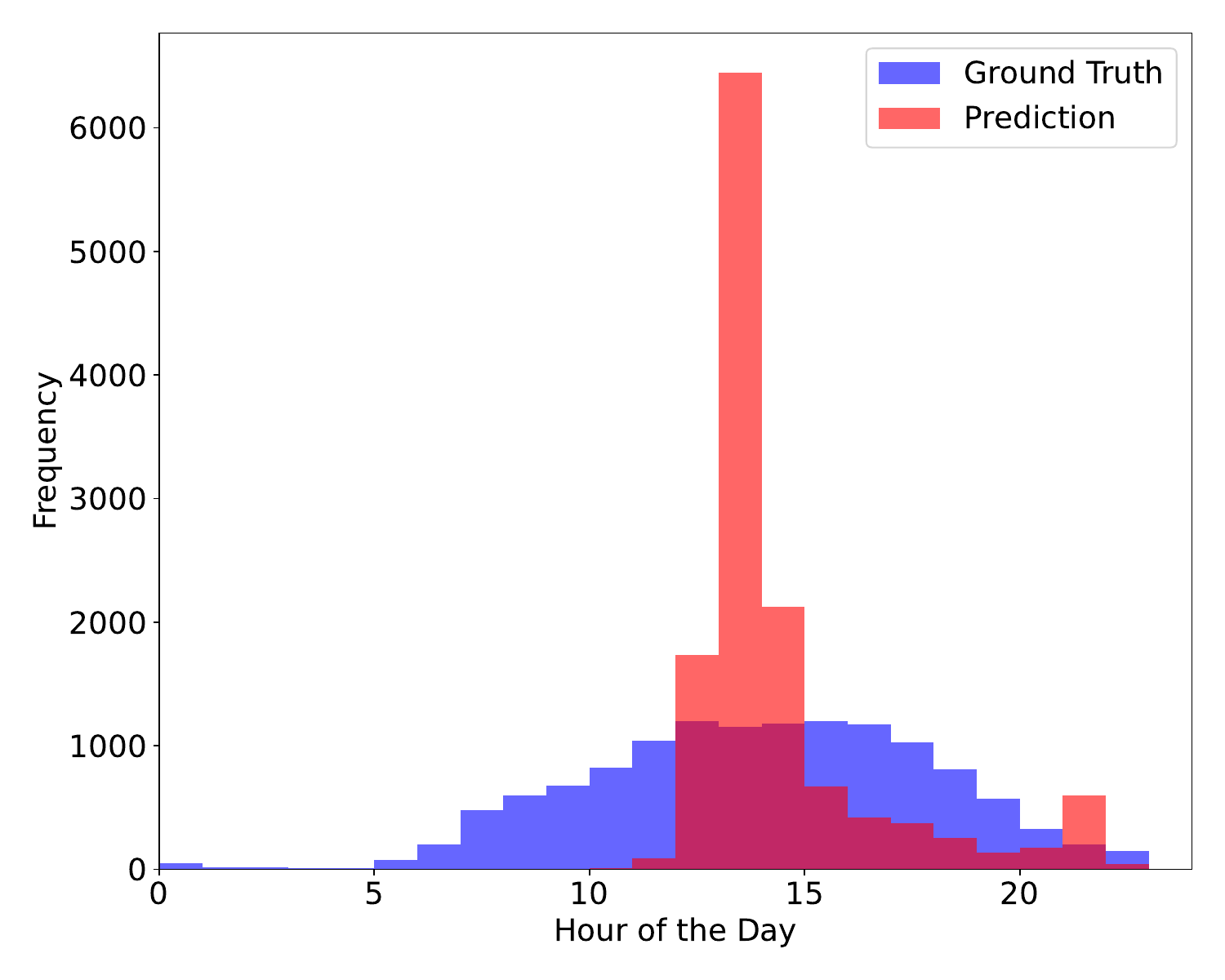}
        \caption{Prediction distribution of cyclic regression}
        \label{fig:regressor_prediction_cyclic}
    \end{subfigure}
    \begin{subfigure}[b]{0.45\linewidth}
        \centering
        \includegraphics[width=\linewidth]{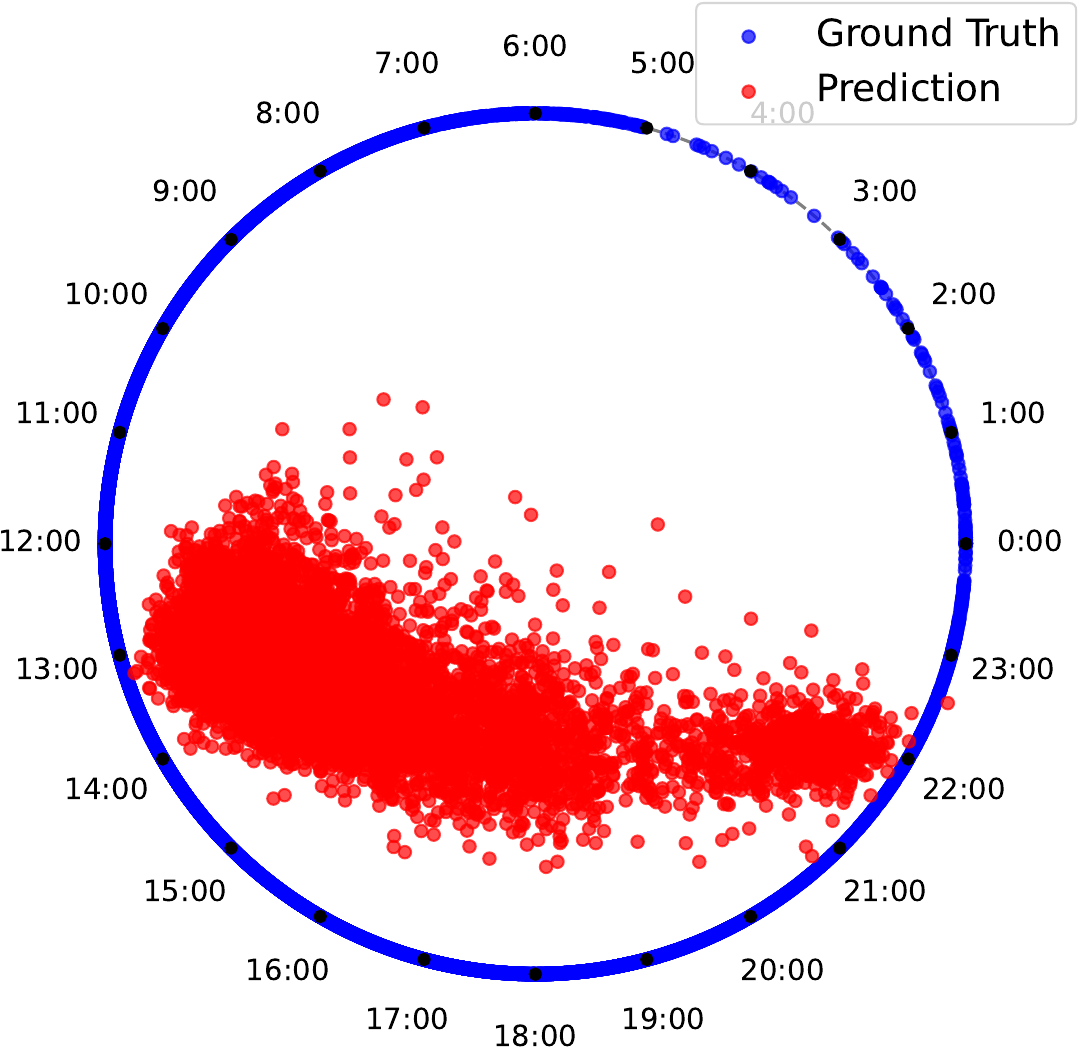}
        \caption{Cyclic encoded labels and predictions}
        \label{fig:cyclic_encoding}
    \end{subfigure}
    \caption{\textbf{Cyclic regression model results,} (a) shows prediction distribution of cyclic regression model, and (b) visualise how the cyclic encoding of predictions differ from the ground truth.}
    \label{fig:cyclic_regression}
\end{figure*}

Let us define the problem setting of clock timestamp regression as follows. Given an image $x$, the objective is to predict the timestamp $y$ in the range $[0, 24)$ hours of the day. In a regression framework, the model $f_\theta$ maps an input image $x$ to a continuous scalar output $\hat{y} = f_\theta(x) \in [0, 24)$.

Consider a dataset $\mathcal{D}$ consisting of images taken at various times throughout the day. Specifically, consider pairs of images $\{(x_i, y_i), (x_j, y_j)\}$ taken during ``symmetric times'' such as sunrise and sunset, where the general light conditions are similar but the ground truth timestamps are different (see \cref{fig:night_sim} and~\ref{fig:sunset_sim}). With very similar inputs and the same model $f_\theta(\cdot)$, it holds that:
\[
f_\theta{(x_i)} \approx f_\theta{(x_j)}
\]
Then the Mean Squared Error (MSE) loss for the regression model over the dataset is defined as:
\[
\mathcal{L}_{\text{MSE}}(\theta) = \frac{1}{|\mathcal{D}|} \sum_{k=0}^{|\mathcal{D}|-1} (y_k - f_\theta(x_k))^2
\]

To find the optimal model parameters $\theta^*$, we minimise this loss function. Ideally, the  goal of the optimiser is:
\[
\nabla_\theta \mathcal{L}(\theta) = 0
\]

For pairs of similar images with different $y$, this optimisation leads to mid-point predictions:
\[
\hat{y}_i \approx \hat{y}_j \approx \frac{y_i + y_j}{2}
\]

This effect leads to local minima in the clock timestamp embedding space in \cref{fig:regr}, particularly when $y_i$ and $y_j$ are at opposite ends of the 24-hour cycle, for example, 00:00 and 23:59. The regression model struggles with the ambiguous nature of time, resulting in systematically biased predictions towards the midpoint of symmetric clock times. Such bias results in incorrect gradient updates that cannot lead to an accurate estimation model for inputs $x_i, x_j$.

The aforementioned phenomenon of similar images with disparate ground truth timestamps prevails in the dataset. As evidence, we visualise the similarity of features using the ResNet-18 backbone throughout hours for the entire dataset in \cref{fig:cossim_resnet18}. Therefore, this overall trend of feature similarity extends the reasoning to the entire dataset, where the predictions $\hat{y}$ are systematically biased towards the average of the whole clock time distribution. The predictions are likely to follow the normal distribution with the same mean value to the ground truth distribution and smaller variance $\sigma$ \citep{murphy2012machine}. 
\[
\hat{y} \sim \mathcal{N}\left(\frac{1}{|\mathcal{D}|} \sum_{k = 0}^{|\mathcal{D}|-1} y_k, \sigma\right)
\]

We conduct corresponding experiments to provide evidence for the claims above. Particularly, we train a regression model using ResNet-101 backbone. The prediction histogram and confusion matrix provided in \cref{fig:regressor_prediction} and \cref{fig:regressor_conf} support our claims. The predictions are heavily concentrated around the average value of the ground truth distribution, while the actual timestamps in the dataset are more evenly distributed throughout the day. This discrepancy highlights the failure of the regression model to capture the cyclic nature of time, resulting in biased predictions of the average of the whole range. \cref{fig:regr_rep} shows that the regression model fails to discern similar images with different timestamps, where the features form disjoint trails on which images features from totally different time periods are nearly overlapped with each other. \cref{fig:pred_regr_rep} further shows how the regression model predicts average timestamps for these images with similar features. These phenomena show that although the regression model managed to learn a certain extent of continuity of time of day from static views, it failed to tackle the ambiguity of clock timestamp given visual inputs with similar illuminations. Therefore, while such a regression model reaches convergence at local minima for the MSE loss, it is not ideal resorts we are looking for.
\begin{figure*}[t]
    \centering
    \begin{subfigure}[b]{0.325\textwidth}
        \centering
        \includegraphics[width=\linewidth]{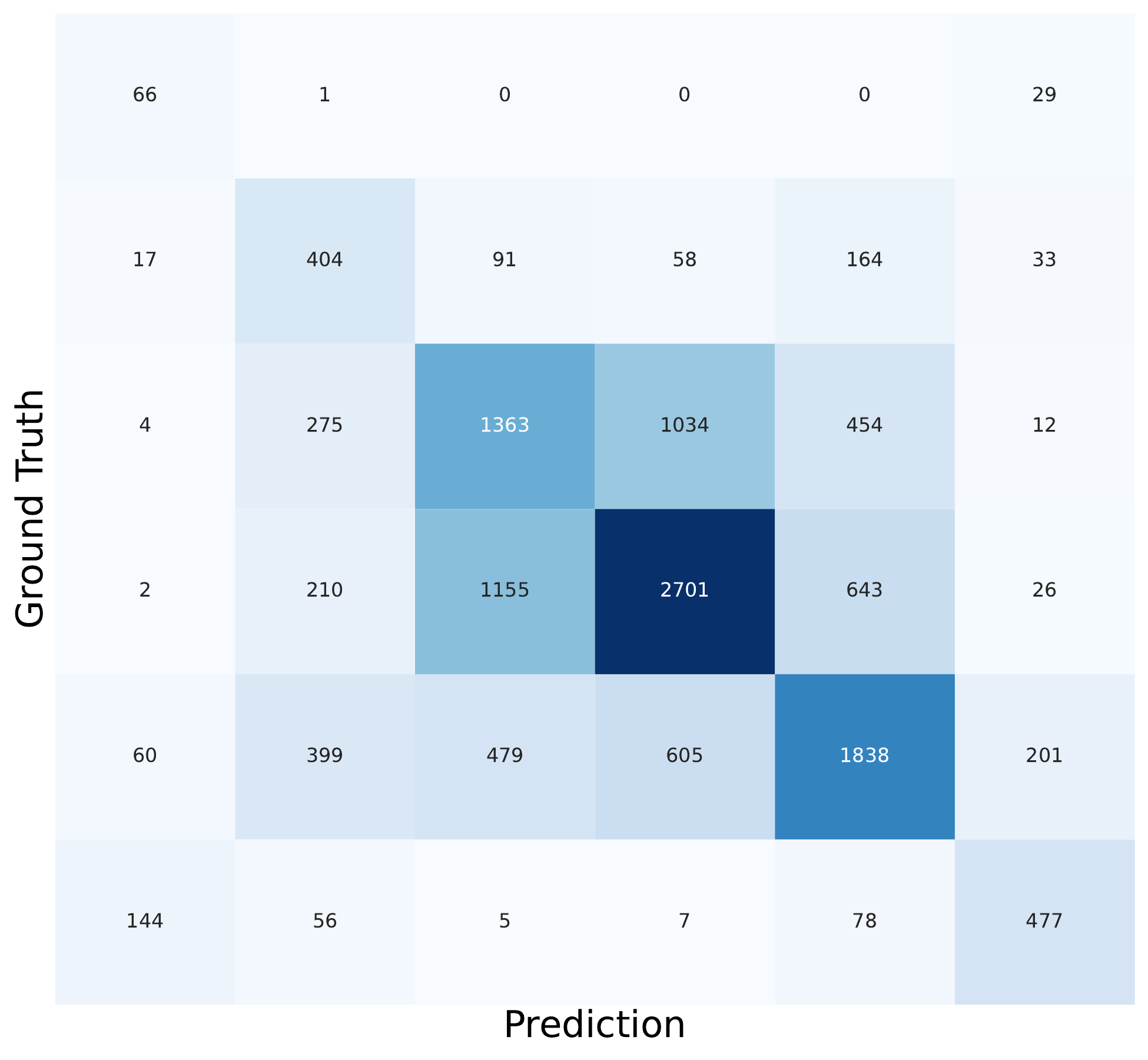}
        \caption{6 class partitioning}
        \label{fig:conf_k_1}
    \end{subfigure}
    \hfill
    \begin{subfigure}[b]{0.3\textwidth}
        \centering
        \includegraphics[width=\linewidth]{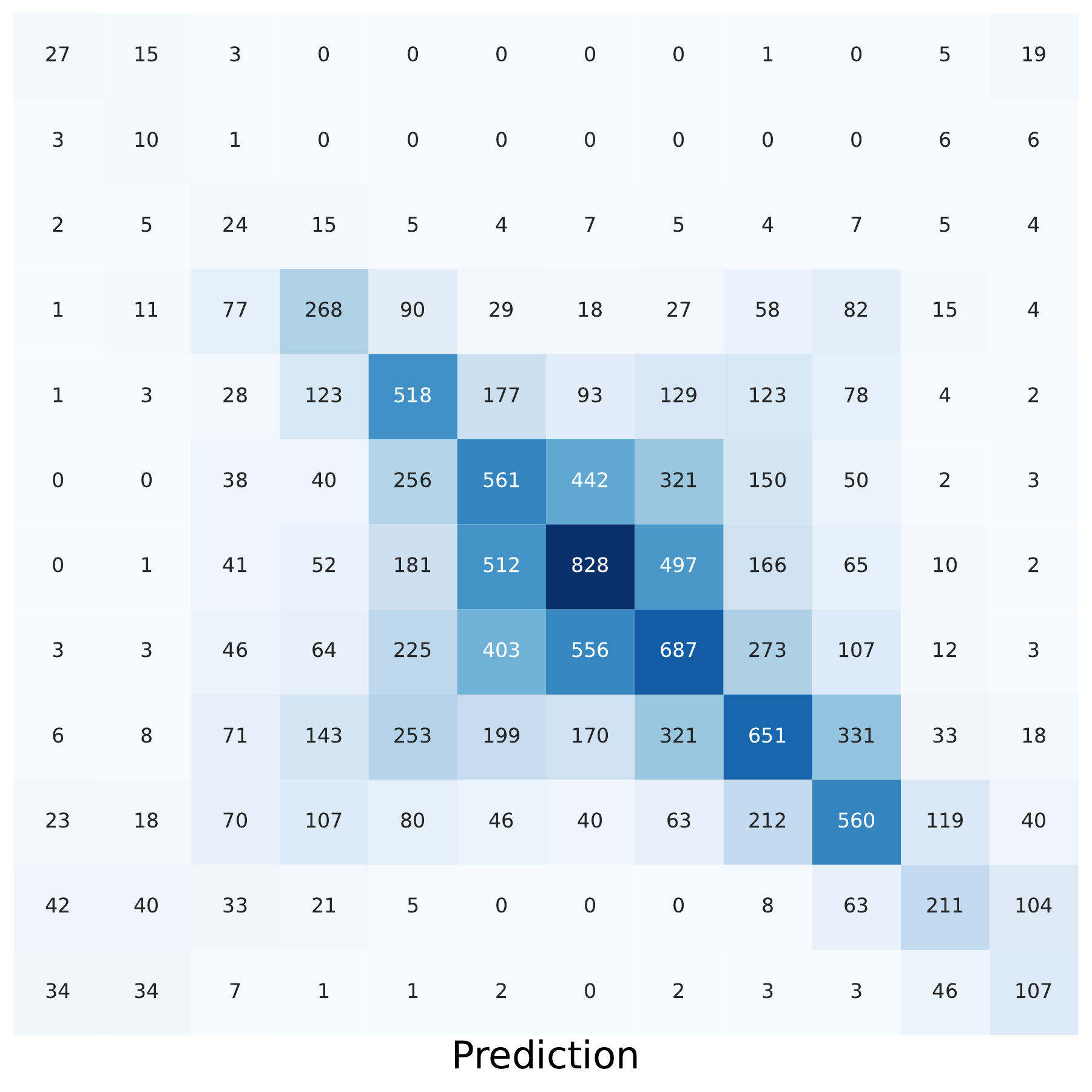}
        \caption{12 class partitioning}
        \label{fig:conf_k_2}
    \end{subfigure}
    \hfill
    \begin{subfigure}[b]{0.3\textwidth}
        \centering
        \includegraphics[width=\linewidth]{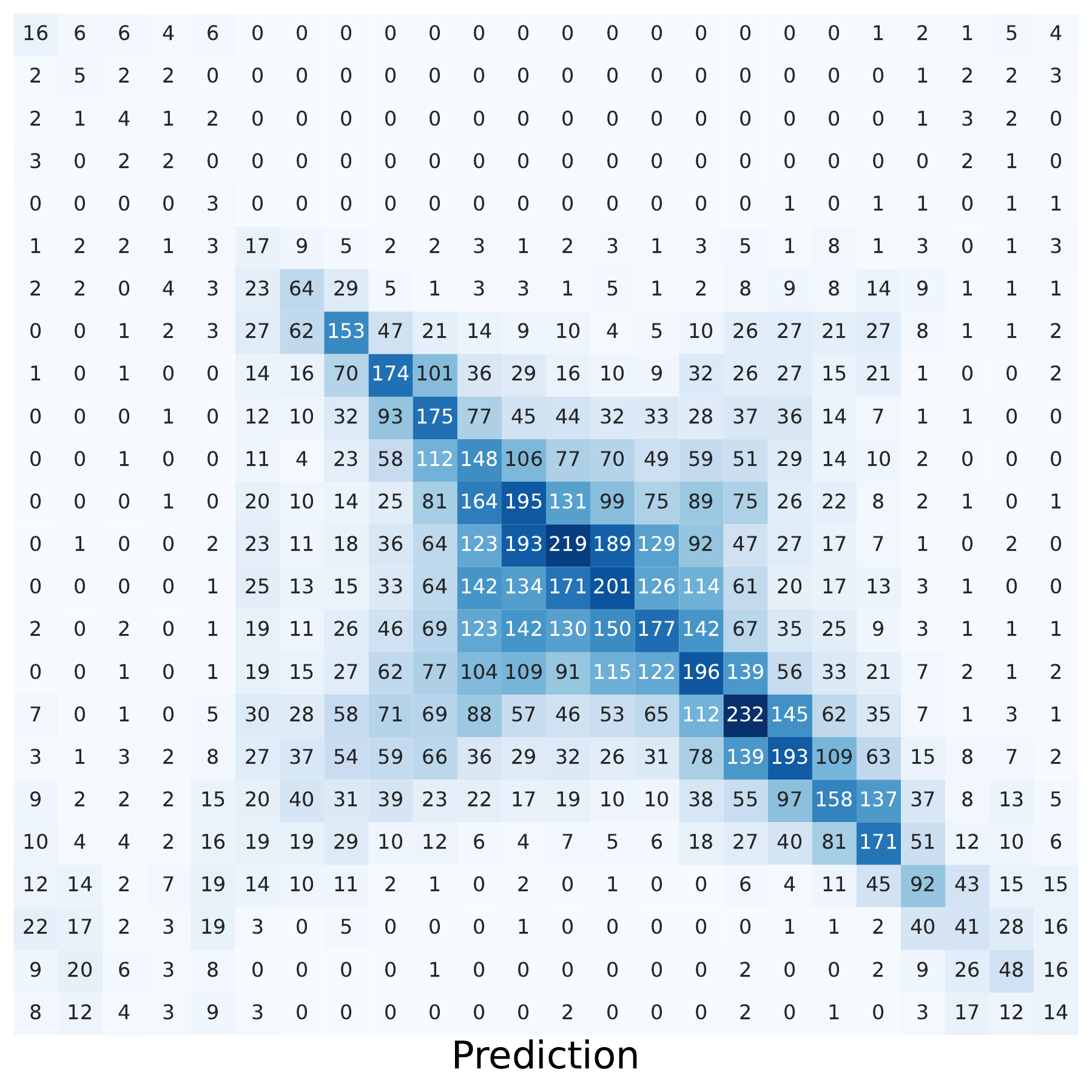}
        \caption{24 class partitioning}
        \label{fig:conf_k_3}
    \end{subfigure}
    \vfill
    \begin{subfigure}[b]{0.325\textwidth}
        \centering
        \includegraphics[width=\linewidth]{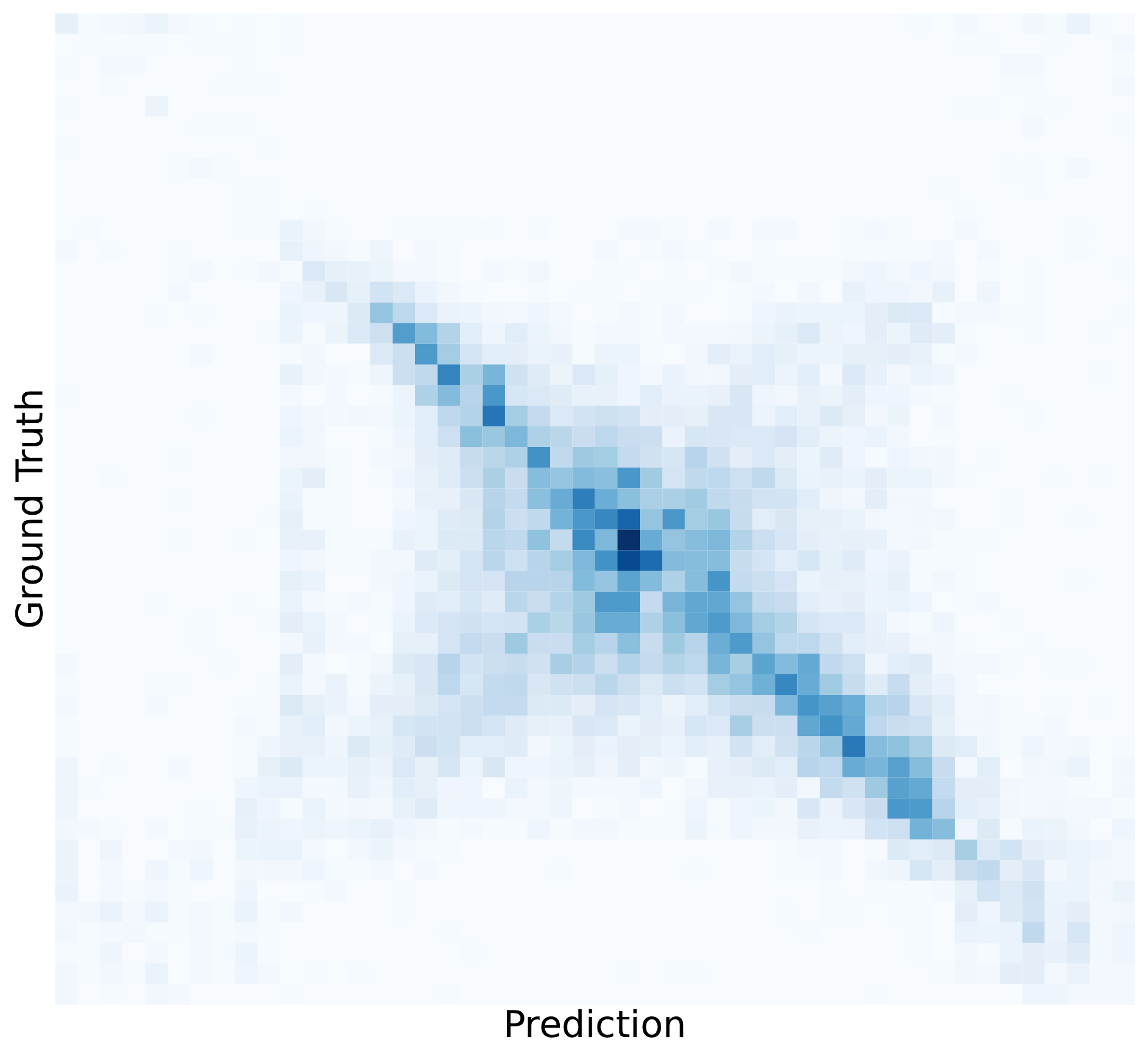}
        \caption{48 class partitioning}
        \label{fig:conf_k_4}
    \end{subfigure}
    \hfill
    \begin{subfigure}[b]{0.3\textwidth}
        \centering
        \includegraphics[width=\linewidth]{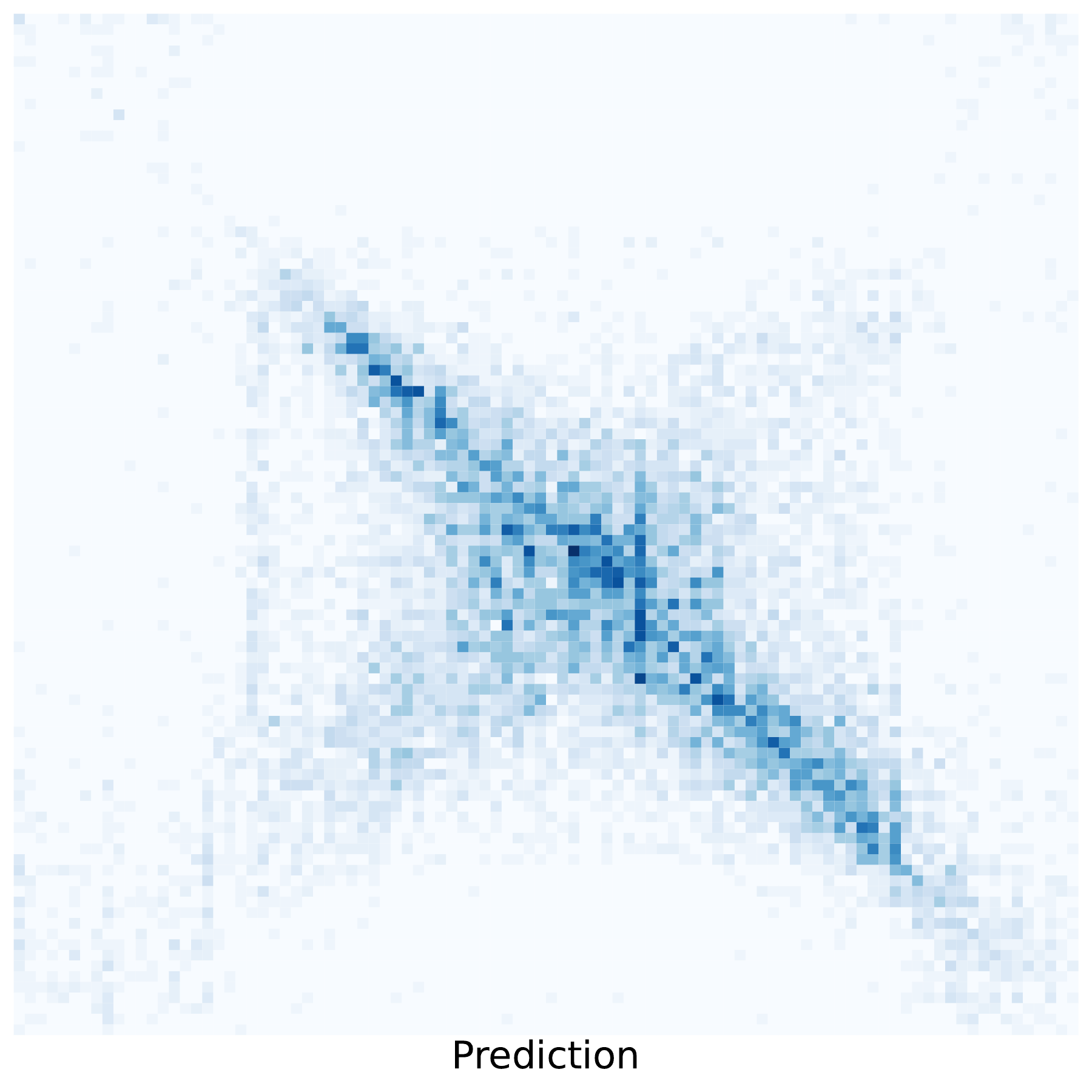}
        \caption{96 class partitioning}
        \label{fig:conf_k_5}
    \end{subfigure}
    \hfill
    \begin{subfigure}[b]{0.3\textwidth}
        \centering
        \includegraphics[width=\linewidth]{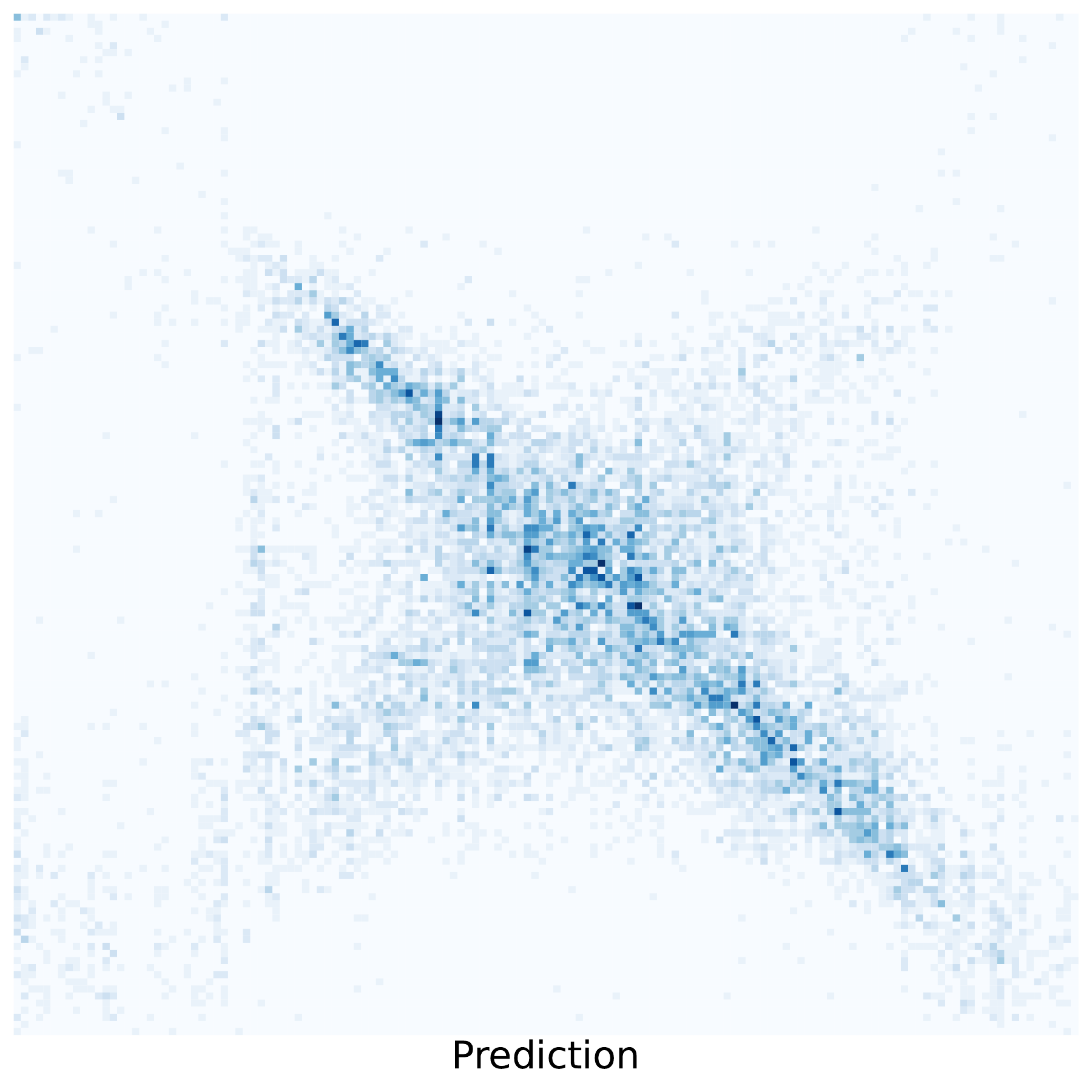}
        \caption{144 class partitioning}
        \label{fig:conf_k_6}
    \end{subfigure}
        \caption{\textbf{Confusion matrices under different number of classes} provide more in-depth comparison of clock timestamp estimation performance.}
    \label{fig:conf_k_ablation}
\end{figure*} 

\begin{figure*}[h!]
    \centering
    \includegraphics[width=0.7\linewidth]{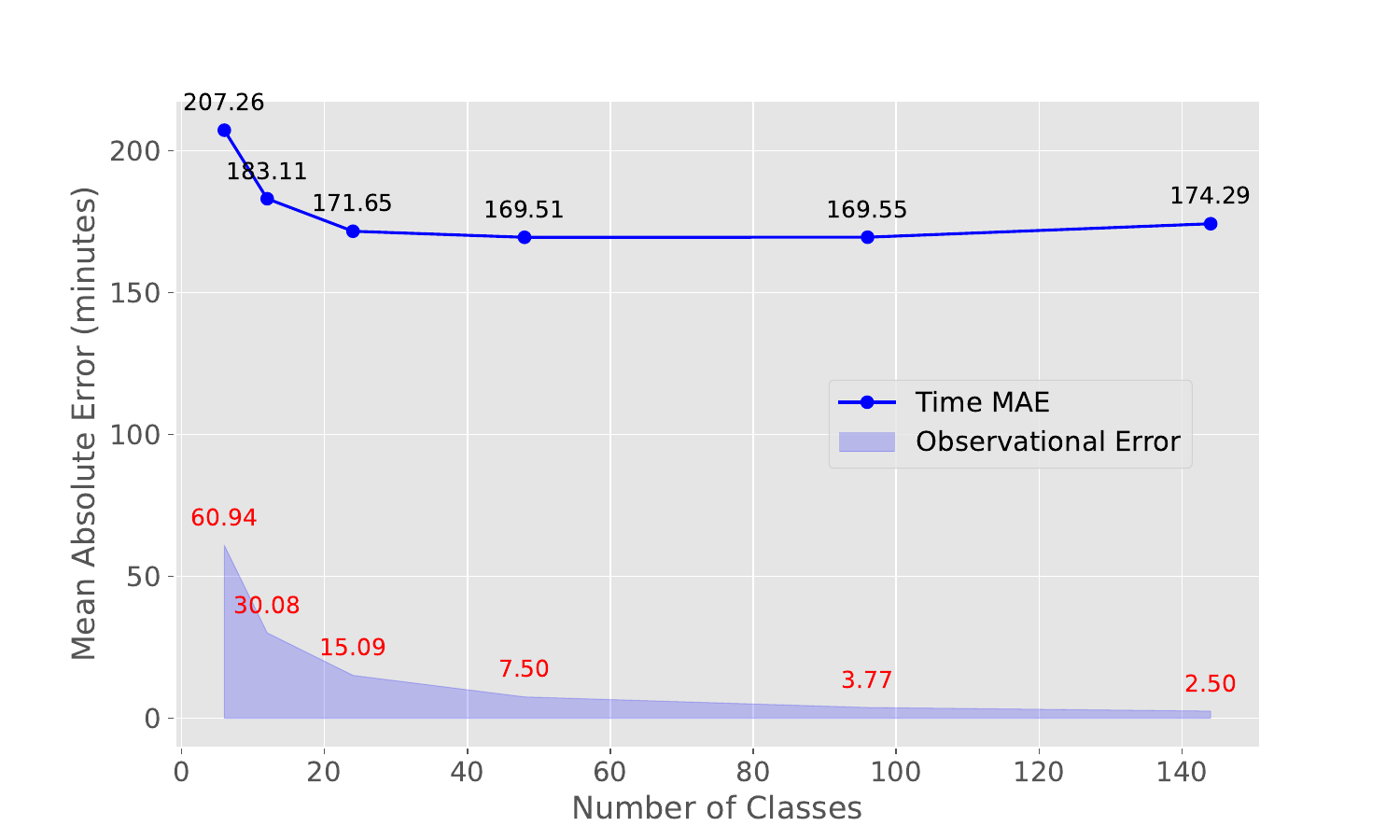}
    \caption{\textbf{Comparative error analysis of different class partitioning schemes,} it shows trends of mean absolute error (MAE) and observational error.}
    \label{fig:k_ablation}
\end{figure*}
\begin{figure*}[h!]
    \centering \includegraphics[width=0.7\linewidth]{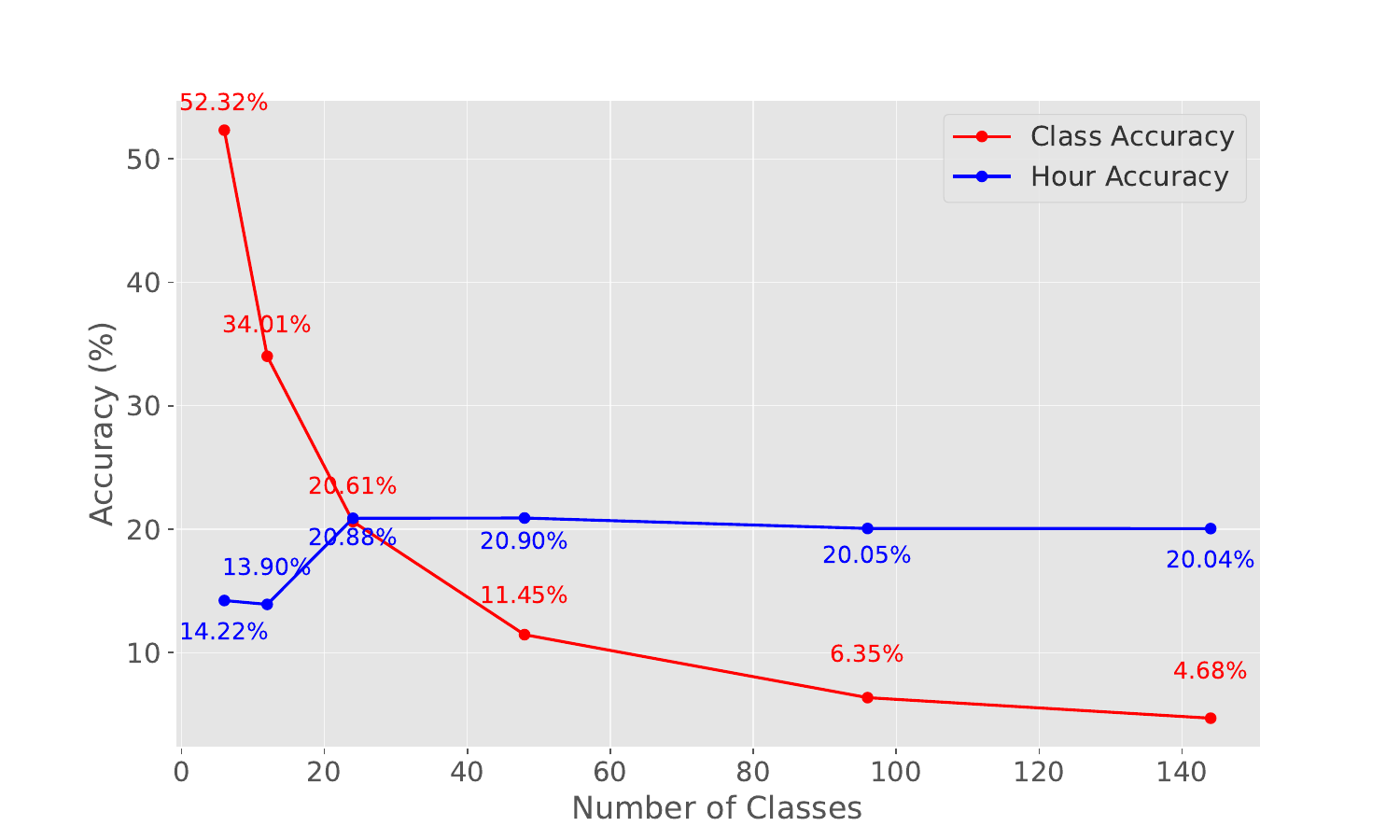}
    \caption{\textbf{Increasing number of classes does not further improve the time prediction accuracy.}, \textcolor{red}{Class Accuracy} represents the raw classification accuracy, and \textcolor{blue}{Hour Accuracy} is calculated by $\frac{1}{|D|} \sum_{i = 0}^{|D|-1} \mathbb{1}_{\left( \left\|\hat{Y}_{i} - Y_{i}\right\|_{1} \leq 30 \ \text{minutes} \right)}$, $\hat{Y}_{i}, Y_{i}$ are prediction and ground truth timestamps correspondingly.}
    \label{fig:k_accuracy}
\end{figure*}
\paragraph{Cyclic vector encoding: }As we identified in the main paper, the regression range for clock timestamp is a disconnected set. Here we present an attempt to solve the discontinuity of the clock timestamp scalar range: we adopted a previous method bridging the gap by trigonometric encoding and decoding to cyclic data \citep{vamplew}. Specifically, it encodes the scalar data $y$ into points on the unit circle $\left(\cos\left(y/y_{max}\right), \sin\left(y/y_{max}\right)\right)$, and decodes the model outputs by reversing this process. Such representation space is proved to be continuous \citep{Zhou_2019_CVPR}. It bridged the gap between the end and the start of the regression value range, which was supposed to be close. We tried this remedy and found that it slightly mitigates the issue of over-concentration on the average values, as shown in \cref{fig:regressor_prediction_cyclic}. 

However, although this modification managed to rescue part of night images that are wrongly predicted toward the mean value of the whole target value range, it still exhibits poor prediction fairness, with most of the predictions falling in certain short time spans. The possible cause for such phenomena could still be the local minima that persist in the MSE loss landscape due to the prevailing timestamp ambiguities we discussed. Another observation in \cref{fig:cyclic_encoding} is that there exists an obvious gap between the distribution of trigonometric encoding of ground truth timestamps and the predictions. This suggests that the cyclical correlation between visual appearances and clock time may not perfectly follow the simple unit-circle assumption in \citet{vamplew}. In contrast, our proposed learnable embeddings for the target clock time labels in TICL can capture more complex correlations between different periods of clock times and visual cues without imposing such assumptions.

Therefore, the regression approaches struggle to properly address the ambiguity between time and visual features. Regression-based solutions are thus not as favourable for the pretext task of image clock time estimation.

\subsection{Ablation study on class partitioning}
\label{sec:class_num}

In the main paper, we adhere to the 24-class classification scheme used in previous methods. As loss of precision may introduce observation errors, we explore the effects of different granularities of class partitioning on pretext tasks. 

To measure the precision loss, we compute observational errors, which are the average difference between actual timestamps and the converted class timestamps. \cref{fig:k_ablation} shows the mean absolute error (MAE) and the observational errors for different partitions of classes. As a part of MAE, observational errors are inherent such that they persist even with perfect class predictions \citep{conforti2020automatic}. Specifically, a small number of classes induces larger MAE, which is reasonable since converting actual timestamps to coarser time-span classes introduces larger additional observational errors.

However, this does not imply that extremely fine partitions should always be used to reduce observational error. We find that finer class partitioning, such as 144 classes, does not further improve the performance. In particular, \cref{fig:conf_k_ablation} presents the performance of the TICL model on the TOC test set under different class partitioning. The overall distribution of predictions exhibits similar patterns despite different granularities. \cref{fig:k_accuracy} highlights both class accuracy and hour accuracy for the model. The visualisation shows that while class accuracy drops significantly as the number of classes increases, the overall hour accuracy remains stable once the number of classes exceeds 24. This degradation in class accuracy with finer partitioning can be attributed to the smaller sample volumes within each class. The smaller the sample volume for each class, the more under-represented it tends to be \citep{sangalli2021constrained}. This suggests a potential drawback of finer class partitioning for downstream tasks involving time class embeddings.

\begin{figure*}[ht]
    \centering
    \begin{subfigure}[b]{0.43\linewidth}
        \centering
        \includegraphics[width=\linewidth]{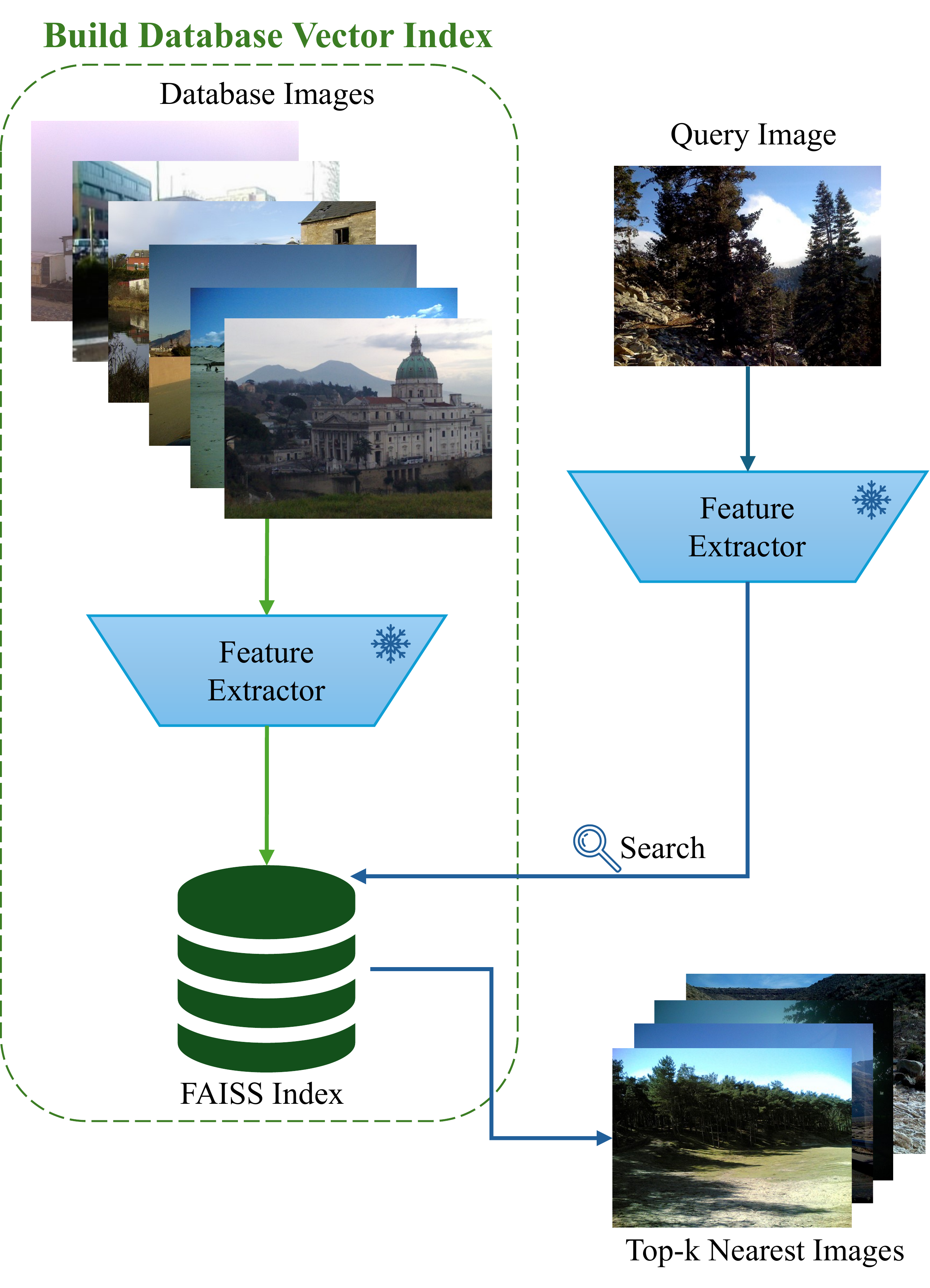}
        \caption{Time-based image retrieval}
        \label{fig:retrieva_modell}
    \end{subfigure}
    \hfill
    \begin{subfigure}[b]{0.38\linewidth}
        \centering
        \includegraphics[width=\linewidth]{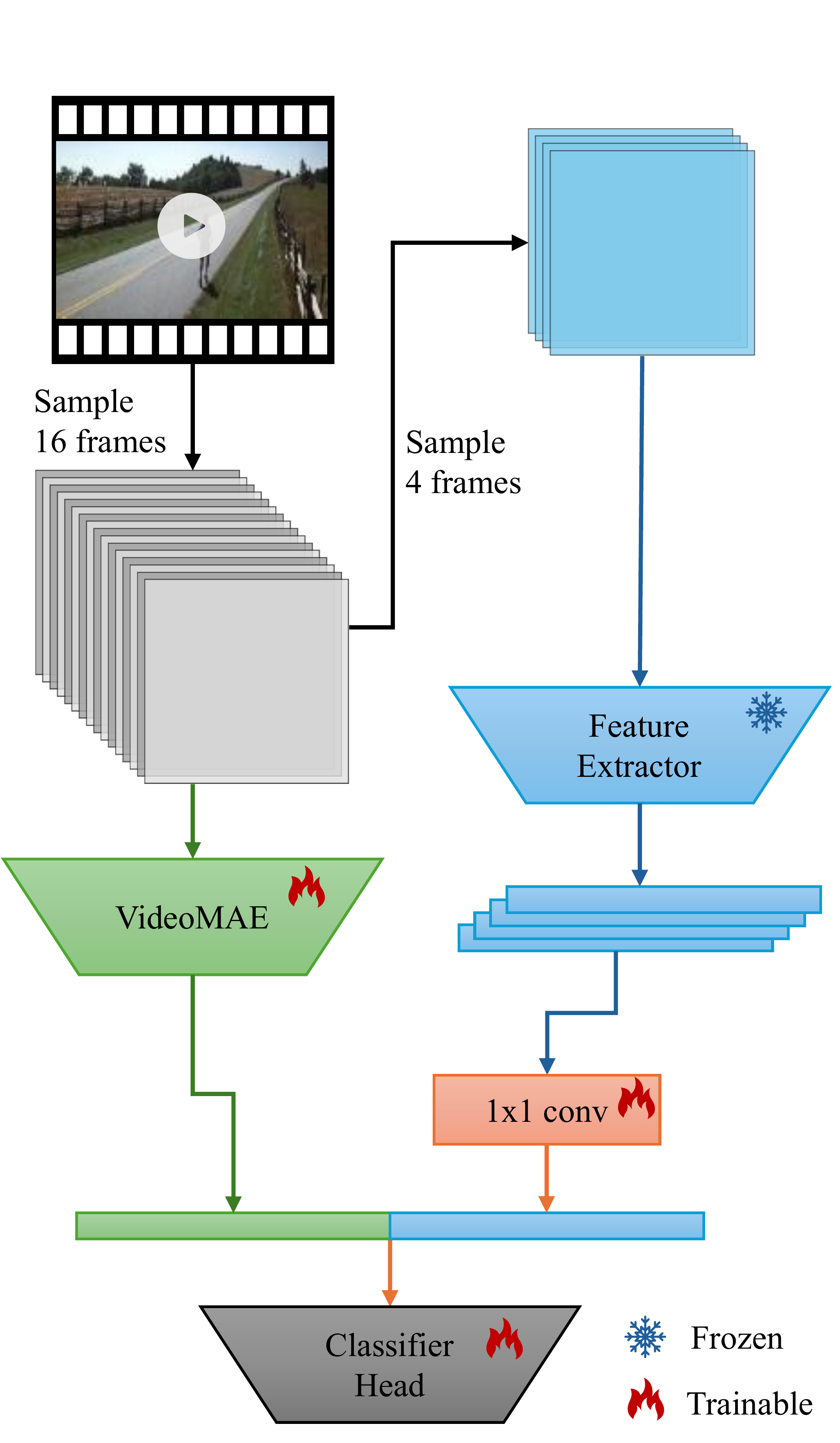}
        \caption{Video scene classification}
        \label{fig:videomae}
    \end{subfigure}
    
    \caption{\textbf{Zero-shot downstream pipelines.} (a) corresponds with experiment pipelines for retrieval in the main paper, which is a zero-shot vector search engine for same-hour images based on FAISS \citep{johnson2019billion}, and (b) shows one of the pipeline for video scene classification in addition to linear probing, in which we test the capabilities of TICL by plugging in the corresponding models to the feature extractor whose outputs are convoluted and concatenated to the backbone features \citep{tong2022videomae}.}
    \label{fig:downstream}
\end{figure*}

Since the difference between clock time estimation performance of the 24-class partition and the optimal result achieved with different class partitioning is within an acceptable range, we choose the 24-class partition as the default in our main work. This choice allows for a fair comparison against previous methods, to ensure that our improvements are due to the proposed techniques rather than variations in class partitioning. Additionally, the 24-class partitioning, which reached $\text{Class Accuracy} \approx \text{Hour Accuracy}$, also ensures that each class can be assigned enough samples so that a robust time class embedding could be learned.

To sum up, the ablation study on number of classes indicates that while the proposed TICL method can easily be extended to finer class partitioning schemes and maintains good hour accuracy and MAE, moderate granularity in class partitioning yields the best results for time estimation tasks. This supports our choice of a 24-class partitioning scheme for consistent benchmarking to previous baselines and verification of our conjecture on visual time awareness.

\CameraReady{
\subsection{Exploration on other contrastive learning methods}
Given the ambiguity of visual appearances with respect to time of day, in addition to the vanilla InfoNCE contrastive learning we applied, another relaxed supervised contrastive learning counterpart is also tested to learn the TICL model following Segsort~\citep{hwang2019segsort}. Concretely, we replace the InfoNCE objective with an hour-aware, cluster-based pulling/pushing scheme: for each hour, embeddings are repeatedly clustered with $k$-means; samples that fall into the \emph{same} hour-specific subcluster within a minibatch (memory bank) are pulled together, while others are pushed apart via a negative log-likelihood objective (as in~\cite{hwang2019segsort}). At inference, we embed a query image once and predict clock time via cosine $k$-nearest-neighbour over the frozen gallery of training embeddings. In the experiment below, We follow defaults in \cite{hwang2019segsort}: the hour-wise subcluster count is set to $n_{\text{sub}}=25$, the inner $k$-means refinement uses \texttt{max\_iter}$=10$ iterations per refresh, and the assignment sharpness is controlled by \texttt{concentration\_constant}$=10$. All remaining training and optimization settings are identical to those in TICL.

\begin{table}[h]
\centering
\caption{TOC and AMOS test results under different contrastive learning methods.}
\label{tab:segsort_both}
\resizebox{\linewidth}{!}{
\begin{tabular}{l|cccc|cccc}
\toprule
& \multicolumn{4}{c}{\textbf{TOC}} & \multicolumn{4}{c}{\textbf{AMOS}} \\
\cmidrule(lr){2-5}\cmidrule(lr){6-9}
\textbf{Contrastive Loss} & Top-1 acc $\uparrow$ & Top-3 acc $\uparrow$ & Top-5 acc $\uparrow$ & Time MAE (min.) $\downarrow$ & Top-1 acc & Top-3 acc & Top-5 acc & Time MAE (min.) \\
\midrule
Segsort-style \citep{hwang2019segsort}              & 18.97 & 39.32 & 52.27 & 171.55 & 12.29 & 28.37 & 39.57 & 209.00 \\
InfoNCE (Classification)   & 20.60 & 49.01 & \textbf{67.82} & 171.65 & \textbf{13.55} & \textbf{38.49} & \textbf{57.28} & \textbf{187.87} \\
InfoNCE ($k$NN)            & \textbf{25.67} & \textbf{49.32} & 66.74 & \textbf{156.24} & 11.14 & 31.01 & 48.84 & 220.94 \\
\bottomrule
\end{tabular}
}
\end{table}

As shown in \cref{tab:segsort_both}, across both datasets, the Segsort-style variant did not consistently improve over InfoNCE. On both TOC AMOS, InfoNCE methods achieves the best performance. We hypothesize two contributing factors: (i) prototype assignments in Segsort are well-suited to separable semantic classes, whereas time-of-day differences can be subtler than object-level semantics; (ii) hour-wise $k$-means over CLIP features tends to form subclusters by object identity early in training, which may misalign with temporal structure and provide weaker supervisory signals. Developing supervised contrastive objectives that more directly encode temporal neighbourhoods remains a promising direction; we therefore retain InfoNCE as the default in TICL.
}

\section{Qualitative Time-based Image Retrieval Results} 
\label{sec:retrieval_supp}
\begin{figure*}[t]
    \centering
    \includegraphics[width=0.9\linewidth]{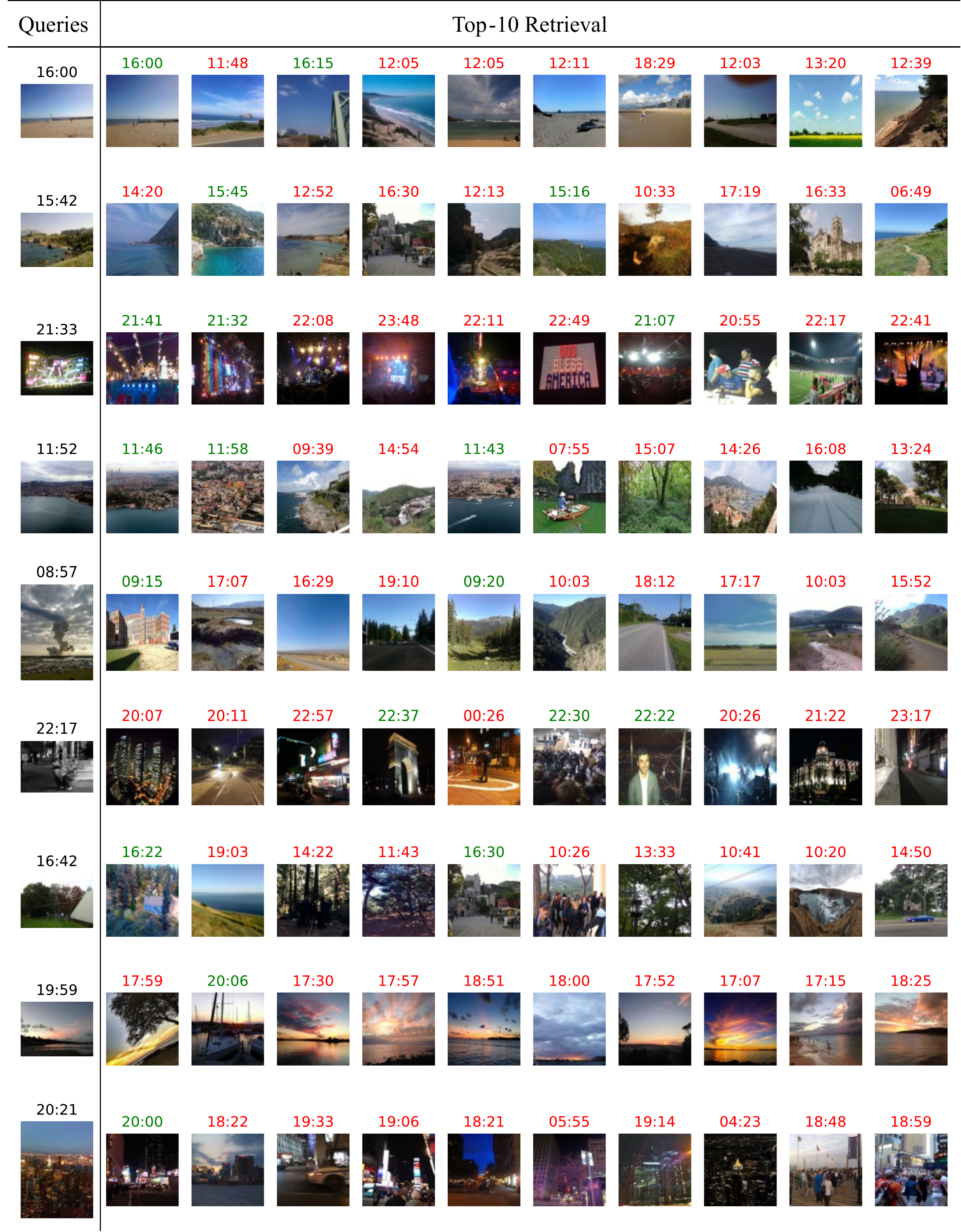}
    \caption{\textbf{Randomly sampled retrieval results.} Each image is annotated with its corresponding timestamp, \textcolor{ForestGreen}{green} captioned images are positive retrieval while \textcolor{red}{red} are negative predictions with $\text{Error} > \text{00:30}$, retrieved images closer to the left have larger similarity to the query images.}
    \label{fig:retrieval}
\end{figure*}

\cref{fig:retrieval} provides a closer look at the retrieved images using the pipeline in \cref{fig:retrieva_modell} as part of a more detailed qualitative evaluation of retrieval performance. Some of the retrieved images have totally different content from the query images, but share similar light conditions. This suggests that our model disentangles the time-awareness from rich semantics of CLIP representations, which have more semantic focus to the subjects. In addition, the negative predictions still share similar illumination to the query images, suggesting the essence and ambiguity of clock time to visual appearances.

\section{Additional Results on Video Scene Classification}
\label{sec:video_supp}

\subsection{Experiment setup} 
\label{sec:video_exp_setup_supp}
The performance of different models on the video scene classification task was evaluated across three datasets, each containing videos with distinct styles. Apart from simple linear probing, we also tested the model's performance fused with/against a baseline method VideoMAE \citep{tong2022videomae}. The detailed fusion architecture is visualized in \cref{fig:videomae}
\begin{itemize}
    \item \textbf{Hollywood2-Scene} \citep{marszalek09} is a movie clip-based dataset with 570 training videos and 582 test videos across 10 scene classes, totalling 20.1 hours. Each video represents a specific dramatic scene with multiple shots, meaning drastic view/subject changes within.
    \item \textbf{YUP++} \citep{6247815} comprises 1200 videos across 20 scenes captured by either stationary or moving cameras. Given the significant differences between the 20 scenes and the fact that the average clip duration is only 5 seconds, the classification task on it is considered less challenging \citep{wang2023flowdynamicscorrectionaction}.
    \item \textbf{360+x} dataset \citep{chen2024x360} is a more recent dataset introduced for holistic dynamic scene understanding with multiple views captured by stationary cameras. It consists of 15 indoor scenes and 13 outdoor scenes, with 1380 clips totalling 67.78 hours. Its multi-view and stationary camera traits enable us to evaluate how our learned time-awareness perform on different types of views individually.
\end{itemize}
\begin{figure*}[ht]
    \centering
    \begin{subfigure}[b]{0.9\linewidth}
        \centering
        \includegraphics[width=\linewidth]{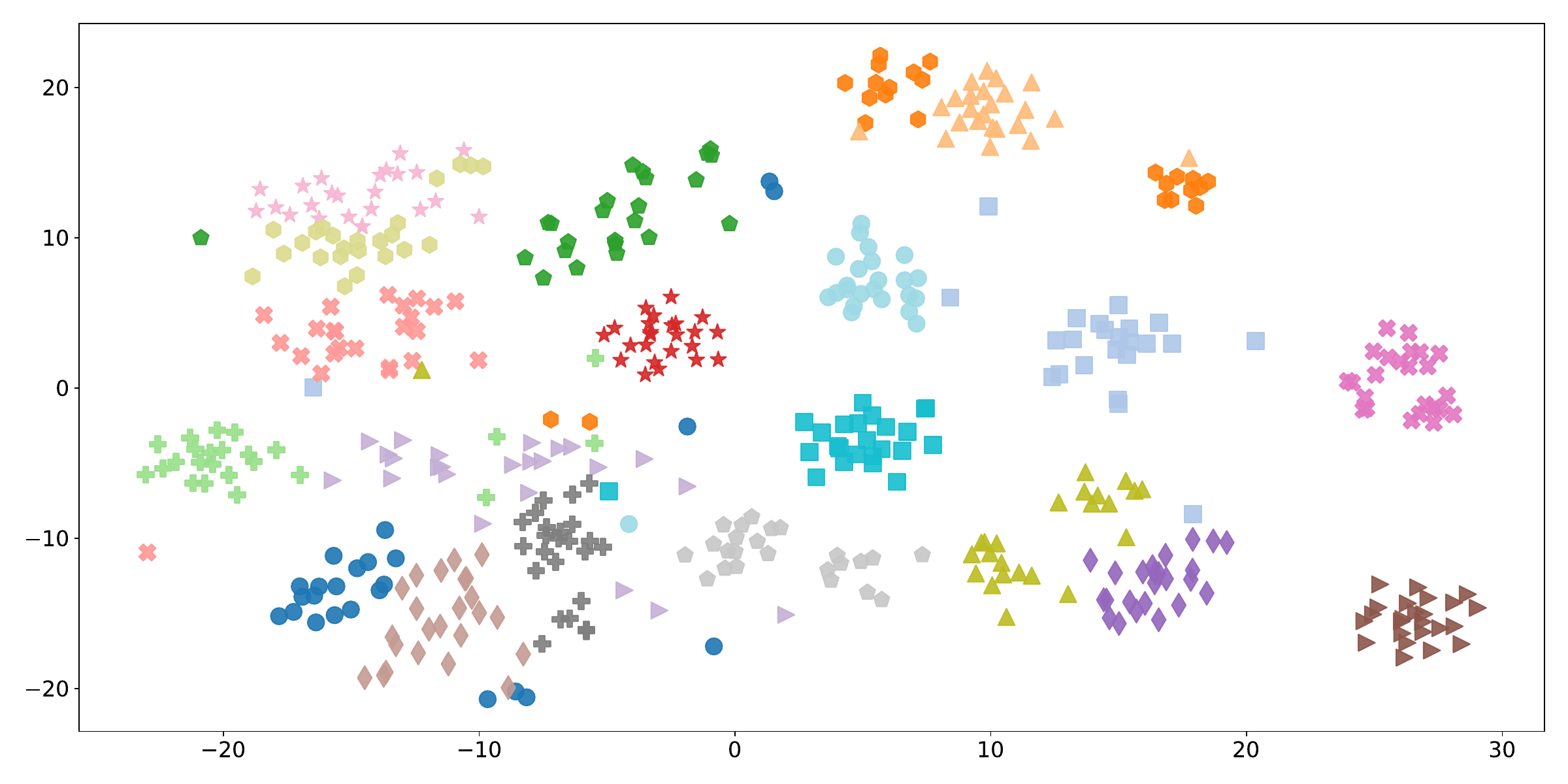}
        \caption{CLIP (ViT-L/14)}
        \label{fig:CLIP_YUP++}
    \end{subfigure}%
    \hfill
    \centering
    \begin{subfigure}[b]{0.9\linewidth}
        \centering
        \includegraphics[width=\linewidth]{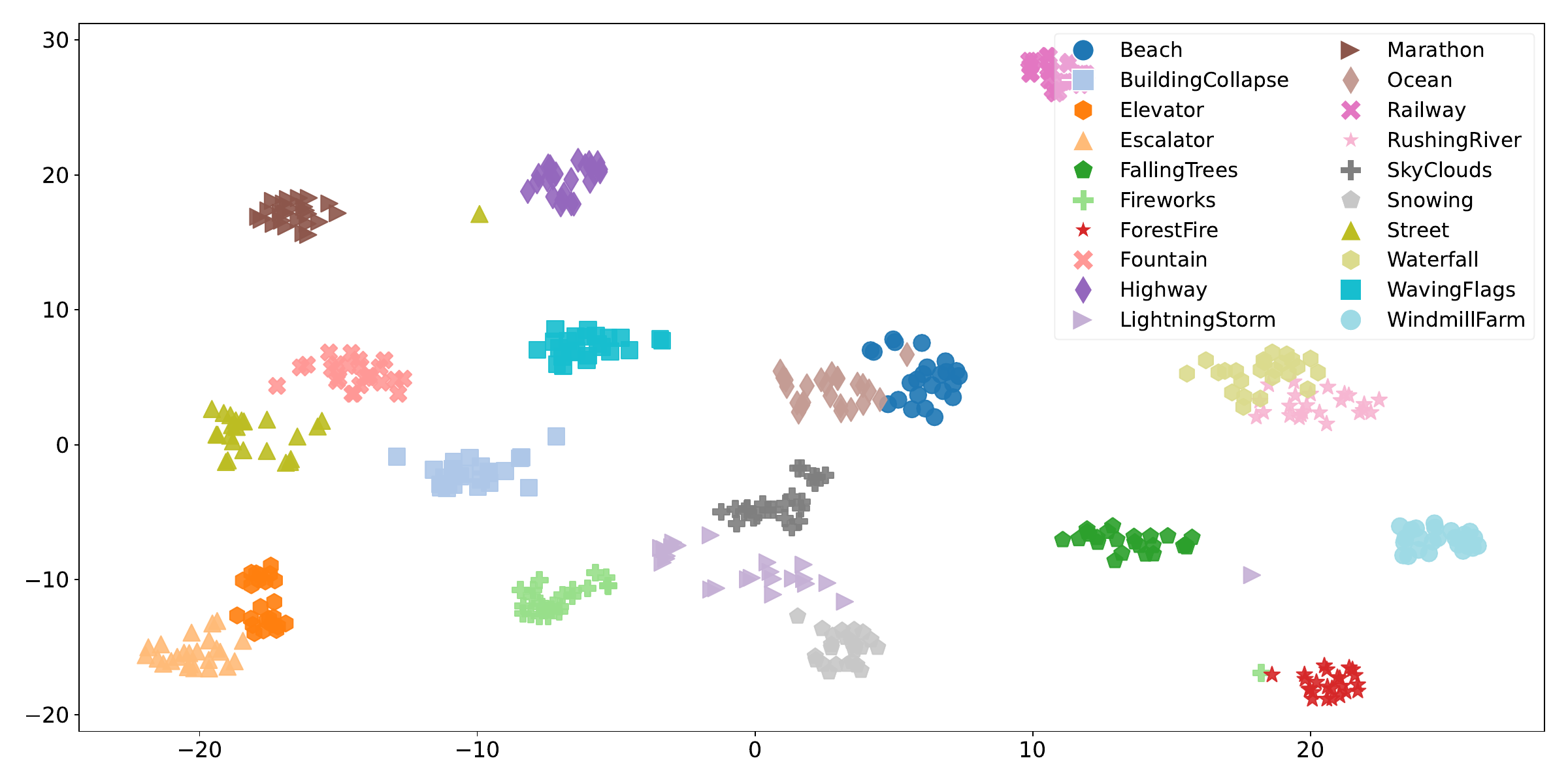}
        \caption{TICL}
        \label{fig:TICL_YUP++}
    \end{subfigure}%
    \hfill
    \caption{\textbf{t-SNE visualisation comparison.} It visualises time-aware video features in YUP++ dataset \citep{6247815}. Each embedding is annotated by their corresponding labels. It exhibits a similar trend to the t-SNE results in the main paper.}
    \label{fig:tsne_YUP++}
\end{figure*}
\paragraph{Hyper-parameters:} For fair comparison, a fixed set of hyper-parameters was used in different experiment trials. Apart from the number of epochs and the learning rate, we followed all the parameter settings in \citet{tong2022videomae}. And we only varied numbers epochs and learning rate for different datasets in. We report the best result achieved for each method tested. Specifically, a training/validation split of 5:1 was applied to each original training dataset to fairly select the best checkpoints for each method.

\begin{table*}[t]
\caption{Hyper-parameters used for video scene classification on different datasets.}
\label{tab:video_hyperparameters}
\centering
\resizebox{\linewidth}{!}{%
\begin{tabular}{l|cccc}
\toprule
\textbf{Hyper-parameter} & \textbf{Hollywood2-Scene} & \textbf{YUP++} &  \textbf{360x (Third-person)} & \textbf{360x (Panoramic)} \\
\midrule
Learning Rate   & $5\times 10^{-5}$ & $5\times 10^{-5}$ & $7\times 10^{-5}$ & $7\times 10^{-5}$ \\
\# Iterations       & 20      & 10      & 20      & 20      \\
\midrule
\multicolumn{5}{l}{\textbf{Default Settings from \citep{tong2022videomae, wolf2020huggingfacestransformersstateoftheartnatural} (Common Across All Datasets)}} \\\midrule
Optimizer Type  & \multicolumn{4}{c}{adamw\_torch\((\beta_1 = 0.9, \beta_2 = 0.999, \epsilon = 10^{-8})\)} \\
LR Scheduler    & \multicolumn{4}{c}{linear} \\
Batch Size      & \multicolumn{4}{c}{2} \\
\bottomrule
\end{tabular}
}
\end{table*}

\subsection{Time embedding coherence on video frames}

As discussed in the main paper, the observed improvements when integrating time-aware features with video classification backbone models could be attributed to the stronger intra-video consistency of these time-aware features.

To provide quantitative evidence of this consistency, we examine the characteristics of time-aware features across frames within each video. The backbone VideoMAE (ViT-B) model takes the input by sampling 16 frames evenly from each video. For the 16 input frames, we observed that the time-aware features of these 16 frames exhibit significantly smaller average variance compared to their CLIP features, as shown in \cref{tab:intra_video_variance}. 

This finding supports our intuition that a natural video that depicts a dynamic scene is typically captured over a short period of the day, leading to relatively small changes in the time-aware features of consecutive frames. In contrast, the CLIP features show more drastic changes between frames, making it harder to summarise consistent frame-wise features into coherent video-level features. The t-SNE visualisation comparisons to these features in main paper and \cref{fig:tsne_YUP++} provide additional results to prove that TICL video features are more separable than CLIP video features. 

Thus, time-aware feature extractors provide more consistency across different frames, making it easier to capture time-related visual priors in videos, which correlate with scene categories. These time-aware video priors eventually improved the video scene recognition performance, as illustrated in the main text.

\label{sec:frame_variance}
\begin{table*}[t]
\caption{\textbf{Mean intra-video feature variance.} It is computed by the mean feature variance of 16 input frames for each video using different models, showing a quantitative evidence of intra-video feature consistency of time-aware models.}
\label{tab:intra_video_variance}
\centering

\resizebox{\linewidth}{!}{%
\begin{tabular}{l|cccc}
\toprule
\textbf{Models} & \textbf{Hollywood2-Scene} & \textbf{YUP++} & \textbf{360+x (Third-person)} & \textbf{360+x (Panoramic)} \\
\midrule
CLIP (ViT-L/14)              & $7.49 \times 10^{-2}$ & $2.49 \times 10^{-2}$ & $3.31 \times 10^{-2}$ & $2.83 \times 10^{-2}$ \\
\citet{salem2022timestamp}   & $3.52 \times 10^{-6}$ & $1.23 \times 10^{-6}$ & $7.55 \times 10^{-7}$ & $7.86 \times 10^{-7}$ \\
\citet{zhai2019learning}     & $2.50 \times 10^{-4}$ & $1.00 \times 10^{-4}$ & $8.50 \times 10^{-5}$ & $7.59 \times 10^{-5}$ \\
TICL (Ours)                  & $3.33 \times 10^{-4}$ & $1.24 \times 10^{-4}$ & $1.44 \times 10^{-4}$ & $1.33 \times 10^{-4}$ \\
\bottomrule
\end{tabular}
}
\end{table*}

However, it is observed that the embeddings in \citet{salem2022timestamp} and \citet{zhai2019learning} have much smaller intra-video feature variances, but they perform worse than the TICL features we proposed. Given that the previous methods produce 128-dimensional time-aware embeddings, which dimensionality is much lower than TICL embeddings, it is expected that they have much smaller variances. Moreover, although previous methods perform moderately better than the baseline methods in the majority of test datasets, their performance degradation in panoramic video datasets suggests a limitation in terms of generalisation ability between different styles of videos, especially for those captured in rare camera views in the 360+x dataset \citep{chen2024x360}. In contrast, TICL utilising a strong foundation model generalised better across different kinds of videos.

In summary, time-aware embeddings could provide a more coherent representation among multiple sequential frames in a video, which are relatively invariant to sudden view/object changes altering the semantic meaning of the frame. Among the time-aware models, TICL gives more robust time-aware priors that generally bring more improvements than all the other time-aware models on different styles of video. 

\section{Additional Results on Time-aware Image Editing}
\label{sec:edit_appendix}
\subsection{Latent optimisation} 
\label{sec:lo_appendix}

\begin{figure*}[h!]
    \centering
    \includegraphics[width=\linewidth]{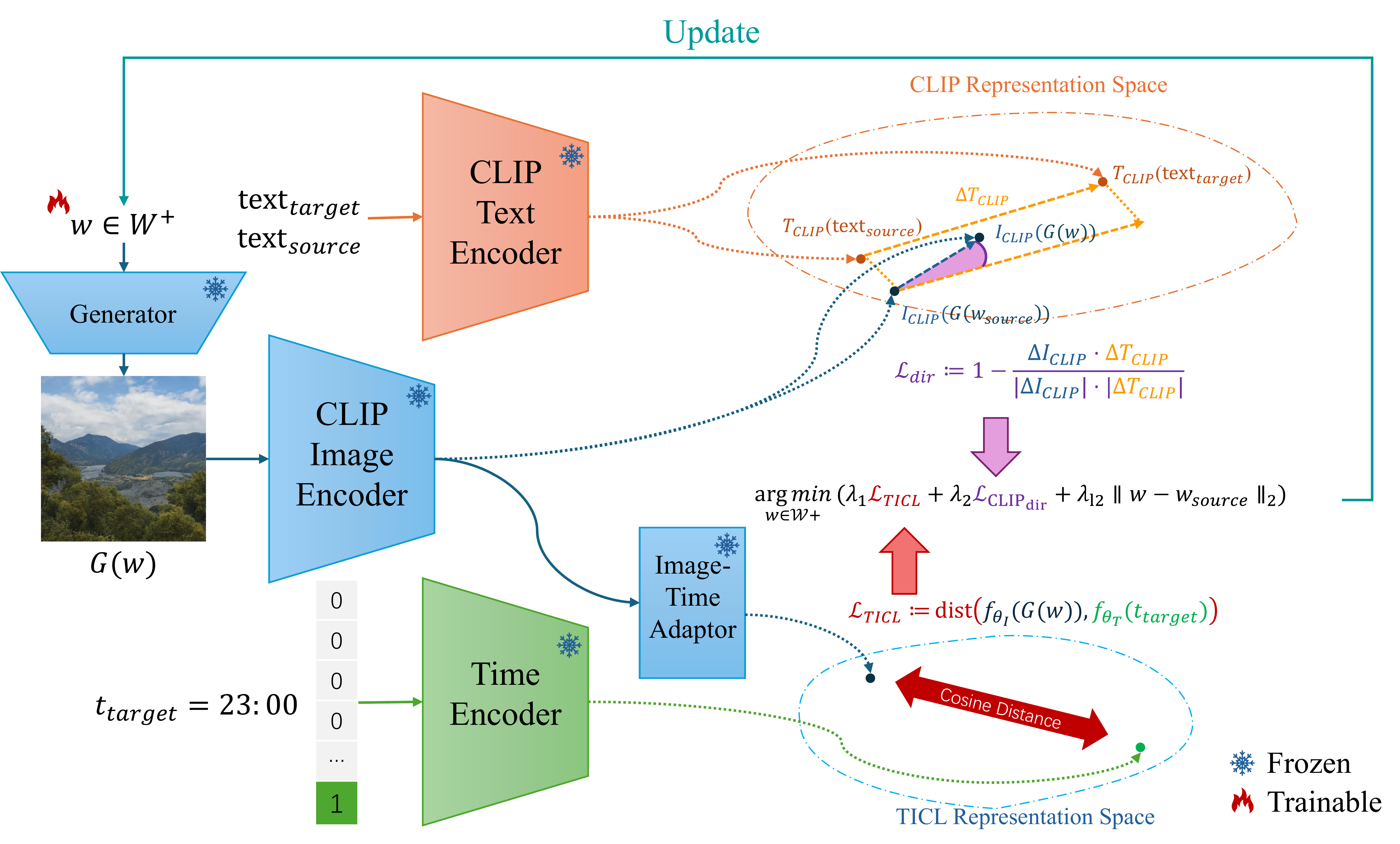}
    \caption{\textbf{Time-aware image editing pipeline.} This is the pipeline for latent optimisation for image editing, where \(w, w_{source}\) represents latent vectors for ongoing edit outcomes and original images, \(t_{\text{target}}\) is the one-hot encoding of the desired time of day for the output image, \(G(\cdot)\) is the generator, \(\mathrm{dist}(\cdot, \cdot)\) computes the cosine distance between two vectors, \(\Delta I_{\text{CLIP}}\) is the difference between CLIP embeddings of the original image, \(\Delta T_{\text{CLIP}}\) stands for the difference between the source and target caption embeddings. \(f_{\theta_I}(\cdot), f_{\theta_T}(\cdot)\) corresponds to components of the TICL model.}
    \label{fig:styleedit}
\end{figure*}

\begin{figure*}[h!] 
    \centering
    \includegraphics[width=0.6\linewidth]{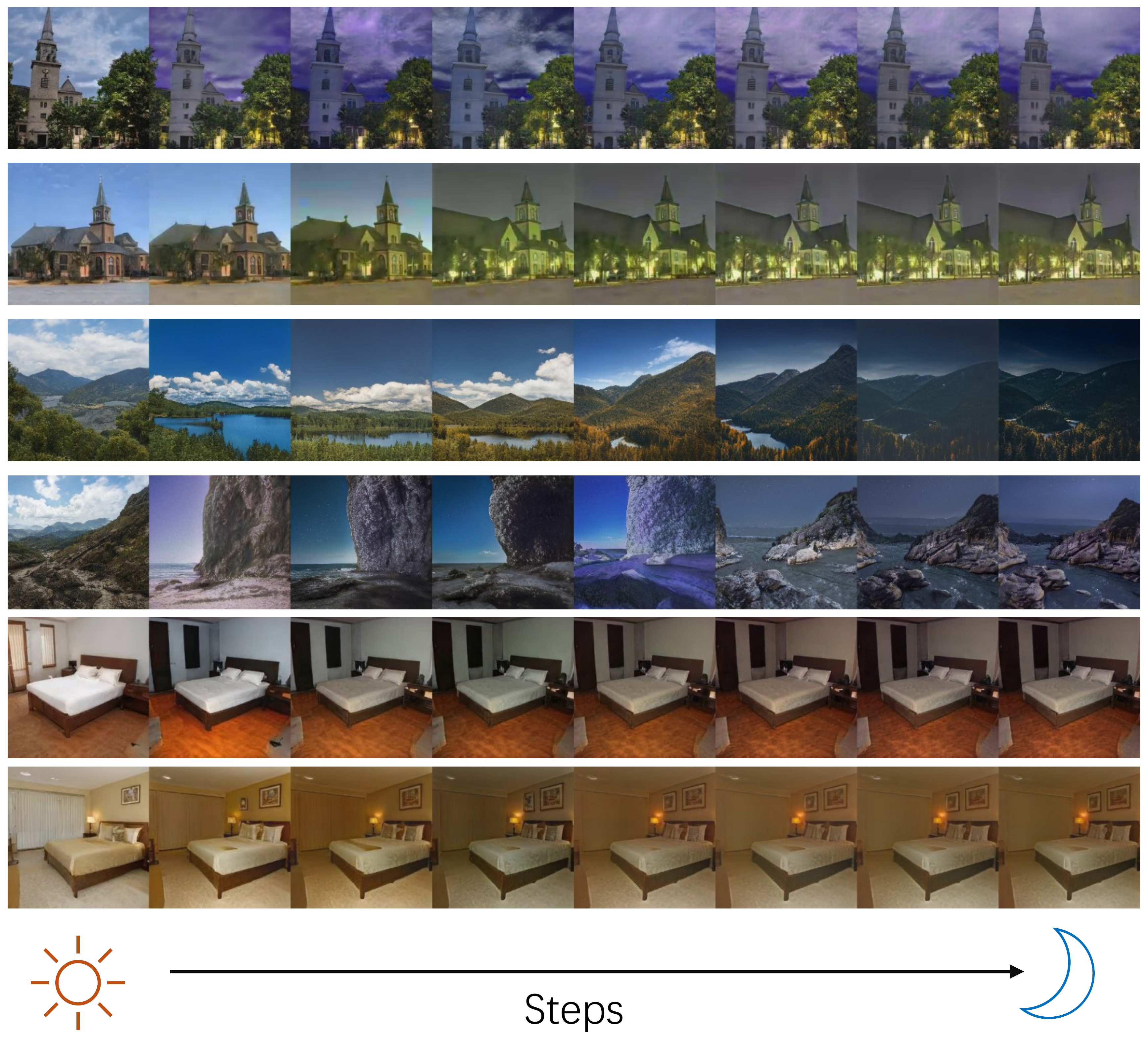}
    \caption{\textbf{Day-to-Night Edits.} An example of transitioning images from daytime to nighttime using latent optimization. This figure shows the progression of edits from various starting points to target times of day 22:00 (The rightmost figures are outputs for each edits.).}
    \label{fig:d2n_edit}
\end{figure*}
\begin{figure*}[h!] 
    \centering
    \includegraphics[width=0.6\linewidth]{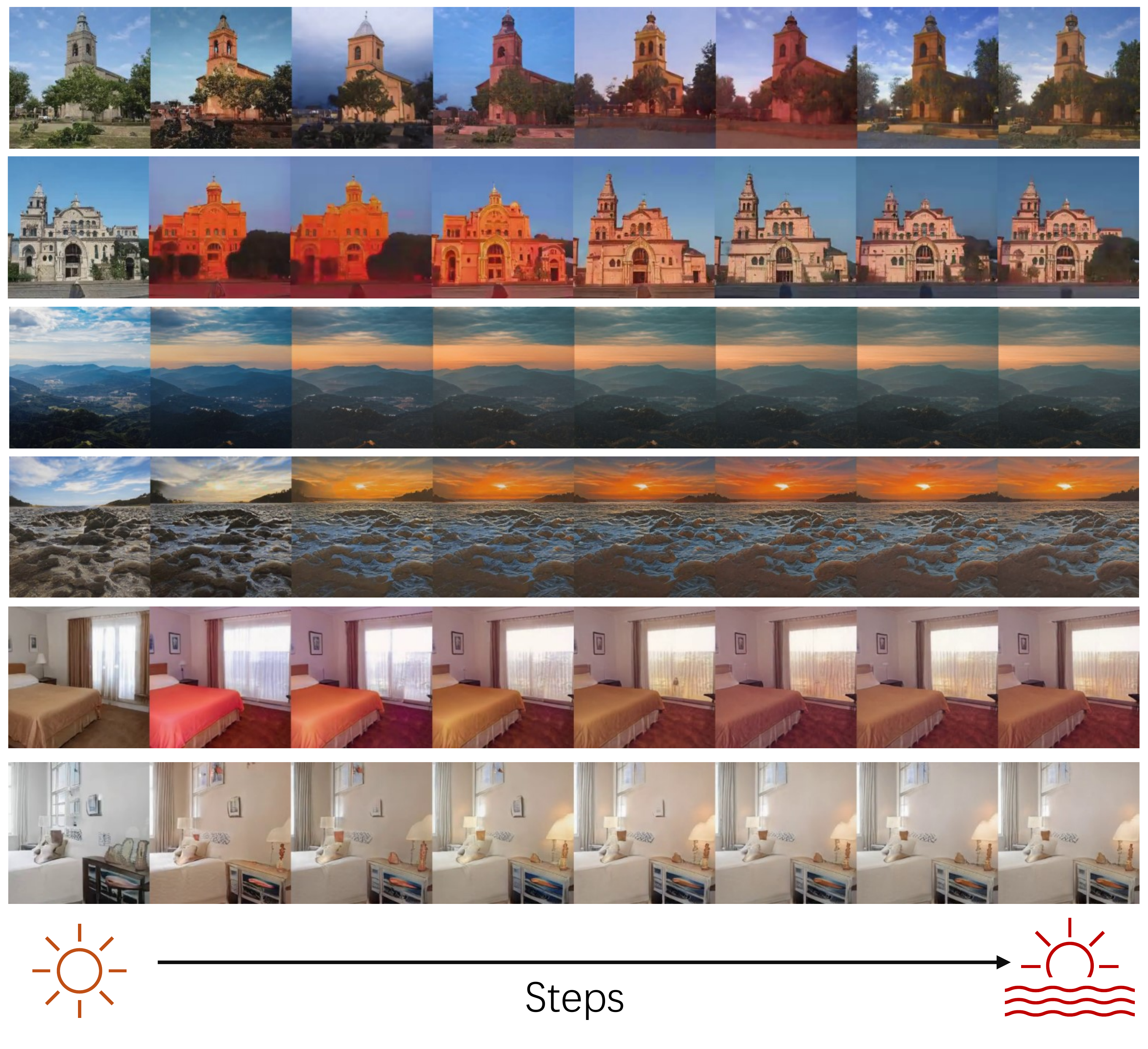}
    \caption{\textbf{Day-to-Evening Edits.} An example of transitioning images from daytime to evening using latent optimization. This figure shows the progression of edits from various starting points to target times of day 19:00 (The rightmost figures are outputs for each edits.).}
    \label{fig:d2e_edit}
\end{figure*}

\begin{figure*}[h!] 
    \centering
    \includegraphics[width=0.6\linewidth]{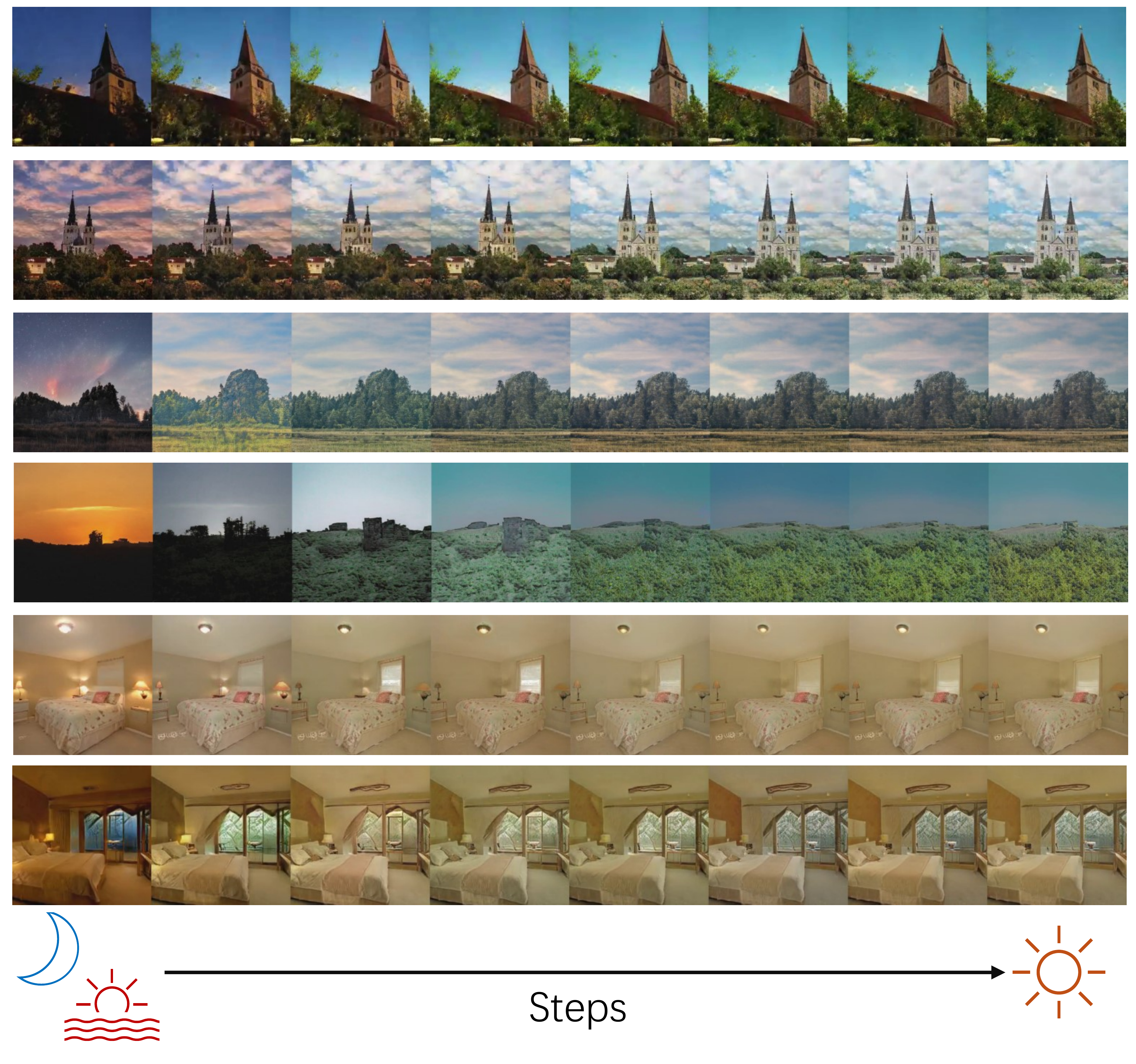}
    \caption{\textbf{Night/Sunset-to-Noon Edits.} An example of transitioning images from nighttime/sunset to noon using latent optimization. This figure shows the progression of edits from various starting points to target times of day 12:00 (The rightmost figures are outputs for each edits.).}
    \label{fig:ne2d_edit}
\end{figure*}

\paragraph{Experiment setup \& Hyper-parameters:} \cref{fig:styleedit} gives an overview of the experiment pipeline we used for the time-aware image editing task. For the main paper's results, each columns results were obtained via the same hyper-parameter setup. Specifically, we set the target timestamps $t_{target}$ as visualized in the figure and fixed the $\lambda_1 = \lambda_2 = 1$, using \texttt{Adam} optimiser with \texttt{lr\_rampup}$ = 0.05$ for all experiments; we varied other hyper-parameters as visualized in the following \cref{tab:combined_hyperparams} \wrt different target time periods of the day and the subject contents of images.

\begin{table}[h!]
\caption{Hyper-parameters for LHQ \citep{pinkney_lhq-sg2-1024}, LSUN-Church, and LSUN-Bedroom \citep{yu15lsun} editing processes.}
\label{tab:combined_hyperparams}
\centering
\resizebox{\linewidth}{!}{
\begin{tabular}{c|ccc|ccc|ccc}
\toprule
\multirow{2}{*}{\textbf{Hyper-parameter}} & \multicolumn{3}{c|}{\textbf{LHQ}} & \multicolumn{3}{c|}{\textbf{LSUN-Church}} & \multicolumn{3}{c}{\textbf{LSUN-Bedroom}} \\ 
\cmidrule(lr){2-4} \cmidrule(lr){5-7} \cmidrule(lr){8-10}
& \textbf{Noon} & \textbf{Evening} & \textbf{Night} & \textbf{Noon} & \textbf{Evening} & \textbf{Night} & \textbf{Noon} & \textbf{Evening} & \textbf{Night} \\ 
\midrule
$\lambda_{L2}$ & 0.001 & 0.001 & 0.0005 & 0.001 & 0.001 & 0.0005 & 0.001 & 0.001 & 0.0005 \\ 
\# iterations  & 50    & 50    & 100    & 50    & 50    & 100    & 50    & 50    & 100    \\ 
lr             & 0.07  & 0.07  & 0.1    & 0.5   & 0.5   & 0.5    & 0.05  & 0.05  & 0.05   \\ 
\bottomrule
\end{tabular}
}
\end{table}
\begin{table}[h!]
\caption{\textbf{FID Scores.} They quantitatively show how realistic the image editing results are for different methods on two image edit directions.}
\label{tab:fid_score}
\centering
\resizebox{0.85\columnwidth}{!}{%
\begin{tabular}{l|cc}
\toprule
                  Methods                        & Day-to-Night $\downarrow$ & Day-to-Sunset $\downarrow$ \\ \midrule
Latent optimisation ($\mathcal{L}_{CLIP}$) \citep{Patashnik_2021_ICCV}                                                                  & 53.55        &       50.60        \\
Latent optimisation ($\mathcal{L}_{\text{CLIP}_{dir}}$) & 50.07        &       50.59        \\
\textbf{Latent optimisation ($\mathcal{L}_{\text{CLIP}_{dir}} + \mathcal{L}_{TICL}$)} &\textbf{48.97}        &               \textbf{50.41} \\ \midrule
StyleGAN NADA \citep{gal2021stylegannada} & 78.80& 66.58 \\
CLIPStyler \citep{kwon2022clipstylerimagestyletransfer} & 71.12& 73.59 \\
\bottomrule
\end{tabular}
}

\end{table}

\paragraph{More qualitative results:} Additional results of latent optimisation based editing are presented. We varied the initial latent vectors and target hours to show the broader capabilities of our approach. \cref{fig:d2n_edit}, \cref{fig:d2e_edit} and \cref{fig:ne2d_edit} provide more examples of time-aware image editing with intermediate results during optimisation steps. The results suggest that our method could be applied to broad time-aware editing directions, which can start from images from various times of day.

\paragraph{Quantitative evaluations (User study): }In addition to the qualitative evaluation results, we also include quantitative metrics to evaluate the synthesis results. \Cref{tab:fid_score} gives FID scores \citep{NIPS2017_8a1d6947} to different edit directions calculated by the official PyTorch implementation of \citet{Seitzer2020FID} on 5000 samples for each methods. Our method outperforms existing methods with a smaller FID score suggesting more realism in the synthesised images. Additionally, we conducted a user study (by using the mean-opinion-score scheme) on the output images. The preference scores for each method are reported in \cref{tab:human_evaluation}, further demonstrating the advantages brought by incorporating time-aware embeddings.

\begin{table}[t]
\caption{\textbf{User study evaluating image editing qualities,} in which we report preference scores and their standard deviation (in brackets). Preference scores range from 1-5, and higher scores mean better preferences.}
\label{tab:human_evaluation}
\centering
\resizebox{0.85\columnwidth}{!}{%
\begin{tabular}{l|cc}
\toprule
                  Methods                        & Day-to-Night $\uparrow$  & Day-to-Sunset $\uparrow$ \\ \midrule
Latent optimisation ($\mathcal{L}_{CLIP}$) \citep{Patashnik_2021_ICCV}                                                                   &   2.80 (0.60)     &       2.84 (0.53)        \\
Latent optimisation ($\mathcal{L}_{\text{CLIP}_{dir}}$) & 2.63 (0.85)       &       3.28 (0.67)        \\
\textbf{Latent optimisation ($\mathcal{L}_{\text{CLIP}_{dir}} + \mathcal{L}_{TICL}$) (Ours)} &\textbf{3.34 (0.64)}        &               \textbf{4.01 (0.58)} \\ \midrule
StyleGAN NADA \citep{gal2021stylegannada} & 2.41 (0.89)&  2.36 (1.17) \\
CLIPStyler \citep{kwon2022clipstylerimagestyletransfer} & 2.08 (0.62)&  1.81 (0.93) \\
\bottomrule
\end{tabular}
}

\end{table}

\subsection{Editing with diffusion models} 
\label{sec:diffusion}

\begin{figure*}[h!]
    \centering
    \includegraphics[width=0.65\linewidth]{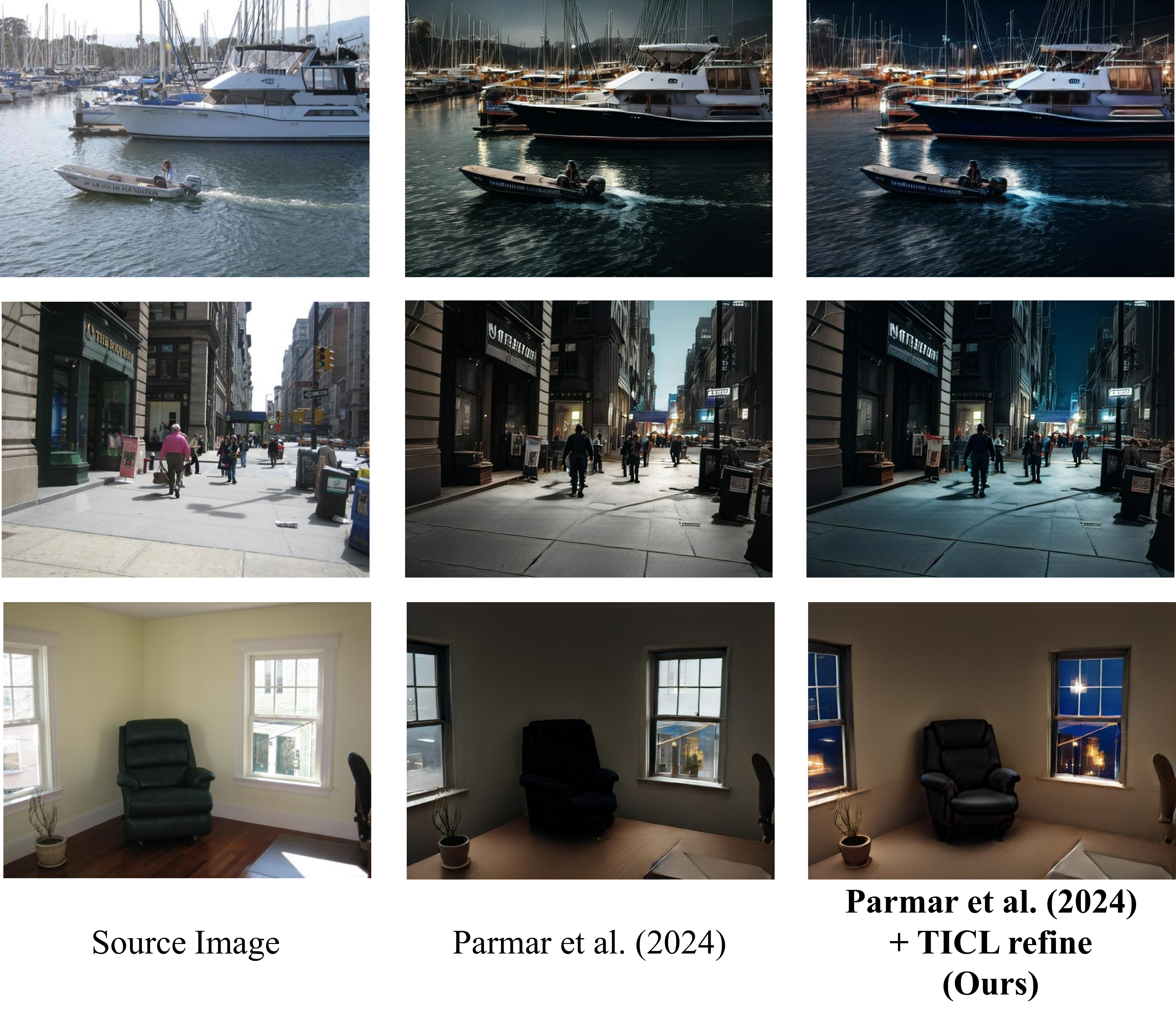}
    \caption{\textbf{Visualisation of Day-to-night edits (Part 1).} Transitioning images from day to night using the diffusion model.}
    \label{fig:diffusion_1}
\end{figure*}

\begin{figure*}[h!]
    \centering
    \includegraphics[width=0.65\linewidth]{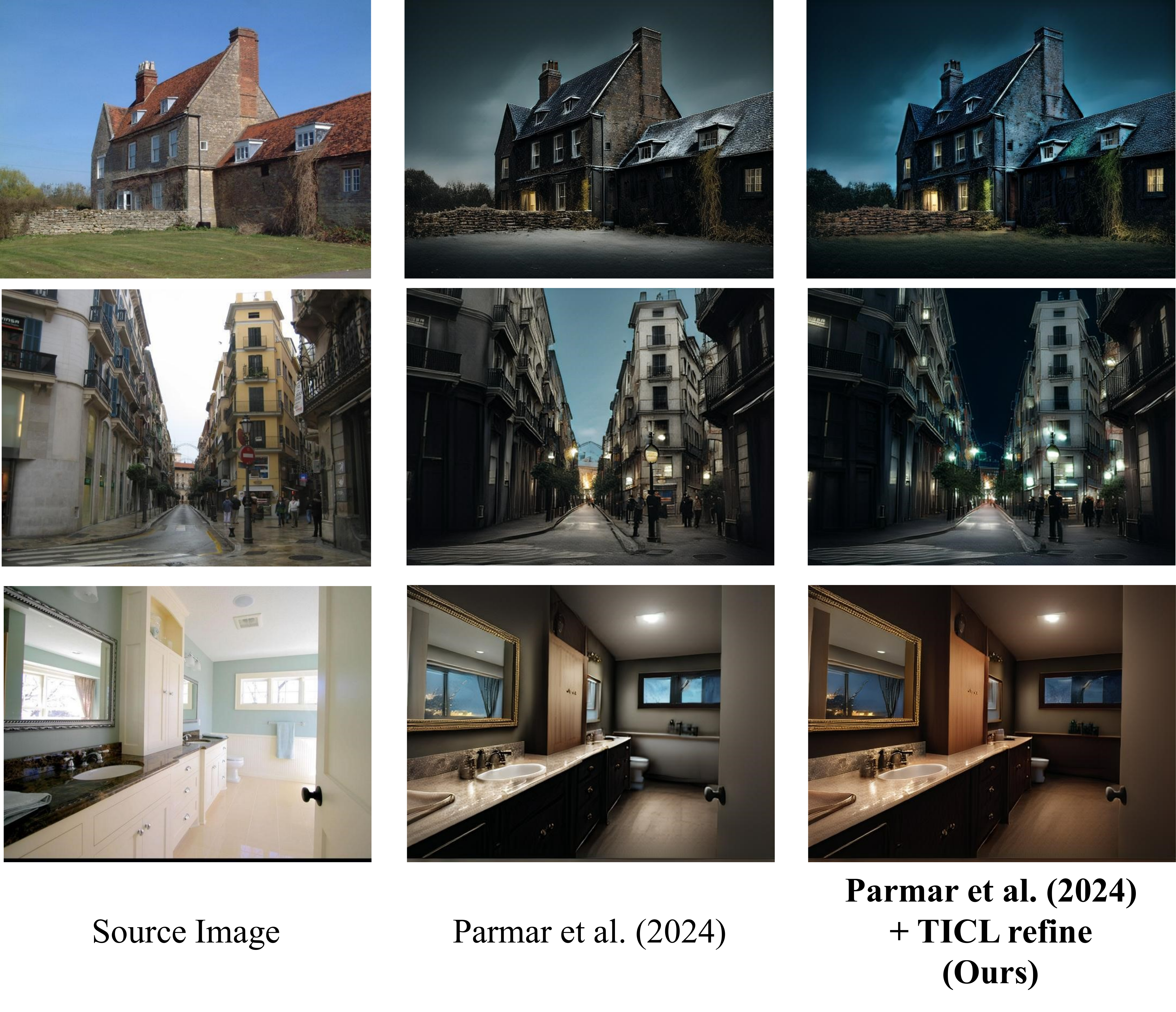}
    \caption{\textbf{Visualisation of Day-to-night edits (Part 2).} Continuing results of time-aware editing using the diffusion model.}
    \label{fig:diffusion_2}
\end{figure*}

\begin{figure*}[h!]
    \centering
    \includegraphics[width=0.75\linewidth]{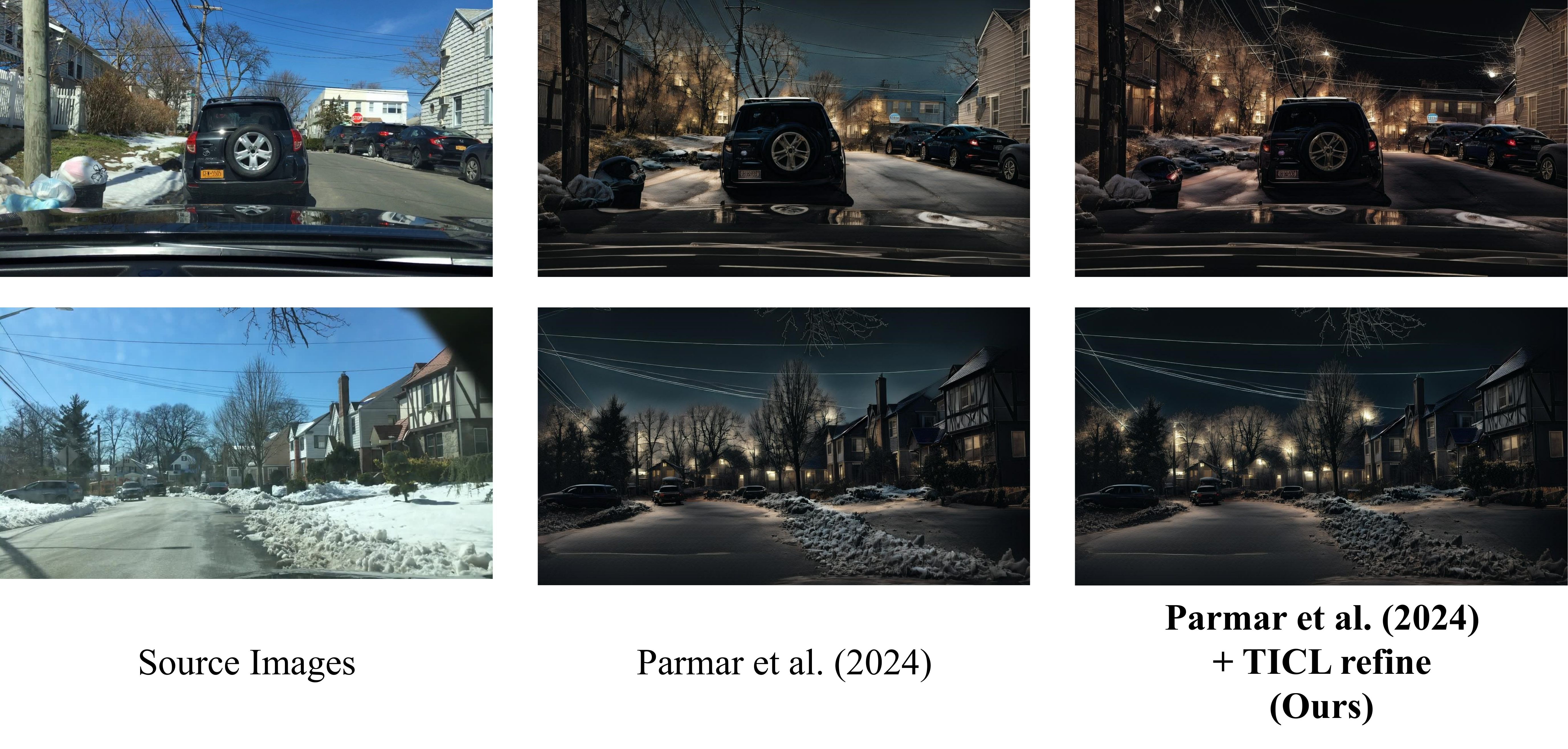}
    \vspace{-3mm}
    \caption{\textbf{Visualisation of Day-to-night edits (Part 3).} Further results demonstrating day-to-night transitions using the diffusion model.}
    \label{fig:diffusion_3}
\end{figure*}

Given that the previous baseline latent optimisation image editing method has limited capabilities, we extend our experiment to a more recent editing method \citet{parmar2024onestepimagetranslationtexttoimage} using diffusion models \citep{ho2020denoisingdiffusionprobabilisticmodels, rombach2021highresolution}. 

\paragraph{Experiment setup \& Hyper-parameters:} Specifically, we optimise the edit target text embedding $E^*_{text}$ to minimise the cosine distance between the time-aware embeddings of the output images and the target clock timestamp embeddings, which is written as:
\[
E^*_{text} = \arg \min _{E_{text}} \text{dist}\left(f_{\theta_I}\left(G\left (x, E_{text}\right )\right), f_{\theta_T}\left(t_{target}\right)\right)
\]
where $E^*_{text}$ is the target text embeddings for the text-based image editing model $G(\cdot, \cdot)$ takes input image $x$ and guidance text embedding $E_{text}$. $f_{\theta_I}, f_{\theta_T}$ corresponds to TICL model components. $dist(\cdot, \cdot)$ measures the cosine distance of two embeddings. It essentially optimises the guidance text embeddings $E_{text}$ to achieve better editing results that visually align with the target time. 

As for hyper-parameters, we applied default experiment settings for the baseline editing process as provided in \citet{parmar2024onestepimagetranslationtexttoimage} with text guidance set to ``\texttt{a photo of} $\{$\texttt{target time period} $\}$''. The subsequent optimisation process to $E_{text}$ uses \texttt{Adam} optimiser with learning rate $= 0.02$ and 10 iterations without any further configuration.

\paragraph{Qualitative results:} As shown in \cref{fig:diffusion_1}, \cref{fig:diffusion_2} and \cref{fig:diffusion_3}, although additional optimisation steps for each edit are required, it refines the existing method with more reasonable synthesis results compared with using purely text editing guidance, further proving the general applicability of the TICL embeddings to the whole image-editing subfield.

\section{Additional Text Queries on Clock Time Class Embeddings}

\begin{figure*}[h!]
    \centering
    \begin{subfigure}[b]{0.4\linewidth}
        \centering
        \includegraphics[width=\linewidth]{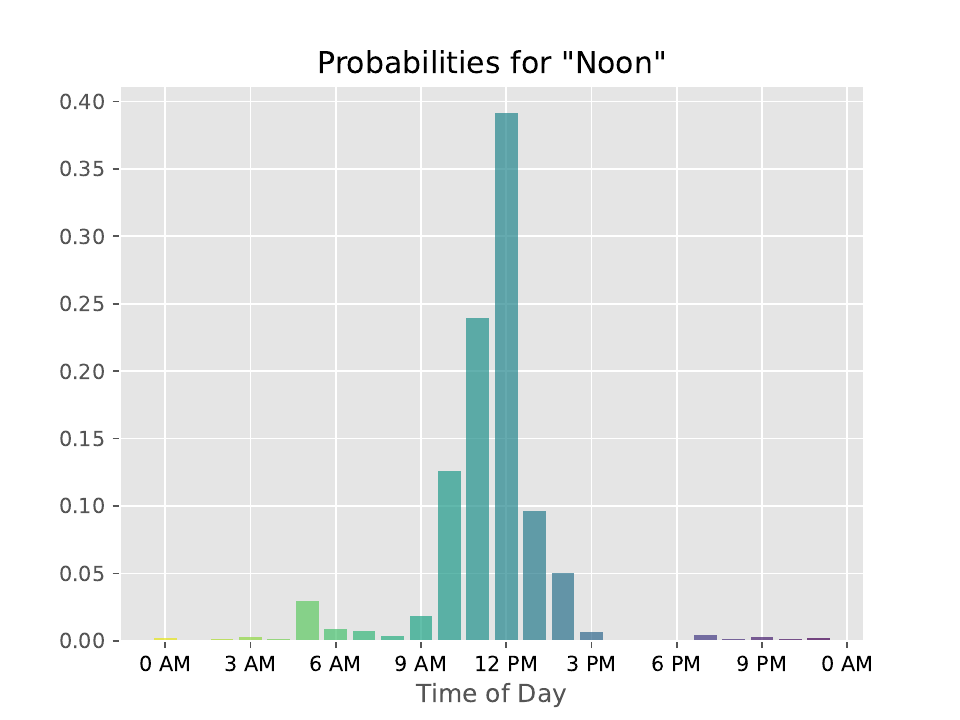}
        \label{fig:noon}
    \end{subfigure}
    \begin{subfigure}[b]{0.4\linewidth}
        \centering
        \includegraphics[width=\linewidth]{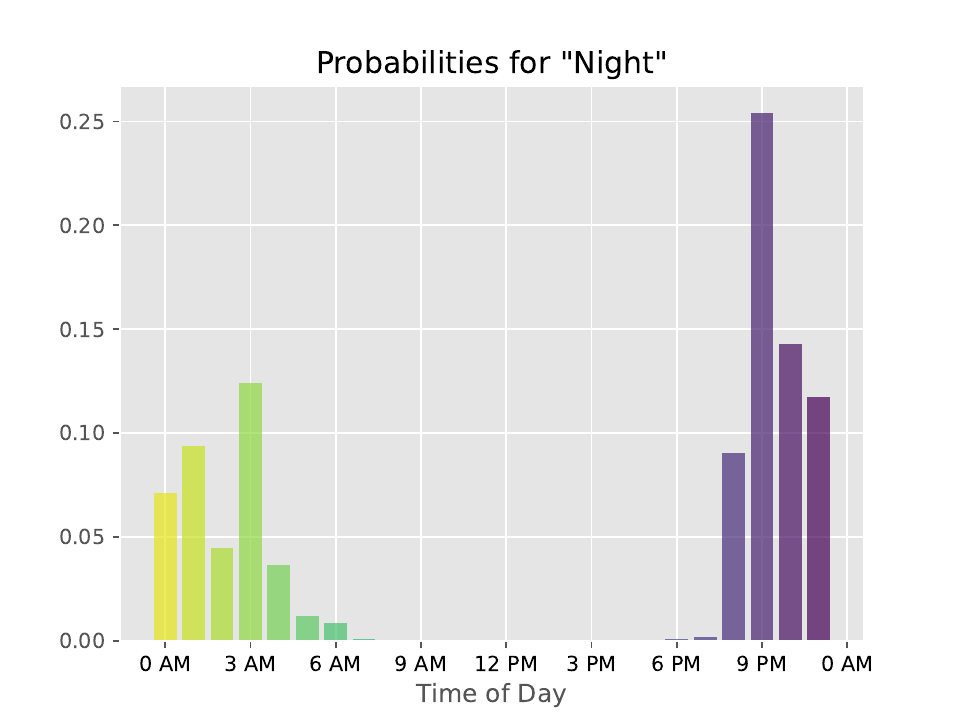}
        \label{fig:night}
    \end{subfigure}
    
    \vspace{-3mm}
    \begin{subfigure}[b]{0.4\linewidth}
        \centering
        \includegraphics[width=\linewidth]{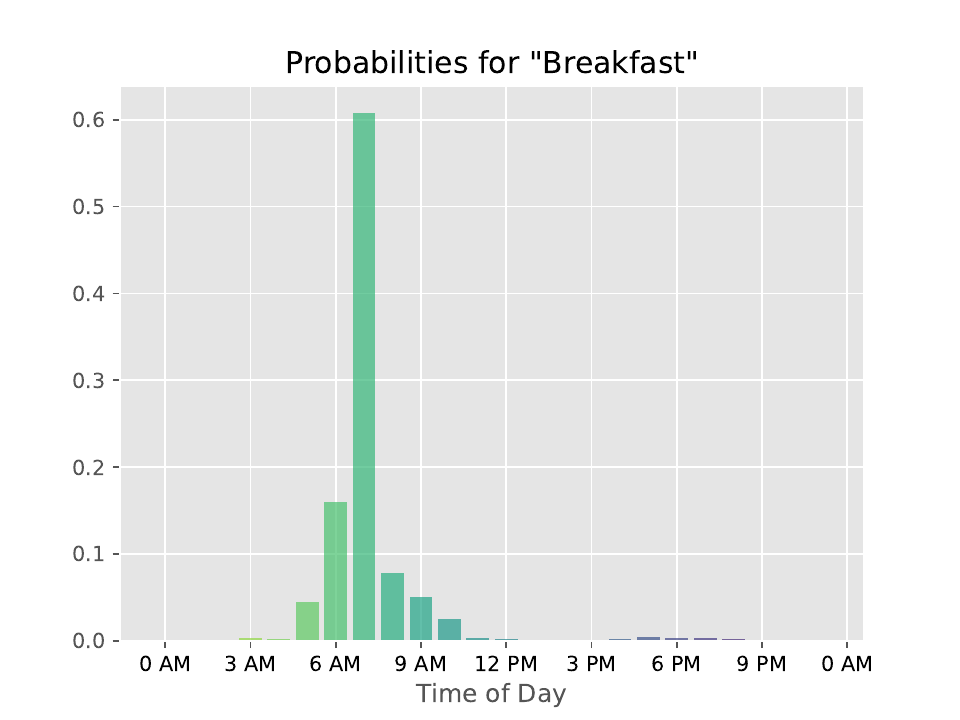}
        \label{fig:breakfast}
    \end{subfigure}
    \begin{subfigure}[b]{0.4\linewidth}
        \centering
        \includegraphics[width=\linewidth]{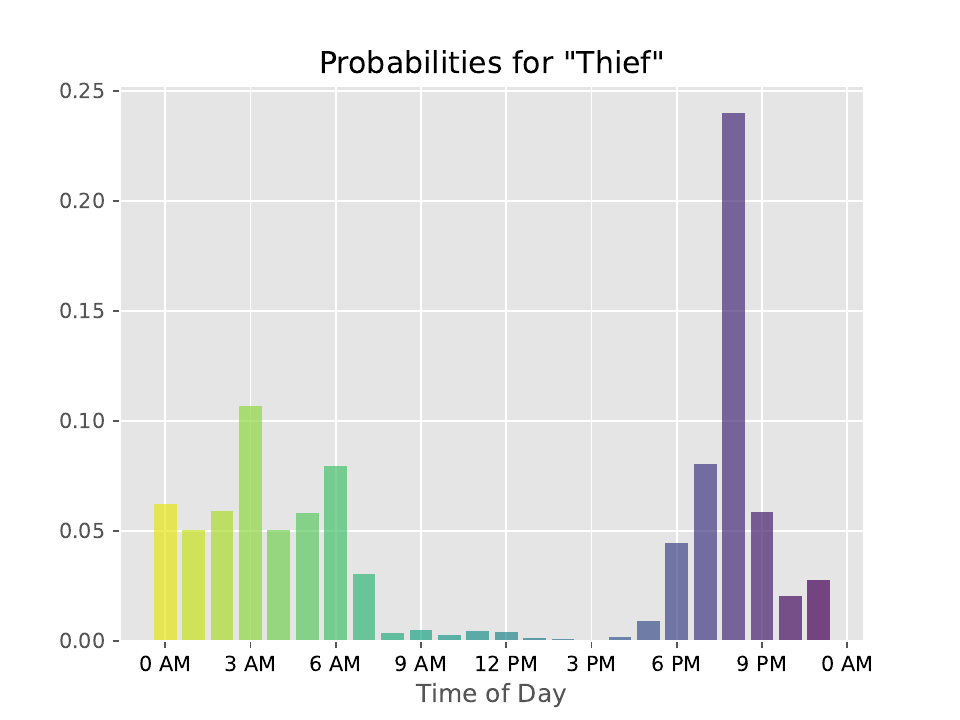}
        \label{fig:thief}
    \end{subfigure}

    \vspace{-3mm}
    \begin{subfigure}[b]{0.4\linewidth}
        \centering
        \includegraphics[width=\linewidth]{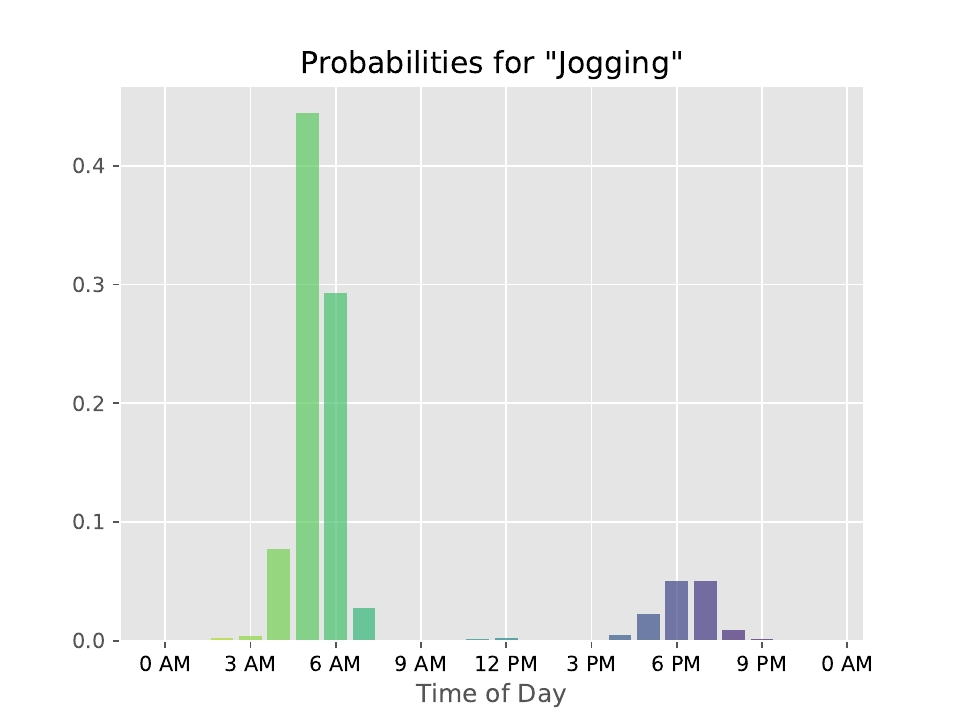}
        \label{fig:jogging}
    \end{subfigure}
    \begin{subfigure}[b]{0.4\linewidth}
        \centering
        \includegraphics[width=\linewidth]{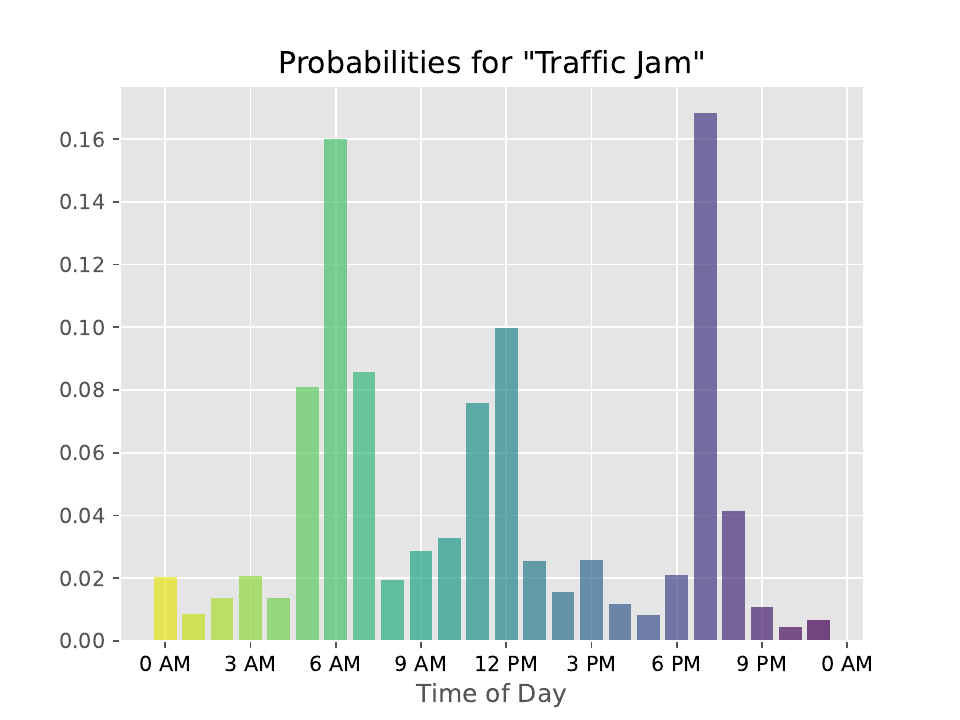}
        \label{fig:traffic_jam}
    \end{subfigure}
    
    \vspace{-3mm}
    \begin{subfigure}[b]{0.4\linewidth}
        \centering
        \includegraphics[width=\linewidth]{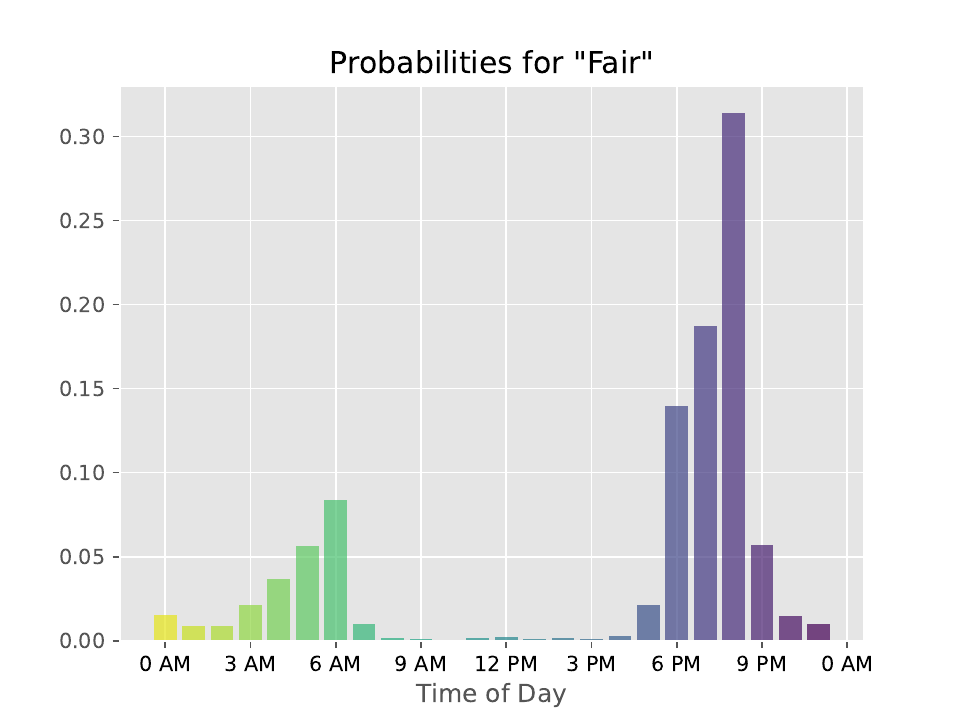}
        \label{fig:bar}
    \end{subfigure}
    \begin{subfigure}[b]{0.4\linewidth}
        \centering
        \includegraphics[width=\linewidth]{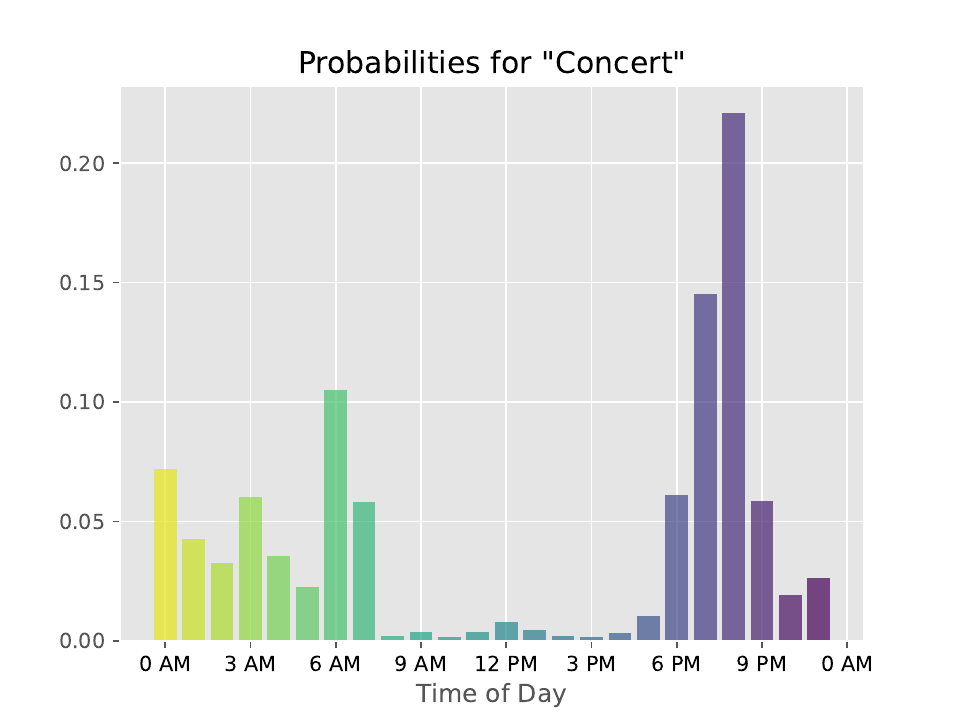}
        \label{fig:concert}
    \end{subfigure}
    \vspace{-3mm}
    \caption{\textbf{Probability measure of the similarity between time classes and text queries.} The x-axis is hour classes and y-axis is probabilities calculated by $\mathbf{Softmax}$.}
    \label{fig:text_queries}
\end{figure*}

In video scene classification tasks, we explored the semantic correlations between clock timestamps and scenes, and here we provide several examples to illustrate these connections. The Time Encoder and Image-Time Adaptor modules are designed to align visual CLIP representations and clock time embeddings. As a result, the learned time embeddings naturally align with CLIP text embeddings. This alignment allows us to factorise text concepts using TICL time embeddings and vice versa. Specifically, for each input text embedding, \(T_{CLIP}\), we compute their similarity with time-class embeddings, \(T_i\), using the Softmax function: 
\[
\mathbf{Softmax} = \frac{\exp \left(T_{CLIP} \cdot T_{i}\right)}{\sum_{j=0}^{|C|-1} \exp \left(T_{CLIP} \cdot T_{j} \right)}
\]
where \(T_i, T_j\) are the TICL class embeddings. This formulation offers a probabilistic measure of the similarity between text embeddings and time classes. The resulting 24-hour class probabilities are shown in \cref{fig:text_queries}. 

The results clearly demonstrate that texts describing specific times of day are directly associated with corresponding time periods. In addition, we also observe indirect associations. For example, the word \texttt{``breakfast''} is by definition related to morning hours, while \texttt{``thief''} is often associated with nighttime activities. These uneven probability distributions across the 24-hour timeline reflect the natural relations between certain events, scenes, or concepts and their corresponding time periods.

However, some irregular trends in the probability distributions indicate that our time-aware embeddings, learned from a limited image dataset, still have room for further improvements, particularly for night-time related concepts, which corresponds with fewer night-time image samples in the dataset. This highlights the need for further improvement of the dataset/model to achieve more robust time-awareness across all clock time periods.

\label{sec:text_query}
    \vspace{-3mm}

%% file: main.bib
@String(CVPR= {IEEE Conf. Comput. Vis. Pattern Recog.})

@String(ICCV= {Int. Conf. Comput. Vis.})

@String(ECCV= {Eur. Conf. Comput. Vis.})

@String(AAAI = {AAAI})

@String(CVPRW= {IEEE Conf. Comput. Vis. Pattern Recog. Worksh.})

@String(CVPR  = {CVPR})

@String(ICCV  = {ICCV})

@String(ECCV  = {ECCV})

@String(CVPRW= {CVPRW})

@article{vamplew,
author = {Adams, Anthony and Vamplew, Peter},
year = {1998},
month = {01},
pages = {},
title = {Encoding and Decoding Cyclic Data},
volume = {16},
journal = {The South Pacific Journal of Natural Science}
}

@inproceedings{Zhou_2019_CVPR,
title={On the Continuity of Rotation Representations in Neural Networks},
author={Zhou, Yi and Barnes, Connelly and Jingwan, Lu and Jimei, Yang and Hao, Li},
booktitle={The IEEE Conference on Computer Vision and Pattern Recognition (CVPR)},
month={June},
year={2019}
}

@article{thomee2016yfcc100m,
author = "Bart Thomee and David A. Shamma and Gerald Friedland and Benjamin Elizalde and Karl Ni and Douglas Poland and Damian Borth and Li-Jia Li",
title = "{YFCC100M}: The New Data in Multimedia Research",
journal = "Communications of the {ACM}",
volume = "59",
number = "2",
pages = "64--73",
year = "2016",
url = "http://cacm.acm.org/magazines/2016/2/197425-yfcc100m/fulltext",
}

@inproceedings{salem2020dynamic,
author = {Salem, Tawfiq and Workman, Scott and Jacobs, Nathan},
booktitle = {{IEEE Conference on Computer Vision and Pattern Recognition (CVPR)}},
title = {{Learning a Dynamic Map of Visual Appearance}},
year = {2020},
}

@inproceedings{DBSCAN,
author = {Ester, Martin and Kriegel, Hans-Peter and Sander, J\"{o}rg and Xu, Xiaowei},
title = {A density-based algorithm for discovering clusters in large spatial databases with noise},
year = {1996},
publisher = {AAAI Press},
booktitle = {Proceedings of the Second International Conference on Knowledge Discovery and Data Mining},
pages = {226–231},
numpages = {6},
location = {Portland, Oregon},
series = {KDD'96}
}

@misc{ho2020denoisingdiffusionprobabilisticmodels,
      title={Denoising Diffusion Probabilistic Models}, 
      author={Jonathan Ho and Ajay Jain and Pieter Abbeel},
      year={2020},
      eprint={2006.11239},
      archivePrefix={arXiv},
      primaryClass={cs.LG},
      url={https://arxiv.org/abs/2006.11239}, 
}

@inproceedings{miao2021vspw,
  title={Vspw: A large-scale dataset for video scene parsing in the wild},
  author={Miao, Jiaxu and Wei, Yunchao and Wu, Yu and Liang, Chen and Li, Guangrui and Yang, Yi},
  booktitle={Proceedings of the IEEE/CVF conference on computer vision and pattern recognition},
  pages={4133--4143},
  year={2021}
}

@article{padilha2022content,
  title={Content-aware detection of temporal metadata manipulation},
  author={Padilha, Rafael and Salem, Tawfiq and Workman, Scott and Andal{\'o}, Fernanda A and Rocha, Anderson and Jacobs, Nathan},
  journal={IEEE Transactions on Information Forensics and Security},
  volume={17},
  pages={1316--1327},
  year={2022},
  publisher={IEEE}
}

@misc{salem2022timestamp,
  title={Timestamp estimation from outdoor scenes},
  author={Salem, Tawfiq and Hwang, Jisoo and Padilha, Rafael},
  year={2022}
}

@misc{zhai2019learning,
      title={Learning Geo-Temporal Image Features}, 
      author={Menghua Zhai and Tawfiq Salem and Connor Greenwell and Scott Workman and Robert Pless and Nathan Jacobs},
      year={2019},
      eprint={1909.07499},
      archivePrefix={arXiv},
      primaryClass={id='cs.CV' full_name='Computer Vision and Pattern Recognition' is_active=True alt_name=None in_archive='cs' is_general=False description='Covers image processing, computer vision, pattern recognition, and scene understanding. Roughly includes material in ACM Subject Classes I.2.10, I.4, and I.5.'}
}

@INPROCEEDINGS{ZhangAIPR,
  author={Zhang, Zeyu and Baker, Callista and Azam-Naseeruddin, Noor and Shen, Jingzhou and Pless, Robert},
  booktitle={2022 IEEE Applied Imagery Pattern Recognition Workshop (AIPR)}, 
  title={What Does Learning About Time Tell About Outdoor Scenes?}, 
  year={2022},
  volume={},
  number={},
  pages={1-6},
  doi={10.1109/AIPR57179.2022.10092235}}

@book{murphy2012machine,
  title={Machine Learning: A Probabilistic Perspective},
  author={Murphy, Kevin P},
  year={2012},
  publisher={MIT Press},
  address={Cambridge, MA},
  chapter={1.4.5},
  isbn={978-0-262-01802-9}
}

@misc{Sharma2016,
  title = {Automated Image Timestamp Inference Using Convolutional Neural Networks},
  author = {Prafull Sharma and Michel Schoemaker and David Pan},
  institution = {Stanford University},
  year = {2016},
  url = {https://cs231n.stanford.edu/reports/2016/pdfs/267_Report.pdf}
}

@inproceedings{MIRFLICKR1M,
    author = {Huiskes, Mark J. and Lew, Michael S.},
    title = {The MIR flickr retrieval evaluation},
    year = {2008},
    isbn = {9781605583129},
    publisher = {Association for Computing Machinery},
    address = {New York, NY, USA},
    url = {https://doi.org/10.1145/1460096.1460104},
    doi = {10.1145/1460096.1460104},
    booktitle = {Proceedings of the 1st ACM International Conference on Multimedia Information Retrieval},
    pages = {39–43},
    numpages = {5},
    location = {Vancouver, British Columbia, Canada},
    series = {MIR '08}
}

@inproceedings{jacobs09webcamgis,
  author = {Jacobs, Nathan and Burgin, Walker and Fridrich, Nick and Abrams, Austin and Miskell, Kylia and Braswell, Bobby H. and Richardson, Andrew D. and Pless, Robert},
  booktitle = {ACM SIGSPATIAL International Conference on Advances in Geographic Information Systems (ACM SIGSPATIAL)},
  title = {The Global Network of Outdoor Webcams: Properties and Applications},
  year = {2009},
  month = nov,
  note = {Acceptance rate: 20.9\%},
  pages = {111--120},
  doi = {10.1145/1653771.1653789},
  pubtype = {oral presentation},
  thumbnail = {/thumbnails/webcam_network.jpg}
}

@inproceedings{jacobs07amos,
  author = {Jacobs, Nathan and Roman, Nathaniel and Pless, Robert},
  booktitle = {IEEE Conference on Computer Vision and Pattern Recognition (CVPR)},
  title = {Consistent Temporal Variations in Many Outdoor Scenes},
  year = {2007},
  month = jun,
  note = {Acceptance rate: 23.4\%},
  pages = {1--6},
  doi = {10.1109/CVPR.2007.383258},
  project = {/research/global-webcam-archive/},
  thumbnail = {/thumbnails/consistent.jpg}
}

@INPROCEEDINGS{CVL_CAMS,
  author={Volokitin, Anna and Timofte, Radu and Van Gool, Luc},
  booktitle={2016 IEEE Conference on Computer Vision and Pattern Recognition Workshops (CVPRW)}, 
  title={Deep Features or Not: Temperature and Time Prediction in Outdoor Scenes}, 
  year={2016},
  volume={},
  number={},
  pages={1136-1144},
  doi={10.1109/CVPRW.2016.145}
}

@inproceedings{geoclip,
  title={GeoCLIP: Clip-Inspired Alignment between Locations and Images for Effective Worldwide Geo-localization},
  author={Vivanco, Vicente and Nayak, Gaurav Kumar and Shah, Mubarak},
  booktitle={Advances in Neural Information Processing Systems},
  year={2023}
}

@misc{radford2021learningtransferablevisualmodels,
      title={Learning Transferable Visual Models From Natural Language Supervision}, 
      author={Alec Radford and Jong Wook Kim and Chris Hallacy and Aditya Ramesh and Gabriel Goh and Sandhini Agarwal and Girish Sastry and Amanda Askell and Pamela Mishkin and Jack Clark and Gretchen Krueger and Ilya Sutskever},
      year={2021},
      eprint={2103.00020},
      archivePrefix={arXiv},
      primaryClass={cs.CV},
      url={https://arxiv.org/abs/2103.00020}, 
}

@article{johnson2019billion,
  title={Billion-scale similarity search with {GPUs}},
  author={Johnson, Jeff and Douze, Matthijs and J{\'e}gou, Herv{\'e}},
  journal={IEEE Transactions on Big Data},
  volume={7},
  number={3},
  pages={535--547},
  year={2019},
  publisher={IEEE}
}

@incollection{MOORE1992101,
title = {Chapter 8 - The organization of the human circadian timing system},
editor = {D.F. Swaab and M.A. Hofman and M. Mirmiran and R. Ravid and F.W. {Van Leeuwen}},
series = {Progress in Brain Research},
publisher = {Elsevier},
volume = {93},
pages = {101-117},
year = {1992},
booktitle = {The Human Hypothalamus in Health and Disease},
issn = {0079-6123},
doi = {https://doi.org/10.1016/S0079-6123(08)64567-7},
url = {https://www.sciencedirect.com/science/article/pii/S0079612308645677},
author = {Robert Y. Moore}
}

@article{DUFFY2009165,
title = {Effect of Light on Human Circadian Physiology},
journal = {Sleep Medicine Clinics},
volume = {4},
number = {2},
pages = {165-177},
year = {2009},
note = {Basics of Circadian Biology and Circadian Rhythm Sleep Disorders},
issn = {1556-407X},
doi = {https://doi.org/10.1016/j.jsmc.2009.01.004},
url = {https://www.sciencedirect.com/science/article/pii/S1556407X09000058},
author = {Jeanne F. Duffy and Charles A. Czeisler}
}

@book{dohrn1996history,
  title={History of the hour: Clocks and modern temporal orders},
  author={Dohrn-van Rossum, Gerhard},
  year={1996},
  publisher={University of Chicago Press}
}

@misc{agarwal2021evaluatingclipcharacterizationbroader,
      title={Evaluating CLIP: Towards Characterization of Broader Capabilities and Downstream Implications}, 
      author={Sandhini Agarwal and Gretchen Krueger and Jack Clark and Alec Radford and Jong Wook Kim and Miles Brundage},
      year={2021},
      eprint={2108.02818},
      archivePrefix={arXiv},
      primaryClass={cs.CV},
      url={https://arxiv.org/abs/2108.02818}, 
}

@ARTICLE{5128907,
  author={He, Haibo and Garcia, Edwardo A.},
  journal={IEEE Transactions on Knowledge and Data Engineering}, 
  title={Learning from Imbalanced Data}, 
  year={2009},
  volume={21},
  number={9},
  pages={1263-1284},
  doi={10.1109/TKDE.2008.239}
}

@inproceedings{tong2022videomae,
  title={Video{MAE}: Masked Autoencoders are Data-Efficient Learners for Self-Supervised Video Pre-Training},
  author={Zhan Tong and Yibing Song and Jue Wang and Limin Wang},
  booktitle={Advances in Neural Information Processing Systems},
  year={2022}
}

@InProceedings{marszalek09,
    author = "Marcin Marsza{\l}ek and Ivan Laptev and Cordelia Schmid",
    title = "Actions in Context",
    booktitle = "IEEE Conference on Computer Vision \& Pattern Recognition",
    year = "2009"
}

@inproceedings{wu2018unsupervised,
  title={Unsupervised Feature Learning via Non-Parametric Instance Discrimination},
  author={Wu, Zhirong and Xiong, Yuanjun and Stella, X Yu and Lin, Dahua},
  booktitle={Proceedings of the IEEE Conference on Computer Vision and Pattern Recognition},
  year={2018}
}

@misc{oquab2023dinov2,
  title={DINOv2: Learning Robust Visual Features without Supervision},
  author={Oquab, Maxime and Darcet, Timothée and Moutakanni, Theo and Vo, Huy V. and Szafraniec, Marc and Khalidov, Vasil and Fernandez, Pierre and Haziza, Daniel and Massa, Francisco and El-Nouby, Alaaeldin and Howes, Russell and Huang, Po-Yao and Xu, Hu and Sharma, Vasu and Li, Shang-Wen and Galuba, Wojciech and Rabbat, Mike and Assran, Mido and Ballas, Nicolas and Synnaeve, Gabriel and Misra, Ishan and Jegou, Herve and Mairal, Julien and Labatut, Patrick and Joulin, Armand and Bojanowski, Piotr},
  journal={arXiv:2304.07193},
  year={2023}
}

@software{torchvision2016,
  title        = {TorchVision: PyTorch's Computer Vision library},
  author       = {TorchVision maintainers and contributors},
  year         = 2016,
  journal      = {GitHub repository},
  publisher    = {GitHub},
  howpublished = {\url{https://github.com/pytorch/vision}}
}

@misc{wolf2020huggingfacestransformersstateoftheartnatural,
      title={HuggingFace's Transformers: State-of-the-art Natural Language Processing}, 
      author={Thomas Wolf and Lysandre Debut and Victor Sanh and Julien Chaumond and Clement Delangue and Anthony Moi and Pierric Cistac and Tim Rault and Rémi Louf and Morgan Funtowicz and Joe Davison and Sam Shleifer and Patrick von Platen and Clara Ma and Yacine Jernite and Julien Plu and Canwen Xu and Teven Le Scao and Sylvain Gugger and Mariama Drame and Quentin Lhoest and Alexander M. Rush},
      year={2020},
      eprint={1910.03771},
      archivePrefix={arXiv},
      primaryClass={cs.CL},
      url={https://arxiv.org/abs/1910.03771}, 
}

@article{sangalli2021constrained,
  title={Constrained optimization to train neural networks on critical and under-represented classes},
  author={Sangalli, Sara and Erdil, Ertunc and H{\"o}tker, Andeas and Donati, Olivio and Konukoglu, Ender},
  journal={Advances in neural information processing systems},
  volume={34},
  pages={25400--25411},
  year={2021}
}

@INPROCEEDINGS{6247815,
  author={Derpanis, Konstantinos G. and Lecce, Matthieu and Daniilidis, Kostas and Wildes, Richard P.},
  booktitle={2012 IEEE Conference on Computer Vision and Pattern Recognition}, 
  title={Dynamic scene understanding: The role of orientation features in space and time in scene classification}, 
  year={2012},
  volume={},
  number={},
  pages={1306-1313},
  doi={10.1109/CVPR.2012.6247815}}

@misc{wang2023flowdynamicscorrectionaction,
      title={Flow Dynamics Correction for Action Recognition}, 
      author={Lei Wang and Piotr Koniusz},
      year={2023},
      eprint={2310.10059},
      archivePrefix={arXiv},
      primaryClass={cs.CV},
      url={https://arxiv.org/abs/2310.10059}, 
}

@inproceedings{chen2024x360,
  title={360+x: A Panoptic Multi-modal Scene Understanding Dataset},
  author={Chen, Hao and Hou, Yuqi and Qu, Chenyuan and Testini, Irene and Hong, Xiaohan and Jiao, Jianbo},
  booktitle={Proceedings of the IEEE/CVF Conference on Computer Vision and Pattern Recognition},
  year={2024}
}

@article{ALIS,
  title={Aligning Latent and Image Spaces to Connect the Unconnectable},
  author={Skorokhodov, Ivan and Sotnikov, Grigorii and Elhoseiny, Mohamed},
  journal={arXiv preprint arXiv:2104.06954},
  year={2021}
}

@inproceedings{Karras2019stylegan2,
  title     = {Analyzing and Improving the Image Quality of {StyleGAN}},
  author    = {Tero Karras and Samuli Laine and Miika Aittala and Janne Hellsten and Jaakko Lehtinen and Timo Aila},
  booktitle = {Proc. CVPR},
  year      = {2020}
}

@article{yu15lsun,
    Author = {Yu, Fisher and Zhang, Yinda and Song, Shuran and Seff, Ari and Xiao, Jianxiong},
    Title = {LSUN: Construction of a Large-scale Image Dataset using Deep Learning with Humans in the Loop},
    Journal = {arXiv preprint arXiv:1506.03365},
    Year = {2015}
}

@InProceedings{Patashnik_2021_ICCV,
    author    = {Patashnik, Or and Wu, Zongze and Shechtman, Eli and Cohen-Or, Daniel and Lischinski, Dani},
    title     = {StyleCLIP: Text-Driven Manipulation of StyleGAN Imagery},
    booktitle = {Proceedings of the IEEE/CVF International Conference on Computer Vision (ICCV)},
    month     = {October},
    year      = {2021},
    pages     = {2085-2094}
}

@Article{liu2022convnet,
  author  = {Zhuang Liu and Hanzi Mao and Chao-Yuan Wu and Christoph Feichtenhofer and Trevor Darrell and Saining Xie},
  title   = {A ConvNet for the 2020s},
  journal = {Proceedings of the IEEE/CVF Conference on Computer Vision and Pattern Recognition (CVPR)},
  year    = {2022},
}

@misc{tan2021efficientnetv2smallermodelsfaster,
      title={EfficientNetV2: Smaller Models and Faster Training}, 
      author={Mingxing Tan and Quoc V. Le},
      year={2021},
      eprint={2104.00298},
      archivePrefix={arXiv},
      primaryClass={cs.CV},
      url={https://arxiv.org/abs/2104.00298}, 
}

@misc{liu2022swintransformerv2scaling,
      title={Swin Transformer V2: Scaling Up Capacity and Resolution}, 
      author={Ze Liu and Han Hu and Yutong Lin and Zhuliang Yao and Zhenda Xie and Yixuan Wei and Jia Ning and Yue Cao and Zheng Zhang and Li Dong and Furu Wei and Baining Guo},
      year={2022},
      eprint={2111.09883},
      archivePrefix={arXiv},
      primaryClass={cs.CV},
      url={https://arxiv.org/abs/2111.09883}, 
}

@misc{gal2021stylegannada,
      title={StyleGAN-NADA: CLIP-Guided Domain Adaptation of Image Generators},
      author={Rinon Gal and Or Patashnik and Haggai Maron and Gal Chechik and Daniel Cohen-Or},
      year={2021},
      eprint={2108.00946},
      archivePrefix={arXiv},
      primaryClass={cs.CV}
}

@misc{kwon2022clipstylerimagestyletransfer,
      title={CLIPstyler: Image Style Transfer with a Single Text Condition}, 
      author={Gihyun Kwon and Jong Chul Ye},
      year={2022},
      eprint={2112.00374},
      archivePrefix={arXiv},
      primaryClass={cs.CV},
      url={https://arxiv.org/abs/2112.00374}, 
}

@article{epstein2022blobgan,
    title={BlobGAN: Spatially Disentangled Scene Representations},
    author={Dave Epstein and Taesung Park and Richard Zhang and Eli Shechtman and Alexei A. Efros},
    journal={European Conference on Computer Vision (ECCV)},
    year={2022}
}

@misc{pinkney_lhq-sg2-1024,
  author = {Justin Pinkney},
  title = {lhq-sg2-1024},
  year = {2024},
  publisher = {Hugging Face},
  note = {StyleGAN2 model trained on the LHQ dataset},
  url = {https://huggingface.co/justinpinkney/lhq-sg2-1024},
}

@inproceedings{Karras2020ada,
  title     = {Training Generative Adversarial Networks with Limited Data},
  author    = {Tero Karras and Miika Aittala and Janne Hellsten and Samuli Laine and Jaakko Lehtinen and Timo Aila},
  booktitle = {Proc. NeurIPS},
  year      = {2020}
}

@misc{parmar2024onestepimagetranslationtexttoimage,
      title={One-Step Image Translation with Text-to-Image Models}, 
      author={Gaurav Parmar and Taesung Park and Srinivasa Narasimhan and Jun-Yan Zhu},
      year={2024},
      eprint={2403.12036},
      archivePrefix={arXiv},
      primaryClass={cs.CV},
      url={https://arxiv.org/abs/2403.12036}, 
}

@ARTICLE{10361537,
  author={Keetha, Nikhil and Mishra, Avneesh and Karhade, Jay and Jatavallabhula, Krishna Murthy and Scherer, Sebastian and Krishna, Madhava and Garg, Sourav},
  journal={IEEE Robotics and Automation Letters}, 
  title={AnyLoc: Towards Universal Visual Place Recognition}, 
  year={2024},
  volume={9},
  number={2},
  pages={1286-1293},
  doi={10.1109/LRA.2023.3343602}}

@inproceedings{conforti2020automatic,
  title={Automatic repair of same-timestamp errors in business process event logs},
  author={Conforti, Raffaele and La Rosa, Marcello and Ter Hofstede, Arthur HM and Augusto, Adriano},
  booktitle={Business Process Management: 18th International Conference, BPM 2020, Seville, Spain, September 13--18, 2020, Proceedings 18},
  pages={327--345},
  year={2020},
  organization={Springer}
}

@misc{Seitzer2020FID,
  author={Maximilian Seitzer},
  title={{pytorch-fid: FID Score for PyTorch}},
  month={August},
  year={2020},
  note={Version 0.3.0},
  howpublished={\url{https://github.com/mseitzer/pytorch-fid}},
}

@inproceedings{NIPS2017_8a1d6947,
 author = {Heusel, Martin and Ramsauer, Hubert and Unterthiner, Thomas and Nessler, Bernhard and Hochreiter, Sepp},
 booktitle = {Advances in Neural Information Processing Systems},
 editor = {I. Guyon and U. Von Luxburg and S. Bengio and H. Wallach and R. Fergus and S. Vishwanathan and R. Garnett},
 pages = {},
 publisher = {Curran Associates, Inc.},
 title = {GANs Trained by a Two Time-Scale Update Rule Converge to a Local Nash Equilibrium},
 url = {https://proceedings.neurips.cc/paper_files/paper/2017/file/8a1d694707eb0fefe65871369074926d-Paper.pdf},
 volume = {30},
 year = {2017}
}

@article{rodriguez2018beyond,
  title={Beyond one-hot encoding: Lower dimensional target embedding},
  author={Rodr{\'\i}guez, Pau and Bautista, Miguel A and Gonzalez, Jordi and Escalera, Sergio},
  journal={Image and Vision Computing},
  volume={75},
  pages={21--31},
  year={2018},
  publisher={Elsevier}
}

@Article{he2019moco,
  author  = {Kaiming He and Haoqi Fan and Yuxin Wu and Saining Xie and Ross Girshick},
  title   = {Momentum Contrast for Unsupervised Visual Representation Learning},
  journal = {arXiv preprint arXiv:1911.05722},
  year    = {2019},
}

@inproceedings{10.1145/3474085.3479207,
author = {Tang, Mingkang and Wang, Zhanyu and LIU, Zhenhua and Rao, Fengyun and Li, Dian and Li, Xiu},
title = {CLIP4Caption: CLIP for Video Caption},
year = {2021},
isbn = {9781450386517},
publisher = {Association for Computing Machinery},
address = {New York, NY, USA},
url = {https://doi.org/10.1145/3474085.3479207},
doi = {10.1145/3474085.3479207},
booktitle = {Proceedings of the 29th ACM International Conference on Multimedia},
pages = {4858–4862},
numpages = {5},
location = {Virtual Event, China},
series = {MM '21}
}

@misc{rombach2021highresolution,
      title={High-Resolution Image Synthesis with Latent Diffusion Models}, 
      author={Robin Rombach and Andreas Blattmann and Dominik Lorenz and Patrick Esser and Björn Ommer},
      year={2021},
      eprint={2112.10752},
      archivePrefix={arXiv},
      primaryClass={cs.CV}
}

@misc{
liu2021fusedream,
title={FuseDream: Training-Free Text-to-Image Generation with Improved CLIP+GAN Space Optimization}, 
author={Xingchao Liu and Chengyue Gong and Lemeng Wu and Shujian Zhang and Hao Su and Qiang Liu},
year={2021},
eprint={2112.01573},
archivePrefix={arXiv},
primaryClass={cs.CV}
}

@misc{kazemi2019time2veclearningvectorrepresentation,
      title={Time2Vec: Learning a Vector Representation of Time}, 
      author={Seyed Mehran Kazemi and Rishab Goel and Sepehr Eghbali and Janahan Ramanan and Jaspreet Sahota and Sanjay Thakur and Stella Wu and Cathal Smyth and Pascal Poupart and Marcus Brubaker},
      year={2019},
      eprint={1907.05321},
      archivePrefix={arXiv},
      primaryClass={cs.LG},
      url={https://arxiv.org/abs/1907.05321}, 
}

@INPROCEEDINGS{8099985,
  author={Carreira, João and Zisserman, Andrew},
  booktitle={2017 IEEE Conference on Computer Vision and Pattern Recognition (CVPR)}, 
  title={Quo Vadis, Action Recognition? A New Model and the Kinetics Dataset}, 
  year={2017},
  volume={},
  number={},
  pages={4724-4733},
  doi={10.1109/CVPR.2017.502}}

@inproceedings{NIPS2007_013a006f,
 author = {Rahimi, Ali and Recht, Benjamin},
 booktitle = {Advances in Neural Information Processing Systems},
 editor = {J. Platt and D. Koller and Y. Singer and S. Roweis},
 pages = {},
 publisher = {Curran Associates, Inc.},
 title = {Random Features for Large-Scale Kernel Machines},
 url = {https://proceedings.neurips.cc/paper_files/paper/2007/file/013a006f03dbc5392effeb8f18fda755-Paper.pdf},
 volume = {20},
 year = {2007}
}

@misc{BLIP,
  doi = {10.48550/ARXIV.2201.12086},
  
  url = {https://arxiv.org/abs/2201.12086},
  
  author = {Li, Junnan and Li, Dongxu and Xiong, Caiming and Hoi, Steven},
  
  title = {BLIP: Bootstrapping Language-Image Pre-training for Unified Vision-Language Understanding and Generation},
  
  publisher = {arXiv},
  
  year = {2022},
  
  copyright = {Creative Commons Attribution 4.0 International}
}

@misc{openai2024gpt4ocard,
      title={GPT-4o System Card}, 
      author={OpenAI and : and Aaron Hurst and Adam Lerer and Adam P. Goucher and Adam Perelman and Aditya Ramesh and Aidan Clark and AJ Ostrow and Akila Welihinda and Alan Hayes and Alec Radford and Aleksander Mądry and Alex Baker-Whitcomb and Alex Beutel and Alex Borzunov and Alex Carney and Alex Chow and Alex Kirillov and Alex Nichol and Alex Paino and Alex Renzin and Alex Tachard Passos and Alexander Kirillov and Alexi Christakis and Alexis Conneau and Ali Kamali and Allan Jabri and Allison Moyer and Allison Tam and Amadou Crookes and Amin Tootoochian and Amin Tootoonchian and Ananya Kumar and Andrea Vallone and Andrej Karpathy and Andrew Braunstein and Andrew Cann and Andrew Codispoti and Andrew Galu and Andrew Kondrich and Andrew Tulloch and Andrey Mishchenko and Angela Baek and Angela Jiang and Antoine Pelisse and Antonia Woodford and Anuj Gosalia and Arka Dhar and Ashley Pantuliano and Avi Nayak and Avital Oliver and Barret Zoph and Behrooz Ghorbani and Ben Leimberger and Ben Rossen and Ben Sokolowsky and Ben Wang and Benjamin Zweig and Beth Hoover and Blake Samic and Bob McGrew and Bobby Spero and Bogo Giertler and Bowen Cheng and Brad Lightcap and Brandon Walkin and Brendan Quinn and Brian Guarraci and Brian Hsu and Bright Kellogg and Brydon Eastman and Camillo Lugaresi and Carroll Wainwright and Cary Bassin and Cary Hudson and Casey Chu and Chad Nelson and Chak Li and Chan Jun Shern and Channing Conger and Charlotte Barette and Chelsea Voss and Chen Ding and Cheng Lu and Chong Zhang and Chris Beaumont and Chris Hallacy and Chris Koch and Christian Gibson and Christina Kim and Christine Choi and Christine McLeavey and Christopher Hesse and Claudia Fischer and Clemens Winter and Coley Czarnecki and Colin Jarvis and Colin Wei and Constantin Koumouzelis and Dane Sherburn and Daniel Kappler and Daniel Levin and Daniel Levy and David Carr and David Farhi and David Mely and David Robinson and David Sasaki and Denny Jin and Dev Valladares and Dimitris Tsipras and Doug Li and Duc Phong Nguyen and Duncan Findlay and Edede Oiwoh and Edmund Wong and Ehsan Asdar and Elizabeth Proehl and Elizabeth Yang and Eric Antonow and Eric Kramer and Eric Peterson and Eric Sigler and Eric Wallace and Eugene Brevdo and Evan Mays and Farzad Khorasani and Felipe Petroski Such and Filippo Raso and Francis Zhang and Fred von Lohmann and Freddie Sulit and Gabriel Goh and Gene Oden and Geoff Salmon and Giulio Starace and Greg Brockman and Hadi Salman and Haiming Bao and Haitang Hu and Hannah Wong and Haoyu Wang and Heather Schmidt and Heather Whitney and Heewoo Jun and Hendrik Kirchner and Henrique Ponde de Oliveira Pinto and Hongyu Ren and Huiwen Chang and Hyung Won Chung and Ian Kivlichan and Ian O'Connell and Ian O'Connell and Ian Osband and Ian Silber and Ian Sohl and Ibrahim Okuyucu and Ikai Lan and Ilya Kostrikov and Ilya Sutskever and Ingmar Kanitscheider and Ishaan Gulrajani and Jacob Coxon and Jacob Menick and Jakub Pachocki and James Aung and James Betker and James Crooks and James Lennon and Jamie Kiros and Jan Leike and Jane Park and Jason Kwon and Jason Phang and Jason Teplitz and Jason Wei and Jason Wolfe and Jay Chen and Jeff Harris and Jenia Varavva and Jessica Gan Lee and Jessica Shieh and Ji Lin and Jiahui Yu and Jiayi Weng and Jie Tang and Jieqi Yu and Joanne Jang and Joaquin Quinonero Candela and Joe Beutler and Joe Landers and Joel Parish and Johannes Heidecke and John Schulman and Jonathan Lachman and Jonathan McKay and Jonathan Uesato and Jonathan Ward and Jong Wook Kim and Joost Huizinga and Jordan Sitkin and Jos Kraaijeveld and Josh Gross and Josh Kaplan and Josh Snyder and Joshua Achiam and Joy Jiao and Joyce Lee and Juntang Zhuang and Justyn Harriman and Kai Fricke and Kai Hayashi and Karan Singhal and Katy Shi and Kavin Karthik and Kayla Wood and Kendra Rimbach and Kenny Hsu and Kenny Nguyen and Keren Gu-Lemberg and Kevin Button and Kevin Liu and Kiel Howe and Krithika Muthukumar and Kyle Luther and Lama Ahmad and Larry Kai and Lauren Itow and Lauren Workman and Leher Pathak and Leo Chen and Li Jing and Lia Guy and Liam Fedus and Liang Zhou and Lien Mamitsuka and Lilian Weng and Lindsay McCallum and Lindsey Held and Long Ouyang and Louis Feuvrier and Lu Zhang and Lukas Kondraciuk and Lukasz Kaiser and Luke Hewitt and Luke Metz and Lyric Doshi and Mada Aflak and Maddie Simens and Madelaine Boyd and Madeleine Thompson and Marat Dukhan and Mark Chen and Mark Gray and Mark Hudnall and Marvin Zhang and Marwan Aljubeh and Mateusz Litwin and Matthew Zeng and Max Johnson and Maya Shetty and Mayank Gupta and Meghan Shah and Mehmet Yatbaz and Meng Jia Yang and Mengchao Zhong and Mia Glaese and Mianna Chen and Michael Janner and Michael Lampe and Michael Petrov and Michael Wu and Michele Wang and Michelle Fradin and Michelle Pokrass and Miguel Castro and Miguel Oom Temudo de Castro and Mikhail Pavlov and Miles Brundage and Miles Wang and Minal Khan and Mira Murati and Mo Bavarian and Molly Lin and Murat Yesildal and Nacho Soto and Natalia Gimelshein and Natalie Cone and Natalie Staudacher and Natalie Summers and Natan LaFontaine and Neil Chowdhury and Nick Ryder and Nick Stathas and Nick Turley and Nik Tezak and Niko Felix and Nithanth Kudige and Nitish Keskar and Noah Deutsch and Noel Bundick and Nora Puckett and Ofir Nachum and Ola Okelola and Oleg Boiko and Oleg Murk and Oliver Jaffe and Olivia Watkins and Olivier Godement and Owen Campbell-Moore and Patrick Chao and Paul McMillan and Pavel Belov and Peng Su and Peter Bak and Peter Bakkum and Peter Deng and Peter Dolan and Peter Hoeschele and Peter Welinder and Phil Tillet and Philip Pronin and Philippe Tillet and Prafulla Dhariwal and Qiming Yuan and Rachel Dias and Rachel Lim and Rahul Arora and Rajan Troll and Randall Lin and Rapha Gontijo Lopes and Raul Puri and Reah Miyara and Reimar Leike and Renaud Gaubert and Reza Zamani and Ricky Wang and Rob Donnelly and Rob Honsby and Rocky Smith and Rohan Sahai and Rohit Ramchandani and Romain Huet and Rory Carmichael and Rowan Zellers and Roy Chen and Ruby Chen and Ruslan Nigmatullin and Ryan Cheu and Saachi Jain and Sam Altman and Sam Schoenholz and Sam Toizer and Samuel Miserendino and Sandhini Agarwal and Sara Culver and Scott Ethersmith and Scott Gray and Sean Grove and Sean Metzger and Shamez Hermani and Shantanu Jain and Shengjia Zhao and Sherwin Wu and Shino Jomoto and Shirong Wu and Shuaiqi and Xia and Sonia Phene and Spencer Papay and Srinivas Narayanan and Steve Coffey and Steve Lee and Stewart Hall and Suchir Balaji and Tal Broda and Tal Stramer and Tao Xu and Tarun Gogineni and Taya Christianson and Ted Sanders and Tejal Patwardhan and Thomas Cunninghman and Thomas Degry and Thomas Dimson and Thomas Raoux and Thomas Shadwell and Tianhao Zheng and Todd Underwood and Todor Markov and Toki Sherbakov and Tom Rubin and Tom Stasi and Tomer Kaftan and Tristan Heywood and Troy Peterson and Tyce Walters and Tyna Eloundou and Valerie Qi and Veit Moeller and Vinnie Monaco and Vishal Kuo and Vlad Fomenko and Wayne Chang and Weiyi Zheng and Wenda Zhou and Wesam Manassra and Will Sheu and Wojciech Zaremba and Yash Patil and Yilei Qian and Yongjik Kim and Youlong Cheng and Yu Zhang and Yuchen He and Yuchen Zhang and Yujia Jin and Yunxing Dai and Yury Malkov},
      year={2024},
      eprint={2410.21276},
      archivePrefix={arXiv},
      primaryClass={cs.CL},
      url={https://arxiv.org/abs/2410.21276}, 
}

@inproceedings{hwang2019segsort,
  title={SegSort: Segmentation by Discriminative Sorting of Segments},
  author={Hwang, Jyh-Jing and Yu, Stella X and Shi, Jianbo and Collins, Maxwell D and Yang, Tien-Ju and Zhang, Xiao and Chen, Liang-Chieh},
  booktitle={Proceedings of the IEEE International Conference on Computer Vision},
  pages={7334--7344},
  year={2019}
}
